\ifdefined\useorstyle

\documentclass[opre,nonblindrev]{informs3}

\usepackage{./macros/packages-or}
\usepackage{./macros/editing-macros}
\usepackage{./macros/statistics-macros-or}
\usepackage{./macros/formatting}

\DoubleSpacedXI 

\usepackage{endnotes}
\let\footnote=\endnote

\usepackage{natbib}
 \bibpunct[, ]{(}{)}{,}{a}{}{,}%
 \def\bibfont{\small}%
 \def\bibsep{\smallskipamount}%
\TheoremsNumberedThrough     
\ECRepeatTheorems

\EquationsNumberedThrough    

\usepackage{hyperref}
\usepackage{pgfplotstable}
\usepackage{graphicx}
\usepackage{subcaption}
\usepackage{float} 
\usepackage{tabularx}
\usepackage{overpic}
\usepackage{tikz}
\usepackage{rotating}
\usepackage{psfrag}
\usepackage{enumitem}
\usepackage{booktabs}

\begin{document}



\RUNAUTHOR{Che and Namkoong}
\RUNTITLE{Adaptive Experimentation at Scale}

\TITLE{Adaptive Experimentation at Scale: A Computational Framework for Flexible
  Batches}

\ARTICLEAUTHORS{%

\AUTHOR{Ethan Che}
\AFF{Decision, Risk, and Operations Division, Columbia Business School, New York, NY, \EMAIL{namkoong@gsb.columbia.edu}} 

\AUTHOR{Daniel Jiang}
\AFF{Meta, University of Pittsburgh, New York, NY, \EMAIL{drjiang@meta.com}} 

\AUTHOR{Hongseok Namkoong}
\AFF{Decision, Risk, and Operations Division, Columbia Business School, New York, NY, \EMAIL{namkoong@gsb.columbia.edu}} 

\AUTHOR{Jimmy Wang}
\AFF{Columbia University, New York, NY, \EMAIL{jw4209@columbia.edu}} 

\ABSTRACT{

Real-world experiments involve batched \& delayed feedback, non-stationarity, multiple objectives \& constraints, and  (often some)  personalization. 
Tailoring adaptive methods to address 
these challenges on a per-problem basis is infeasible, and a static design remains the de facto standard.
Focusing on short-horizon ($\le 10$) adaptive experiments, we move away from bespoke algorithms and present a \emph{mathematical programming} formulation that can flexibly incorporate a 
wide range of objectives, constraints, and statistical procedures. 
We formulating a dynamic program based on central limit approximations, which  enables the use of scalable optimization methods based on auto-differentiation and GPU parallelization.  
To evaluate our framework, we implement a simple heuristic planning method (``solver'') and benchmark it across hundreds of problem instances involving non-stationarity, 
personalization, and multiple objectives \& constraints. 
Unlike bespoke methods (e.g., Thompson sampling variants), our mathematical programming 
framework provides consistent gains over static randomized control trials and exhibits 
robust performance across problem instances.

}



\maketitle

%


\else

\documentclass[11pt]{article}
\usepackage[numbers]{natbib}
\usepackage{./macros/packages}
\usepackage{./macros/editing-macros}
\usepackage{./macros/formatting}
\usepackage{./macros/statistics-macros}
\usepackage{multirow}
\usepackage{booktabs}


\begin{document}

\abovedisplayskip=8pt plus0pt minus3pt
\belowdisplayskip=8pt plus0pt minus3pt


\begin{center}
  {\huge Optimization-Driven Adaptive Experimentation} \\
  \vspace{.5cm} {\Large Ethan Che$^{1}$ ~~~ Daniel Jiang$^{2,3}$ ~~~ Hongseok Namkoong$^{1}$  ~~~ Jimmy Wang$^{1}$} \\
  \vspace{.2cm}
  {\large $^{1}$Columbia University, $^{2}$Meta, $^{3}$University of Pittsburgh} \\
  \vspace{.2cm}
  \texttt{\{eche25, namkoong\}@gsb.columbia.edu, drjiang@meta.com, jw4209@columbia.edu}
\end{center}


\begin{abstract}%
  
\end{abstract}

\fi

\section{Introduction}
\label{section:introduction}

Experimentation forms the foundation of scientific decision-making, from
natural and social sciences to industry. Yet even for online platforms with
hundreds of millions of users, achieving sufficient statistical power is
costly~\cite{KohaviLoSoHe09,KohaviDeFrLoWaXu12, KohaviDeFrWaXuPo13,
  ButtonIoMoNoFlRoMu13, CziborJiLi19}. Adaptive experimentation can
significantly improve efficiency by focusing resources on promising
treatments, and expand the set of scientific hypotheses that can be tested
beyond the typical handful of treatment options. However, significant practical challenges remain in applying standard adaptive
algorithms:
\vspace{-.1cm}
\begin{itemize}[itemsep=-3pt, leftmargin=0.55cm]
\item \textbf{Batching, delayed feedback, and short horizons}: Standard algorithms are designed to be
updated after every observation, but experiments are typically conducted with a few large batches and infrequent updates to the sampling policy due to
infrastructure constraints and delayed feedback \citep{BakshyEtAl18,
OfferWestortCoGr20, JobjornssonScMuFr22}. This results in problem formulations with limited adaptivity, or, in other words, formulations with finite and short horizons.
\item \textbf{Nonstationarity}: Real-world experiments operate in non-stationary
  environments. In a week-long experiment comparing multiple layouts for the
  landing page of a fashion retailer, behavior of customers who visit the page
  on the weekend is different from those that visit on Monday.

\item \textbf{General objectives and metric constraints}: Across a range of
  outcomes/metrics/rewards, there is a wide range of objectives and
  constraints that practitioners may consider such as within- vs. post-experiment performance.
\item \textbf{Cost/budget constraints}: Practitioners running a field experiment 
  wish to find the best treatment while satisfying budget constraints and ethical constraints 
  involving the social welfare of participants during the experiment. 
  Other constraints may include requiring
  sufficient sample coverage for post-experiment inference
  \citep{OfferWestortCoGr20, ZhangJaMu20}, requiring fairness \citep{chen20a},
  and ensuring safety \citep{amani19}. 
\end{itemize}

\begin{figure}[t]
    \vspace{-1cm}
  \centering
  \includegraphics[width=1.1\textwidth]{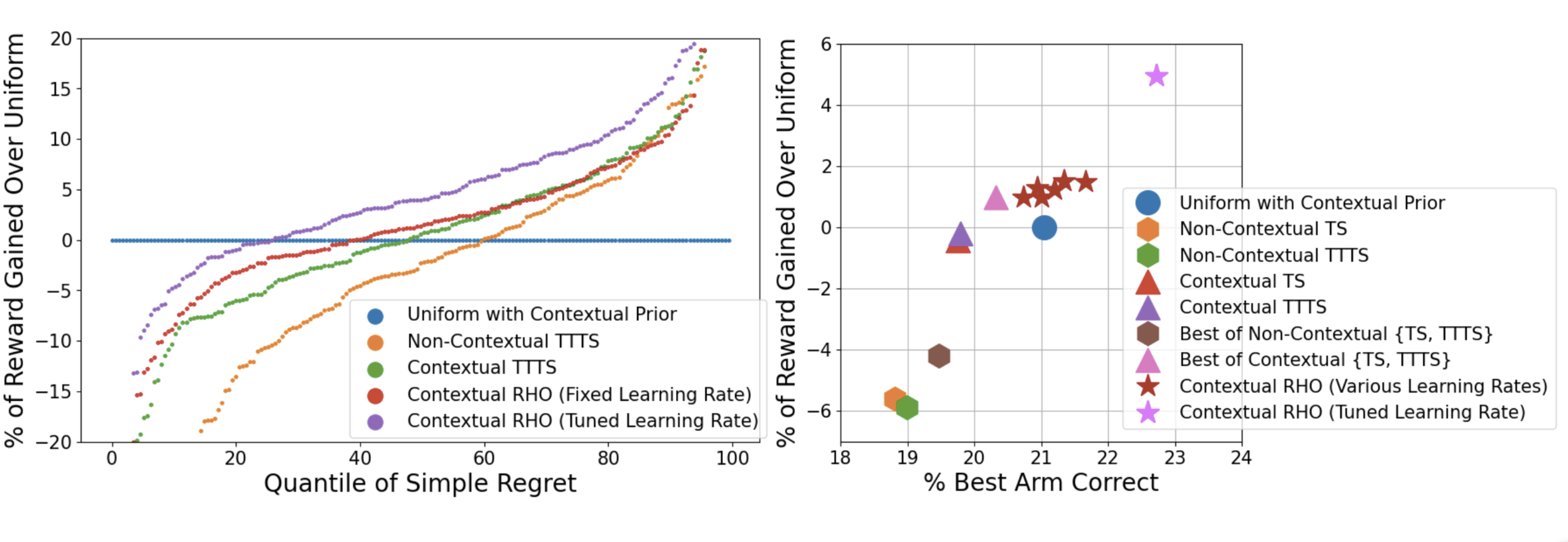}
  \caption{\textbf{(Real-world non-stationarity)} Benchmark results on 241 non-stationary settings based on 78 real
    experiments run at ASOS, a fashion retailer with over 26 million active
    customers as of 2024~\citep{liu2021datasets}. Treatment effects vary
    significantly over days. We simulate 750
    batched experiments across instances, resulting in 180,750 evaluations
    across different batch sizes and different policies.  Plots shown for
    batch size $n_{t}=100$ and $T = 10$.  Contextual policies model non-stationarity effects, but still pick a single arm at the end of the experiment; we compare $\model$ with variants of Thompson
    sampling, including Contextual Top-Two Thompson sampling
    (TTTS)~\cite{QinRusso2023} tailored for non-stationary settings.
    \textbf{Right:} \emph{Adaptive algorithms often do worse than uniform,
       overfitting on initial, temporary performance.}
    $\algo$ (stars) is the only adaptive policy with greater average reward
    and choose the best arm more often compared to uniform sampling,
    outperforming bandit algorithms even under model misspecification (non-contextual), across
    a wide range of learning rates.  \textbf{Left:} Quantile of simple regret
    across 241 settings (normalized by that of uniform allocation).  $\algo$
    outperforms uniform on more instances (60.5\%) compared to other adaptive
    policies (51.8\% for Top-Two Thompson Sampling).  TS-based policies tend
    to be more fragile on difficult instances; when it underperforms Uniform
    it does so 10.7\% on average (compared to 6.9\% for $\algo$).}
  \label{fig:main_plot}
\end{figure}

\noindent \newline \textbf{Long-horizon bandit  algorithms   (Section~\ref{section:setting}).} To address these challenges, the standard algorithm development
paradigm takes a problem-specific approach; researchers craft tailored methods
that deliver strong theoretical guarantees for problems with frequent opportunities to reallocate measurement effort (``long horizon'')~\cite{JoulaniGySz13, PerchetRiChSn16,
  JunJaNoZh16, AgarwalEtAl17, Pike-BurkeAgSzGr18, GroverEtAl18, LattimoreSz19,
  GaoHaReZh19, VernadeCaLaZaErBr20, EsfandiariKaMeMi21,
  QinRusso2023,fiez2024three}.  However, it is infeasible for practitioners to
develop a new algorithm for each new setting they encounter in practice. Deploying algorithms to settings that are slightly different than what they
were designed for can lead to highly suboptimal performance that is even worse than
that of static uniformly randomized allocations (see
Figure~\ref{fig:main_plot}). Additionally, maintaining a portfolio of
problem-specific methods for each possible instance leads to fragmented infrastructures that are
cumbersome to manage.

To understand why the literature takes this approach, consider the adaptive
experiment sequential decision making problem
  \begin{align}
    \label{eqn:oae-dp-informal}
  \scalebox{0.95}{$\displaystyle \min_{\pi_{t}(\mc{H}_t) \ge 0}
    \left\{ \mathbb{E}\left[\sum_{t=0}^T
    \mathsf{Objective}_t(\pi_t; \mc{H}_t)\right] 
  ~~\bigg|~~
    \sum_{t=0}^{T} \mathsf{Cost}_t(\pi_{t}, \mc{H}_t) \leq B,
  ~~1^\top \pi_{t}(\mc{H}_t) \leq 1~~ \forall t
  \right\}$,}
\end{align}
where $T$ is the number of updates to the sampling allocation (horizon), $\mc{H}_t$
denotes historical data (outcomes, contexts, treatment assignments) up to the
$t$-th batch, $\pi_t$ denotes the policy the modeler needs to learn, and $B$ is the budget 
constraint of the experiment. In this paper, we focus on short-horizon ($T \le 10$) problems to model batched real-world experiments where reallocation of measurement effort is costly.

The dynamic program~\eqref{eqn:oae-dp-informal} is intractable since the dimension
of the states $\mc{H}_t$ is commensurate with the total number of
observations. 
Even though they are usually designed for good performance as $T$ grows large,
bandit algorithms \citep{LattimoreSz19} are commoly applied as an approximate solutionmethod.
 They optimize a surrogate such
as an upper confidence bound (UCB) derived from concentration inequalities or a model drawn from a posterior distribution (e.g., Top-Two Thompson
sampling). While these proxies do not directly measure the regret objective of
interest, they nonetheless serve as useful approximations that guide sampling
decisions to reduce regret as the horizon $T$ increases.  However, these
proxy-based approaches are difficult to adapt to the aforementioned challenges that arise in real-world experiments, and it is unclear whether their performance guarantees
will be maintained after suitable modifications. 

As a result, most experiments in practice are non-adaptive
\citep{SculleyEtAl15, AgarwalEtAl16}.
In this paper, we focus on  short-horizon batched adaptive experiments ($T \le 10$) and do not consider large-horizon behavior. Our goal is to uniformly improve upon non-adaptive designs across multiple problem instances.

\noindent \newline \textbf{Mathematical programming via tractable reformulation (Section~\ref{section:algorithms}).}
We present a scalable and flexible \emph{mathematical
  programming} view for \emph{batched} adaptive experiments. It is instructive to use
deterministic optimization as a fable. There, a practitioner writes down a problem instance
consisting of decision variables, an objective, and constraints in a standardized
interface (e.g., CVX~\cite{GrantBo11}) that gets passed to a general solver (e.g.,
Gurobi~\cite{Gurobi}, MOSEK~\cite{mosek}).  We argue \emph{short-horizon} adaptive experimentation problems can be viewed in an analogous way. Practitioners should be able to express their objectives, constraints, and structure of the assignment policy through a
\emph{standardized and flexible language} that includes simple objectives like
best arm identification, or complex ones like best top-$k$ selection while
minimizing cumulative regret. Under this mathematical programming view, an
effective adaptive method must perform well across a wide set of problem
instances, rather than focus only on a specific set of problems.

Toward this vision, we simplify the dynamic program~\eqref{eqn:oae-dp-informal} to a tractable optimization problem that
we can heuristically solve using modern computational tools (auto-differentiation and GPU parallelization). Our main observation is 
that a \emph{batched} view of the experimentation
process allows us to model the uncertainty around sufficient statistics
necessary to make allocation decisions, instead of attempting to model
unit-level outcomes whose distributions are commonly unknown and leads to
intractable dynamic programs with combinatorial action
spaces~\cite{BhatFaMoSi20}.
We consider any outcomes whose conditional means can be learned via a loss
minimization problem over a parametric model class $\theta \in \Theta$.  (In a
vanilla multi-armed setting, the true parameter $\theta\opt$ corresponds to
average rewards.)  Instead of an entire batch of data, we
compress the information into a sufficient statistic, the empirical risk
minimizer $\what{\theta}_t$, which is calculated over each batch.

Letting $n_t$ be the batch size at  reallocation epoch $t$, the central limit approximation implies
\begin{equation}
  \label{eqn:likelihood-informal}
  \what{\theta}_t \sim N(\theta\opt, n_t^{-1} F(\pi_t))
  ~~\mbox{for some known function}~F(\cdot)~\mbox{of sampling allocation}~\pi_t.
\end{equation}
Here, we view $\what{\theta}_t$ as a noisy observation of the true parameter
$\theta\opt$, or equivalently, that the \emph{data} $\what{\theta}_t$ has
approximately Gaussian likelihoods given the parameter $\theta\opt$.
We model the experimenter's evolving beliefs on the true parameter $\theta\opt$
as $\what{\theta}_t$'s are observed. In particular, a Gaussian prior on $\theta\opt$ with prior mean $\beta_0$ and variance $\Sigma_0$
and the approximate
likelihood~\eqref{eqn:likelihood-informal} gives conjugate posteriors
\[
(\beta_0, \Sigma_0) \xrightarrow{\what{\theta}_1} (\beta_1, \Sigma_1) \xrightarrow{\what{\theta}_2} \cdots \xrightarrow{\what{\theta}_{T}} (\beta_T, \Sigma_T).
\]
By modeling posterior beliefs $(\beta_t, \Sigma_t)$ as states, we arrive at the
\emph{batch limit dynamic program} ($\model$) with known dynamics
  \begin{align}
    \label{eqn:bldp-informal}
  \min_{\pi_{t}(\beta_t, \Sigma_t) \ge 0}
    \left\{ \mathbb{E}\left[\sum_{t=0}^T
    \mathsf{Objective}_t(\pi_t; \beta_t, \Sigma_t)\right] 
  \Big|\,
        \sum_{t=0}^{T}\mathsf{Cost}(\pi_{t}) \leq B, \,
  ~~1^\top \pi_{t}(\beta_t, \Sigma_t) \leq 1~ \forall t 
  \right\}.
\end{align}

\begin{figure}[t]
  \centering
  \includegraphics[width=0.65\textwidth]{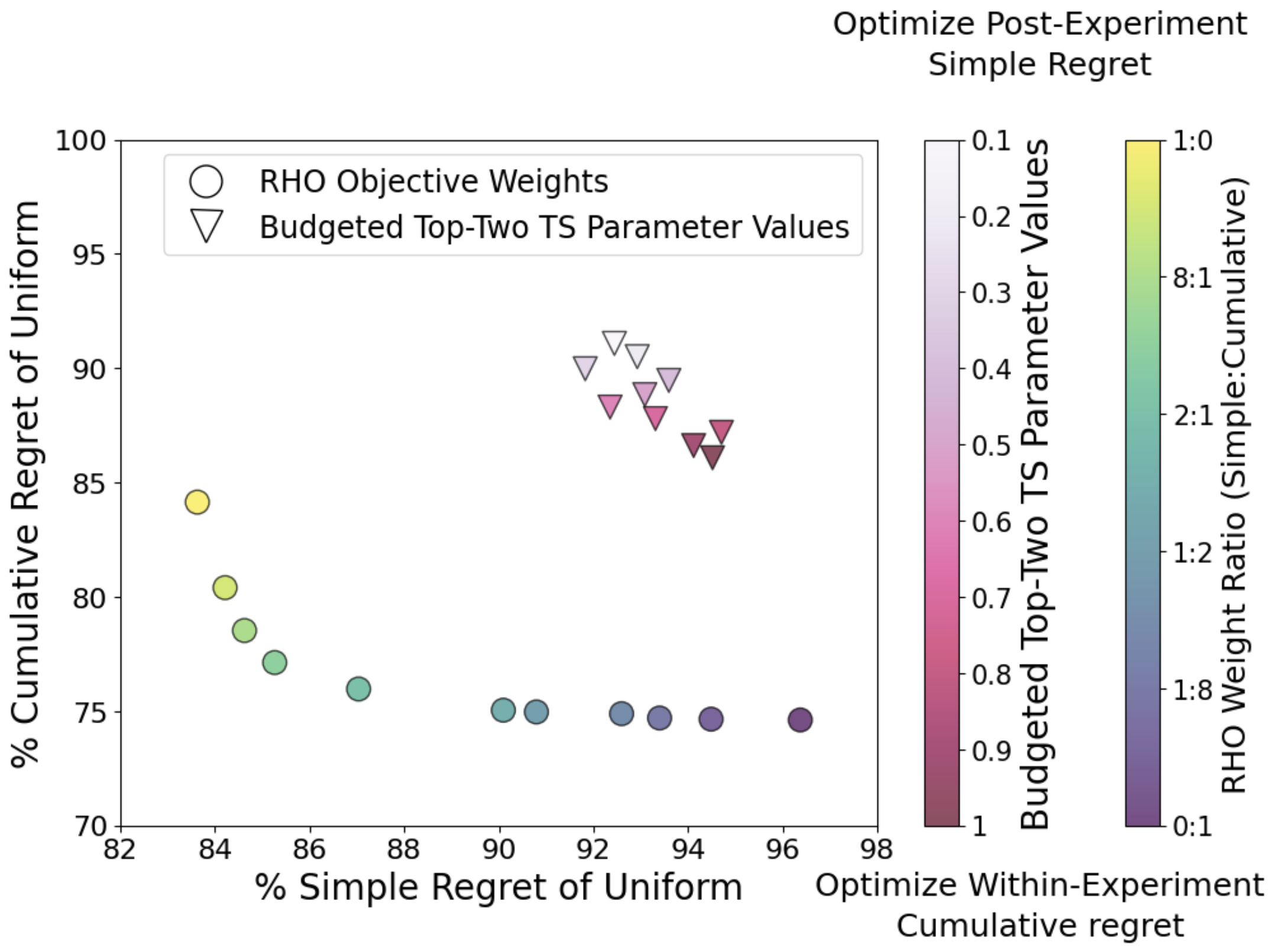}
  \caption{\textbf{Satisfying budget constraints while optimizing simple plus cumulative regret.} Pareto frontier of the tradeoff between simple and cumulative regret while maintaining a fixed budget of $100$ units across the experiment. Each arm is associated with a fixed cost. $\algo$ can exactly satisfy the budget constraint in expectation via projected gradient descent. Additionally, $\algo$ trades off 
  between simple and cumulative regret in a principled manner by setting weights equal to the number of individuals under each objective. We are not aware of any existing adaptive algorithm designed for this setting, so we combine Budgeted TS~\cite{xia2015thompson} designed for budget constraints and Top-Two TS~\cite{Russo20} which can tradeoff between simple and cumulative regret~\cite{qin2024optimize}. In contrast to $\algo$, it is difficult to tune Budgeted Top-Two TS in an exact and principled way to satisfy the objective and constraints. This method involves a scaling parameter $b\in [0, \infty)$ to satisfy the budget constraint and a scaling parameter $\alpha \in [0,1]$ to trade off between simple and cumulative regret. These parameters are difficult to tune as adjusting one parameter directly affects the other. Keeping the budget parameter $b$ constant while varying $\alpha$ to tradeoff between simple and cumulative regret yields average budget costs from 82.2 to 105.2 units, potentially violating the constraint. In contrast, keeping $\algo$'s budget costs fixed yields average budget costs varying from 94.1 to 99.5 units.  
  }
  \label{fig:pareto-constraints}
\end{figure}

To concretely illustrate the benefits of the tractable mathematical
programming reformulation $\model$~\eqref{eqn:bldp-informal}, consider an experiment ubiquitous in practice involving both minimizing \textit{within experiment} (cumulative) regret and \text{post experiment} (simple) regret while satisfying a budget constraint.  For example, experimenters running a field experiment in the social sciences strive to maximize the welfare of participants (cumulative regret) due to ethical considerations while satisfying cost constraints and exploring enough to determine the efficacy of each treatment (simple regret). An online platform wishes to conduct an experiment while limiting negative user experience (cumulative regret) and ensuring experiment constraints are satisfied. To our knowledge, no tailored algorithm for this setting even for long horizons, even though separate algorithms have been proposed to satisfy budget constraints~\cite{xia2015thompson} and optimize the trade-off between simple and cumulative regret~\cite{Russo20, qin2024optimize, caria2023adaptive}. \cite{wilder2024treat} proposes a method to learn treatment effects while treating those in need and satisfying budget constraints, but can only handle single stage designs. In contrast, this setup is easy to model in the $\model$ as we detail in Section~\ref{section:bayesian-mdp}; see Section~\ref{section:experiments} for an empirical demonstration. Figure~\ref{fig:pareto-constraints} shows that our formulation is able to efficiently tradeoff between these objectives in a principled manner, while adaptations of existing methods are difficult to tune and satisfy objectives and constraints in a principled or exact manner. 

\looseness-1 Central limit approximations are universal in statistical inference, and we use it to model the
sufficient statistics $\what{\theta}_t$ as a Gaussian distribution
(``likelihood'').
We do not require the experimenter to be Bayesian
\emph{when analyzing the data}, but combine frequentist CLT approximations~\eqref{eqn:likelihood-informal} with Bayesian principles to \emph{optimize
  adaptive designs}. In our framework, Gaussianity is a \emph{result} of the CLT
rather than a blanket assumption, and
justifying the CLT approximation~\eqref{eqn:likelihood-informal} requires significant effort.
The standard CLT only holds
conditional on the history, and proving joint convergence over $t = 1, \ldots, T$ requires careful control of the \emph{rate} at which convergence occurs. Our main theoretical contribution (Section~\ref{section:asymptotics})
proves this under weak conditions, by combining
 M-estimation theory and Stein's method. 

\noindent \newline \textbf{Optimizing short-horizon adaptive experiments (Section~\ref{section:rho}).} 
Since the $\model$~\eqref{eqn:bldp-informal} has simple closed-form dynamics (posterior
updates),  we can simulate trajectories (``roll-outs'') under various treatment
allocations and \emph{plan} future sampling allocations for short horizons $(T \le 10)$. As an initial step toward approximate solution methods for $\model$, we introduce a simple method based on the
\emph{model-predictive control} (MPC) design principle:
\emph{residual horizon optimization} ($\algo$). Whenever new
observations are obtained, the policy simulates trajectories from the $\model$
to solve for an optimal \emph{static} sequence of treatment allocations to use
for the rest of experiment. By re-solving this sequence after every batch, the
policy remains adaptive while working flexibly with batched feedback.  
$\algo$'s main advantage is that regardless of the complexity of  objectives, constraints, and
predictable non-stationarity patterns,   it is  guaranteed to have a lower Bayesian regret than static designs. While static designs appear to be a weak benchmark initially,  we empirically find  bandit algorithms struggle to consistently outperform a standard A/B test when facing aforementioned practical challenges. 

Since dynamics are differentiable with
respect to sampling allocations, our proposed algorithm can optimize lookahead
policies (with short horizons) over any set of objectives and constraints that can be expressed as a
function of posterior beliefs (e.g., top-$k$-sum, simple, and cumulative regret)
through stochastic gradient descent methods.  The use of pathwise stochastic
gradients obtained through auto-differentiation provides a effective and
simple method compared to REINFORCE-style policy gradient methods.   Our formulation is not
only applicable for personalized treatments or policy learning, but  also
 improving upon the efficiency and
robustness of standard bandit experiments.
Given the vast literature on adaptive methods, we provide a necessarily abridged review in Sections~\ref{section:algorithms} and~\ref{section:discussion}.

\noindent \newline \textbf{Rigorous empirical benchmarking (Section~\ref{section:experiments}).}
In our mathematical programming framework, adaptive designs are learned based
on scalable optimization algorithms regardless of the problem instance. 
To assess the potential of our mathematical programming framework,
we empirically validate our simple solution approach $\algo$ over diverse problem instances. In particular, we benchmark adaptive designs against static uniform
allocations (``A/B tests'', the de facto standard), along with state-of-the-art adaptive methods like Thompson sampling~\cite{agrawal13thompson} and its extensions that account for (predictable) non-stationarity~\cite{QinRusso2023} and various constraints and objectives~\cite{xia2015thompson, Russo20, qin2024optimize}. Despite the preliminary and limited nature of $\algo$, it consistently outperforms these state-of-the-art methods over a wide variety of settings involving non-stationarity,
personalization, multiple objectives, and constraints. In Figure~\ref{fig:main_plot}, we provide a preview of our empirical work, where we find that $\algo$
exhibits a unique robustness to non-stationarity. Figure \ref{fig:pareto-constraints} shows that $\algo$ can trade off between the cumulative and simple regret objectives while exactly maintaining a fixed budget in expectation via projected gradient descent. This shows that incorporating multiple, possibly competing objectives and constraints into our mathematical programming framework is straightforward. We are not aware of existing methods designed for this setting despite its ubiquity. Other methods such as Budgeted Thompson Sampling~\cite{xia2015thompson} and Top-Two Thompson Sampling~\cite{Russo20} can theoretically tradeoff between these objectives individually, but there is no principled approach to setting the values of these parameters. Furthermore, Figure \ref{fig:pareto-ranking} displays $\algo$'s ability to efficiently trade off between a combined cumulative and simple regret objective in a personalized ranking environment, showing our framework's applications in diverse areas.



\section{Batched Adaptive Experimentation}
\label{section:setting}

We begin by formalizing short-horizon adaptive experiments as a sequential decision-making problem and highlight the challenges of solving this problem directly. Given the widespread practice of conducting experiments in a few, large
batches, we consider a \emph{batched adaptive experiment} composed of $T$
sequential epochs (or ``batches'') and $K$ treatment arms.  Each epoch
$t = 0,\ldots,T-1$ comprises of $n_{t}$ sampling units, where each unit
$i =1, \ldots, n_t$ has a \emph{context} $X_{t,i} \in \mathcal{X} \subseteq \R^{p}$
drawn i.i.d. from a potentially time-varying batch context distribution $\mu_{t}$.

In this work, we restrict attention to \emph{predictable} non-stationarity and assume that the context distributions $\mu_t$ are known to the modeler. While
this may seem like a strong requirement, this is often naturally satisfied when
modeling temporal non-stationarity (e.g., day-of-the-week or novelty
effects). Temporal non-stationarity can be modeled by having the context be
the epoch $x = t$: in this case future context arrival distributions are
$\mu_{s}(x) = \indic{x = s}$ and are deterministic.  In more complicated settings,
e.g., if the context are highly user-specific features, we can instead use
historical samples from the context distributions $\mu_{s}$ as an approximation.

The experimenter is interested in a stochastic \emph{outcome}
$R_{t,i} \in \R$, along with the effect of each treatment arm $a\in [K]$ on the
outcome.  In many applications, the experimenter is interested in multiple
outcome metrics, in which case $R_{t,i} \in \R^{m}$ for $m>1$. While the
framework we introduce can seamlessly handle this case, we focus on a single
outcome metric for clarity of presentation. We denote
$\mathcal{D}_{t} = \{ (X_{t,i},A_{t,i},R_{t,i}) \}_{i=1}^{n_{t}}$ to be the data
collected in batch $t$.

The experimenter allocates treatments to sampling units to optimize the
learning objective subject to constraints, both of which may depend on
outcomes. Defining $\mathcal{S}_{K} := \{(x_{1}, ..., x_{K}) \mid x_{i} \geq 0 \text{ and } \sum_{i=1}^{K} x_{i} \; \forall i \in [K]\}$ be the "filled" simplex, the decision variables are \emph{sampling allocations}
$p_{t}(\cdot \,|\, x) \in \mathcal{S}_{K}$, which are conditional probabilities over treatment arms $a\in [K]$. This
includes non-contextual sampling allocations  which allocate a fixed
fraction of the sampling units to each treatment arm. Adaptivity means  allocations may  depend on the \emph{history} $ \mathcal{H}_{t} := \bigcup_{s=0}^{t-1} \mathcal{D}_{s}$ of sampled contexts, arm allocations, and
rewards observed before
epoch $t$.

We allow $p_{t}(\cdot \,|\, x)$ to sum to less than one instead of lying on the $K$-simplex, in order to satisfy potential budget constraints. Notably, we allow the zero-vector as an action, i.e. $p_{t}(\cdot \,|\, x) = \textbf{0}$ allowing the experimenter to abstain from assigning a treatment to take a zero-cost action to maintain feasbility during the experiment. 

Experiments are often evaluated based on the quality of downstream deployment decisions after the experiment is run. To model this, 
we can also consider a
\emph{post-experiment deployment decision} of a treatment arm or even a policy
allocating treatments over a population distribution $X \sim \mu$. We
characterize this as the \emph{post-experimental} phase, in which the experimenter
uses the data collected during the experiment to make a final treatment
allocation.
\begin{definition}
\label{defn:post_exp}
The final epoch $t = T$ is the \textbf{post-experimental phase} after the batched adaptive experiment performed during epochs $t=0,1, \ldots, T-1$. 
During this phase, based on the history $\mathcal{H}_{T}$,
the experimenter selects a final arm allocation decision over treatment arms $p_{T}(\cdot \,|\, x) \in \Delta_{K}$
for units from a \textbf{population distribution} $\mu$.
\end{definition}

The final allocation $p_T$ naturally depends on sampling allocations made
during the experiment.
Due to non-stationarity, the sampling units that arrive during the experiment may be drawn from distributions $\mu_{t}$ that are very
different from the population distribution $\mu$.  Therefore, if the adaptive design does
not account for non-stationarity, it may overfit to data that is not representative of the treatment effects for the population distribution. 

\subsection{Reward Models}
\label{section:rewards}

Given an observed context $X_{t,i} = x$ and assigned action $A_{t,i} = a$,
the stochastic rewards $R_{t,i}$ follow the conditional distribution
\begin{equation*}
  R_{t,i} \mid X_{t,i} = x, \, A_{t,i} = a \sim \nu_t(\cdot \mid x, a),
\end{equation*}
with mean $r(x,a) := \E [R_{t,i} \,|\, X_{t,i} = x, A_{t,i} = a]$.
Crucially, we do not assume $\nu_{t}(\cdot \,|\,x,a)$ is known to the
experimenter.  We assume that the primary object of interest, the conditional mean
reward $r(x,a)$, is given by a general parametric model
\begin{equation}
  \label{eqn:parametric reward model}
 r(x,a)=f(x, a; \theta^\ast)
  ~~~\mbox{for some}~~\theta\opt \in \R^{d}.
\end{equation}
Common examples include linear and linear-logistic
contextual models, given respectively by
\begin{align}\label{eqn:linear reward model}  
r(x,a)=\phi(x,a)^{\top}\theta\opt \quad \textnormal{and} \quad 
r(x,a)=\sigma(\phi(x,a)^{\top}\theta\opt),
\end{align} 
where $\phi(x,a):\mathcal{X}\times[K]\to\mathbb{R}^{d}$ is a known feature mapping
and $\sigma(x) = 1/(1 + e^{-x})$ is the sigmoid.

The set of reward models we consider can model a range of application
scenarios.  Contexts can range from user-specific features to
day-of-the-week or geographic fixed effects. 

\vspace{0.5em}
\begin{example}[Additive treatment effects with confounders]
  Let $\theta\opt = (\theta\opt_{1}, \ldots, \theta\opt_{K}, \bar{\theta}\opt)\in \mathbb{R}^{K+p}$
  consist of an unknown linear parameter $\bar{\theta}\opt \in \mathbb{R}^p$, along with an additive treatment effect $\theta\opt_a \in \mathbb{R}$ associated with each arm $a \in [K]$. Under the model
    $r(x,a) = \phi(x,a)^\top \theta\opt
    = x^\top \bar{\theta}\opt + \theta\opt_a$,
  there exists a single arm that maximizes expected reward for all contexts $x$.
\end{example}

\begin{example}[Mixed contextual treatment effects]
  Let $\theta\opt = (\bar{\theta}\opt_{1}, \ldots, \bar{\theta}\opt_{K}) \in \mathbb{R}^{Kp}$ consist of
  action embeddings $\bar{\theta}\opt_{a}\in\mathbb{R}^{p}$. Under the model
    $r(x,a) = \phi(x,a)^{\top}\theta\opt =x^{\top}\bar{\theta}\opt_{a}$,
  the optimal action varies across $x$, this setup can model personalization (when $x$
  represents user-specific features), or  predictable non-stationarity (when $x$ represents the day-of-the-week).
\end{example}

\vspace{-.1cm}
We assume the experimenter's statistical procedure for estimating $\theta\opt$
is specified through a \emph{loss function} $\ell$, mirroring common practice
as well as the theoretical framework of M-estimation.  This includes
estimation procedures such as least-squares, maximum likelihood, quantile
regression, as well as robust (Huber) losses.  In order to connect this loss
function to the underlying parameter of interest, we assume that the true
parameter $\theta\opt$ minimizes the expected loss under a suitable
population distribution, along with standard regularity assumptions.
\begin{assumption}[True parameter specified by loss minimization]
\label{assumption:parametric}
  Let $\ell(\theta \,|\, x, a, r)$ be a convex, twice continuously differentiable
  loss in $\theta$ for all $(x,a,r)$. Under a population distribution over
  contexts $X_{i} \sim \mu$, a reference action distribution
  $A_{i} \sim p(\cdot\,|\,X_{i})$, and under the reward distribution
  $R_{i} \sim \nu_{t}(\cdot \,|\, X_{i},A_{i})$, $\theta\opt$ is the unique
  minimizer of the population loss $L(\theta)$:
  \begin{align}
    \label{eqn:population_loss}
    & \theta\opt = \arg \min_{\theta \in \Theta}
      \left\{ L(\theta)
      := \arg \min_{\theta \in \Theta}
      \E \bigl [\ell(\theta \,|\, X_i, A_i, R_i) \bigr]\right\}.
  \end{align}
  Furthermore, $L(\theta)$ has a positive definite Hessian on a neighborhood
  of $\theta\opt$.
\end{assumption}
For the linear and the linear-logistic models, the corresponding loss functions are as follows,
\[
  \ell(\theta \,|\, X_{i},A_{i},R_{i}) = | R_{i} - \phi(X_{i},A_{i})^{\top} \theta |^{2},
  \quad \ell(\theta \,|\, X_{i},A_{i},R_{i}) = \log\left( 1 + e^{-R_{i} \phi(X_{i},A_{i})^{\top} \theta} \right).
\]
In order to estimate $\theta\opt$ from a dataset $\mathcal{D}$ consisting of $n$
i.i.d. tuples of contexts, arms, and rewards $\bigl\{(X_{i}, A_{i}, R_{i})\bigr\}_{i=1}^{n}$,
the experimenter solves the empirical risk minimization problem
    $\what{\theta}_n = \arg \min_{\theta \in \Theta}
    \left\{L_{n}(\theta \,|\, \mathcal{D})
      \defeq \frac{1}{n}\sum_{i=1}^{n} \ell(\theta \,|\, X_{i}, A_{i}, R_{i})
    \right\}$.



\subsection{Challenges of Optimization for Adaptive Experimentation}
\label{section:oae-dp}
For short horizon adaptive experiments~\eqref{eqn:oae-dp-informal}, a natural approach is to frame the problem as a Markov decision process (MDP) and use approximate dynamic programming to optimize the experimenter's objective (e.g. cumulative or simple regret). In the absence of assumptions on the reward distribution $R_{t, i} \sim \nu(\cdot \,|\, x,a)$, states are the history of observed data $\mathcal{H}_{t}$, which 
is intractable as the dimension of the history is proportional to the total number of sampling units seen so far $\sum\nolimits_{s=0}^{t}n_{s}$. 
This is huge in
  typical experiments (e.g., $n_{s} \ge 10^{5}$ in typical A/B tests) and the curse of dimensionality prevents the use of standard dynamic programming techniques. It is unclear how to properly distill the history $\mathcal{H}_{t}$ into something more manageable without additional assumptions on the distribution $\nu(\cdot \,|\, x,a)$ of rewards $R_{t,i}$. 
  The state transition $\mathcal{H}_{t+1} = \mathcal{H}_{t} \cup \mathcal{D}_{t}$ relies on newly observed rewards, but the reward-generating distribution $R_{t,i} \stackrel{iid}{\sim} \nu(\cdot \,|\, x,a)$ is unknown.

It may be possible to design approaches for overcoming these issues, such as (online) reinforcement learning or constructing a Bayesian posterior over $\nu$, but again, since the state is $\mathcal{H}_t$, such approaches will be impractical. The intractability of solving this dynamic program has been widely recognized
in the bandit literature and has served as motivation for specialized algorithms that are guaranteed to perform well only in large horizon problems. 
As a result, they may not be able to manage exploration-exploitation trade-offs for short horizons, let alone trade-offs among multiple objectives and constraints.


\label{section:rewards}

\section{Batch Limit Dynamic Program}
\label{section:algorithms}



We now derive a tractable approximation with \emph{known dynamics} that only relies on
$\mathcal O(d^2)$-dimensional states where $d$ is the dimension of the parameter
$\theta\opt$ (Assumption~\ref{assumption:parametric}). 
We leverage the fact
that typical experiments feature large enough batch sizes~\citep{BakshyEtAl18,
  OfferWestortCoGr20, JobjornssonScMuFr22} to admit a central limit
approximation, a universal practice in statistical inference. 
In particular, we use a normal approximation for relevant suffficient statistics 
at the batch level.

Our work directly extends the framework
of~\citet{CheNa23} in simple multi-armed settings to allow nonstationarities,
multiple objectives, constraints, and personalized policies. Our mathematical
programming view is enabled by our main theoretical results which derive
sequential local asymptotic normality for M-estimation problems.  Our analysis
carefully leverages Stein's method to address the sequential, adaptive nature
of the problem and provides the most general formulation of classical local
asymptotic normality and normal experiments to date~\cite{VanDerVaart98}.
Overall, the $\model$ provides a substantially more general framework than their
Gaussian sequential experiment over finite arm rewards and this generalization
allows experimenters to update a flexible and tractable Bayesian model over
the solution to a general loss minimization problem. 
Methodologically, we
extend the MPC principle to formulate a general planning algorithm---which we
still term $\algo$ for conceptual ease---that can handle any objectives and
constraints. Empirically, we demonstrate the performance of $\algo$ in Section~\ref{section:experiments} across an extensive
array of problem instances motivated by practice, compared to the simple simulation
examples in~\citet{CheNa23}.



\subsection{Large-Batch Statistical Approximations}
\label{section:large-batch}

We derive our tractable dynamic programming formulation informally in this section, deferring a rigorous treatment to Section~\ref{section:asymptotics}. Our main theoretical insight is that the following approximations are valid in the large batch limit where normal approximations become valid.

Given data $\mc{D}_{t}$ collected in batch $t$ with sampling allocation $p_{t}(\cdot\,|\,x)$, 
the experimenter can obtain an estimate $\what{\theta}_{t}$
of $\theta\opt$ by minimizing the empirical risk
\begin{equation}
  \label{eqn:batch-estimator}
  \what{\theta}_{t}
  = \argmin_{\theta \in \Theta}
  \left\{L_{n_{t}} (\theta \,|\, \mc{D}_{t})
    := \frac{1}{n_{t}} \sum_{i=1}^{n_{t}} \ell(\theta \,|\,X_{t,i}, A_{t,i}, R_{t,i})
  \right\}.
\end{equation}
Our key observation is that according to standard intuition from M-estimation, under large batch sizes (i.e. $n_{t} \to \infty$) the
distribution of $\what{\theta}_{t}$ is asymptotically Gaussian, i.e.,
\begin{equation}
  \label{eqn:informal-clt}
  \what{\theta}_{t} 
  ~~\mbox{approximately follows}~~N\bigl(\theta\opt,
    n_t^{-1} H_{t}^{-1} I_{t} H_{t}^{-1} \bigr),
\end{equation}
where the Hessian $H_{t}$ and the gradient covariance matrix $I_{t}$ are
defined as follows
\begin{subequations}
  \label{eqn:fisher}
  \begin{align}
    H_{t} &\defeq \E\left[ \sum\nolimits_{a} p_{t}(a\,|\,x) \, \nabla_{\theta}^{2} \ell(\theta\opt \,|\,X_{t,i}, a, R_{t,i}) \right],
            \label{eqn:hessian}\\
    I_{t} &\defeq \E\left[ \sum\nolimits_{a} p_{t}(a\,|\,x) \, \nabla_{\theta} \ell(\theta\opt \,|\,X_{t,i}, a, R_{t,i}) \, \nabla_{\theta} \ell(\theta\opt \,|\,X_{t,i}, a, R_{t,i})^{\top} \right].
            \label{eqn:grad_cov}
  \end{align}
\end{subequations}
For example, for the squared loss $\ell(\theta\,|\,x,a,r) = |r - \phi(x,a)^\top \theta|^{2}$,  the central limit
approximation~\eqref{eqn:informal-clt} recovers the standard asymptotics for
the ordinary least squares (OLS) estimator:
\[
    \what{\theta}_{t} 
    ~~\mbox{approximately follows}~~
    N\left(\theta\opt, n_{t}^{-1} \, \E \biggl[\sum_{a} p_{t}(a\,|\, x)\phi(X_{t,i},a) \, \phi(X_{t,i},a)^{\top}\biggr] \right).
\]
Our main theoretical
result proves that the sequence $(\what{\theta}_{0},\ldots,\what{\theta}_{T-1})$
converges jointly in distribution to a \emph{sequential Gaussian experiment},
where the each observation is Gaussian conditional on past data. Let $G_{t} \,|\, G_{0}, \ldots, G_{t-1} \sim N(\theta\opt, n_{t}^{-1} H_{t}^{-1}I_{t}H_{t}^{-1})$ be 
the limit Gaussian observation from population level quantities at time $t$. 
The main
analytical challenge in Section~\ref{section:asymptotics} comes from the fact that each $\what{\theta}_{t}$ depends
on the realizations of previous estimates
$\what{\theta}_{0},\ldots, \what{\theta}_{t-1}$, as the sampling policy uses
previous measurements to determine the sampling allocations. This requires us
to derive a novel theoretical framework that carefully quantifies how the
dependency impacts the rate at which the central limit approximation becomes
valid. Our joint convergence result holds under minimal assumptions around the
underlying sampling policy $\{\pi_t\}_{t=0}^{T-1}$, even allowing for $\{\pi_t\}_{t=0}^{T-1}$ to sample arms with
zero probability as well as allowing for $H_{t}$ to be non-invertible.


\subsection{A Bayesian MDP Formulation}
\label{section:bayesian-mdp}
So far, we have argued that the estimator
$\what{\theta}_{t}$ of \eqref{eqn:batch-estimator} provides information about
$\theta\opt$ contaminated by Gaussian noise with a known distribution.
Alternatively, we can think of this approximation as a data-generating model
where given $\theta\opt$, the observation $\what{\theta_t}$ follows a Gaussian
distribution. We adopt Bayesian principles to reason through the modeler's
uncertainty on $\theta\opt$ and guide adaptive designs, but do not assume the
experimenter is Bayesian---meaning they can use standard frequentist tools for
final statistical inference based on the data collected from the adaptive
experiment. 

Given a Gaussian likelihood $\P(\what{\theta}_t \,|\, \theta\opt)$, one can efficiently
perform Bayesian inference of $\theta\opt$ under a Gaussian prior
$\theta\opt \sim N(\beta_{0}, \Sigma_{0})$.  By conjugacy, the posterior will also be Gaussian.
\begin{fact}
  \label{fact:conjugate-normal}
  Under a Gaussian prior $\theta\opt \sim N(\beta_{0}, \Sigma_{0})$, the
  posterior distribution of $\theta\opt$ after observing a Gaussian random
  variable $G \sim N\left(\theta, n_{t}^{-1}H^{-1} I H^{-1} \right)$ is $\theta\opt \,|\, G \sim N(\beta, \Sigma)$, where
  \begin{align}
    \label{eqn:posterior_update}
    \Sigma^{-1} := \Sigma_{0}^{-1} + n_{t} H I^{-1} H
     \quad \textnormal{and} \quad \beta := \Sigma\left( \Sigma_{0}^{-1} \beta_{0} + n_{t} H I^{-1} H G \right).
  \end{align}
\end{fact}
\noindent Instead of keeping track of the entire history $\mc{H}_{t}$ of all
observations collected during the experiment, the experimenter can simply keep
track of the posterior distribution of $\theta\opt$ parameterized by
$(\beta_{t}, \Sigma_{t})$. This posterior is updated after every batch $t$,
using the asymptotic characterization of $\what{\theta}_{t}$.

The dynamics of the sequence of posterior states $\{(\beta_{t}, \Sigma_{t}) \}_{t=0}^{T}$ are controlled by the sampling allocations $p_{t}$ chosen by the experimenter, and they follow a simple closed-form expression.

\begin{lemma}
    \label{lemma:mdp}
    Under a Gaussian prior $\theta\opt \sim N(\beta_0, \Sigma_0)$, given observations
    $\{G_{t} \}_{t=1}^{T}$ where
    $G_{t} \,|\, G_{1:t-1} \sim N(\theta^{*}, n_{t}^{-1}H_{t}^{-1} I_{t} H_{t}^{-1})$, the
    joint distribution of posterior states
    $\{(\beta_{t}, \Sigma_{t}) \}_{t=0}^{T}$ is characterized by the
    following recursive equations:
    \begin{equation}
      \label{eqn:dynamics}
        \Sigma_{t+1}^{-1}
         \defeq \Sigma_{t}^{-1} + n_{t} H_{t} I_{t}^{-1} H_{t} \quad \textnormal{and} \quad \beta_{t+1}
        := \beta_{t} + (\Sigma_{t} - \Sigma_{t+1})^{1/2} \,Z_{t},      
    \end{equation}
    where $H_{t}$ and $I_{t}$ are defined in~\eqref{eqn:hessian}
    and~\eqref{eqn:grad_cov} respectively and
    $Z_0, \ldots, Z_{T-1} \simiid N(0, I_{d})$.
  \end{lemma}
\noindent Lemma~\ref{lemma:mdp} makes use of
the reparameterization trick for Gaussian random variables; we defer derivation details to Section~\ref{section:proof-mdp}.

With the transition dynamics given in Lemma \ref{lemma:mdp}, we can formulate an MDP.
Given sampling allocations $p_{0:T} = (p_0, p_1, \ldots, p_T)$, where we recall that each $p_t$ maps contexts in $\mathcal X$ to an allocation in $\Delta_K$, we evaluate the $p_{0:T}$ by
the loss $J_T(p_{0:T})$, which is a sum of per-period objective
functions $c_t: (p_{t}, \beta_{t}, \Sigma_{t}) \mapsto \R$ that depend on the sampling allocation and the current posterior $(\beta_{t},\Sigma_{t})$ at
each epoch $t$. We also assume known per-period $b$-dimensional cost functions $g_t: (p_{t}, \beta_{t}, \Sigma_{t}) \mapsto \mathbb R_{\ge0}^b$, along with a cumulative budget $B \in R_{\ge0}^b$. The \emph{remaining budget} at time $t$ is computed by the simple transition equation 
\begin{equation}
    B_{t+1} = B_{t} - g_t(p_t, \beta_t, \Sigma_t),
    \label{eq:budget_transition}
\end{equation}
with $B_0 = B$. Since $(\beta_{t}, \Sigma_{t})$ characterizes the posterior at epoch $t$, it is sufficient
to consider policies $\pi = \{\pi_{t}\}_{t=0}^{T}$
that map the posterior state and the remaining budget to a sampling allocation $\pi_{t}: (\beta_{t}, \Sigma_{t}, B_t) \mapsto p_{t}(\cdot \,|\, x)$. We assume that for any $t = 0, 1, \ldots, T$ and any state $(\beta_{t}, \Sigma_{t}, B_t)$, there exists an action $p_t(\cdot\,|\,x)$ that satisfies the remaining budget, i.e., $g_t(p_t, \beta_t, \Sigma_t) \le B_t$.\footnote{ This condition is not difficult to satisfy in our framework. Since we allow for the zero-vector as an action, i.e., $p_t(\cdot \, | \, x) = \mathbf{0}$, one way is to ensure that $g_t$ assigns zero cost to the $p_t(\cdot \, | \, x) = \mathbf{0}$. This way, the experimenter can always take zero cost actions and maintain feasibility at any point in the experiment.} This means that a feasible policy exists.

Given the prior $\theta\opt \sim N(\beta_{0}, \Sigma_{0})$, we consider the objective of finding a policy $\pi$
that minimizes the \emph{Bayesian policy loss}:
\begin{equation*}
  \min_{\pi}
  \left\{\E_{0}[J_{T}(\pi)] \defeq
    \E_{\theta\opt \sim N(\beta_{0}, \Sigma_{0})} \left[
      \sum\nolimits_{t=0}^{T} c_{t}(\pi_{t}, \beta_{t}, \Sigma_{t}) \right]  ~\Big| ~
    \sum\nolimits_{t=0}^{T} g_{t}(\pi_{t}, \beta_t, \Sigma_t) \leq B  \right\},
\end{equation*}
where $\E_{t}[\,\cdot\,] \defeq \E[\,\cdot\,|\, \beta_{t}, \Sigma_{t}]$ denotes expectation under the posterior distribution of $\theta\opt$ at epoch $t$. 
We can observe how $J_{T}(\pi)$ generalizes more familiar notions of regret, such as Bayesian cumulative regret or Bayesian simple regret, depending on the specification of $c_{t}$.

Note the following approximation is used to formulate and optimize the planning problem, but the resulting sampling allocations are to be used in the
\emph{original} batched experiment.
\begin{definition}
  \label{def:asymptotic_bayes} 
  The \textbf{batch limit dynamic program}
  $(\model)$ is an MDP where the state is given by
  $(\beta_{t}, \Sigma_{t}, B_t)$ and actions are the sampling allocations
  $p_{t}(\cdot\,|\,x)$.  The state transition of $(\beta_{t+1}, \Sigma_{t+1}, B_{t+1})$ for
  epoch $t+1$, starting from the state $(\beta_{t}, \Sigma_{t}, B_t)$ and under a
  sampling allocation $p_{t}(\cdot\,|\,x)$ is given by \eqref{eqn:dynamics} and \eqref{eq:budget_transition}.
  The goal is to minimize Bayesian loss
  over policies $\pi_{t}: (\beta_{t},\Sigma_{t}, B_t) \mapsto p_{t}(\cdot \,|\, x)$ that are feasible (i.e., satisfy the budget constraint):
  \begin{align}
    \label{eqn:bldp-obj}
    \min_{\pi}
    \left\{ V^{\pi}_{0}(\beta_{0}, \Sigma_{0}, B_0)
    := \E_{0} \left[\sum\nolimits_{t=0}^{T}
    c_{t}(\pi_{t}, \beta_{t}, \Sigma_{t}) \right] ~\Big| ~
    \sum\nolimits_{t=0}^{T} g_{t}(\pi_{t}, \beta_{t}, \Sigma_{t}) \leq B \right\}.
  \end{align}
  Given a feasible policy $\pi$, the value function is
  $V^{\pi}_{t}(\beta_{t}, \Sigma_{t}, B_t) = \E_{t} \left[\sum\nolimits_{s=t}^{T}
    c_{s}(\pi_{s}, \beta_{s}, \Sigma_{s}) \right]$.
\end{definition}

By virtue of restricting attention to policies that only depend on the batch
posterior states~\eqref{eqn:dynamics}, the $\model$ avoids the issues that
make the dynamic program~\eqref{eqn:oae-dp-informal} intractable to solve.
\begin{enumerate}[itemsep=0pt]
\item \textbf{Obviating the need for a sophisticated unit-level Bayesian model}.  Instead of the full data
  $\mc{D}_{t} = \{ (X_{t,i},A_{t,i},R_{t,i}) \}_{i=1}^{n_{t}}$ observed in
  batch $t$, $\model$ only considers sufficient statistics
  $\what{\theta}_t$, as introduced in \eqref{eqn:batch-estimator}, modeled through posterior
  states.  Compressing the states this way removes the need to posit a
  Bayesian model for individual rewards common in the Bayesian MDP literature~\cite{GhavamzadehMaPiTa15}, which requires knowledge of the data-generating distribution of rewards $R_{t,i} \sim \nu(\cdot \,|\, x,a)$.  In
  particular, the central limit approximation~\eqref{eqn:informal-clt}
  provides a formal justification for using Gaussian likelihoods, meaning that
  Gaussianity is now a result of the CLT, not an unverifiable assumption made
  by the modeler.
\item \textbf{Dimensionality reduction of the state space}.  The states of the
  $\model$ are $(\beta_t, \Sigma_t, B_t)$, which has dimension $d + d^{2} + b$.  This
  is orders of magnitude smaller than the size of the history
  $|\mathcal{H}_{t}|$ for typical experiments.
\item \textbf{Closed-form state transitions}. Although the reward
  distribution is unknown, by making use of the asymptotic normality of the M-estimator $\what{\theta}_{t}$, state transitions in the $\model$ can be written in a certain closed form that involves i.i.d. Gaussian random variables. It is easy to simulate rollouts for even very long horizons.
\item \textbf{Differentiable state transitions}.  The effect of the sampling
  allocation on the posterior state $(\beta_{t}, \Sigma_{t})$ is mediated
  through $H_{t}$ and $I_{t}$ which are both \emph{linear} in $p_t$, smoothing
  the partial feedback. This means that state transitions are differentiable
  along any sample path, and one can compute a sample-path gradient
  $\nabla_{p_{t}} c_{s}(p_{s}, \beta_{s}, \Sigma_{s})$ for any $s>t$ with
  modern auto-differentiation frameworks (e.g., PyTorch~\cite{PyTorch19} or
  TensorFlow~\cite{TensorFlow16}) to directly optimize over sampling
  allocations.
\end{enumerate}

\subsection{Objectives and Constraints}
\label{section:objectives}

We now discuss multiple objectives and constraints that can be tractably optimized under $\model$.
Unlike  bandit algorithms focusing on particular
functional forms, our optimization-driven framework can handle flexible functions of posterior states as objectives and
constraints.

\vspace{1em}
\begin{example}[Cumulative regret]
\label{ex:cumregret}
To define \emph{cumulative regret}, we consider sampling allocations made \emph{during} the experiment (i.e. $p_{t}$ for $t<T$):
\begin{align*}
    \label{eq:cumulative_regret}
    \mathsf{BayesRegret}_{T}(p_{0:T}) 
    & = \E_0 \left[
        \sum\nolimits_{t=0}^{T-1} c_{t}(p_{t}, \beta_{t},\Sigma_{t}) \right] \nonumber \\
    & = \E_0 \left[ \sum_{t=0}^{T-1} n_{t} \, \E_t \, \E_{x\sim\mu_{t}} \left[\max_{a} \, r(x,a) -   \sum\nolimits_{a} p_{t}(a\,|\,x)\,r(x,a) \right] \right], 
\end{align*} 
where we have defined
$c_t(p_t, \beta_{t},\Sigma_{t}) = n_{t}\, \E_t \, \E_{x\sim\mu_{t}} \bigl[\max_{a} r(x,a) - \sum\nolimits_{a}
  p_{t}(a|x) \, r(x,a) \bigr]$. Recall that $\E_{t}[\,\cdot\,] \defeq \E[\,\cdot\,|\, \beta_{t}, \Sigma_{t}]$, so that the expectation is taken over the conditional distribution of $r(x,a)$ given $(\beta_{t},\Sigma_{t})$ Note that the regret of each period is weighted by the size of the batch $n_t$.
\end{example}




On the other hand,  the \emph{simple (post-experiment) regret} models a common goal in typical A/B tests where a treatment arm
is deployed to all users at the end of the experiment. In such a setting, the experimenter is concerned with the post-experiment performance of the final deployment decision $p_{T}$, which typically affects far more users than those that participated in the experiment.

\vspace{1em}
\begin{example}[Simple regret]
\label{example:simple_regret}
We define the \emph{simple
regret} as the sub-optimality of the final (non-contextual) arm allocation
$p_{T}\in \Delta_K$ compared to the arm with the highest average reward:
\begin{align}
    \mathsf{BayesSimpleRegret}_{T}(p_{0:T}) &= \E_0\left[c_T(p_T, \beta_T, \Sigma_T)\right] \nonumber \\
    &= \E_0\left[ n_{T}\left(\max_{a} \E_T \,  \E_{x\sim\mu}\bigl[r(x,a) \bigr] - \sum\nolimits_{a} p_{T}(a) \, \E_T \, \E_{x\sim\mu}\bigl[r(x,a)\bigr] \right) \right],
\end{align}
where $c_t(p_t, \beta_t, \Sigma_t) = 0$ for $t < T$ and 
\[
c_T(p_T, \beta_T, \Sigma_T) = n_{T}\left(\max_{a} \E_T \, \E_{x\sim\mu}\bigl[r(x,a) \bigr] - \sum\nolimits_{a} p_{T}(a) \, \E_T \, \E_{x\sim\mu}\bigl[r(x,a)\bigr]\right).
\]
Again, the $\E_T$ signifies that we take expectation over the conditional distribution of $r(x,a)$ under $(\beta_{T},\Sigma_{T})$. Importantly, for simple regret, $p_T$ does not depend on the context $x$.
\end{example}


Rather than identifying a single best arm to be deployed for all contexts, an
extension to settings requiring personalization gives rise to the \emph{policy regret} where we compare the final policy for assigning treatment
arms against the optimal policy that knows $\theta\opt$.

\vspace{1em}
\begin{example}[Policy regret]
In defining \emph{policy regret}, we compare the final policy for assigning treatment
arms against the optimal policy that knows $\theta\opt$:
\begin{align*}
    \mathsf{BayesPolicyRegret}_{T}(p_{0:T}) 
    &= \E_0 \left[ 
    n_{T}\left(\E_T \, \E_{x\sim\mu} \left[\max_{a} \, r(x,a) -  \sum\nolimits_{a} p_{T}(a\,|\,x) \, r(x,a) \right]\right) \right],
\end{align*}
with $c_t$ and $c_T$ appropriately defined as in previous examples.
This is closely related to the cumulative regret of Example \ref{ex:cumregret}, except that policy regret is defined in the \emph{post-experiment phase} (see Definition \ref{defn:post_exp}).
\end{example}


Our framework can handle more exotic objectives. For example, in recommendation systems, experimenters may want to select \emph{multiple} arms, for which a top-$k$-sum objective might be useful.

\vspace{1em}
\begin{example}[Top-$k$-sum regret]
The goal here is to measure the regret between the sum of the $k$ highest mean rewards and a set of $k$ arms chosen by the experimenter at the end of experiment.
\begin{align*}
  \label{eq:topk_regret} 
  &\mathsf{BayesTop}\text{-}k\text{-}\mathsf{Sum}_{T}(p_{0:t}) \\
   &= \E_{0} \left[\E_{T} \,\E_{x\sim\mu} \left [\text{top-}k\text{-sum}\bigl[(r(x, a))_{a \in \mathcal A} \bigr]  
    - \sum\nolimits_{A_{k}} p_{T}(A_{k} \,|\, x ) \, \sum\nolimits_{a \in A_{k}} \!\! r(x,a)  \right] \right],
\end{align*}
where for any vector $v \in \R^{K}$,
$\mathop{\text{top-}k\text{-sum}}(v) = \max \{ v^{\top} q : 1^{\top}q = k, \, q\geq 0\}$
takes the sum of the top $k$ elements and
$p_T$ is modified to assign probability to \emph{$k$-subsets of actions} $A_{k}$ (rather than actions themselves). For $k=1$, this is identical to Bayesian simple regret.
\end{example}

\vspace{1em}
\begin{example}[Arm-cost budget constraints]
When each treatment arm has an associated cost (e.g., an e-commerce platform experiments with different levels of promotions), we may want to specify an overall budget for the experiment. This can be achieved by setting $g_t(p_t, \beta_t, \Sigma_t) = q^{\top}p_{t}$, where $q$ is a vector of arm costs.
\end{example}

\vspace{0.5em}
\begin{example}[Sample coverage guarantees]
In some cases, we may want to guarantee that all arms are sampled with some minimum probability, i.e., $p_t(a \,|\, x) \ge 0.1$. This can accomplished in our framework by establishing $T \cdot K$ constraints via appropriate definitions of $g_t$ and $B$.
\end{example}
\section{Optimization Algorithm ($\algo$)}
\label{section:rho}

\begin{algorithm}[t]
    \caption{\label{alg:rho} $\algofull$}
    \begin{algorithmic}[1]
      \State Input: Prior $(\beta_0, \Sigma_0)$, objectives $c_{s}$, constraints $(g_{t}, B)$, batch context distributions $\mu_{0},\ldots,\mu_{T-1}$
      \State Initialize $(\beta_0, \Sigma_0), B_{0} = B$.
      \For{each epoch $t \in 0, \ldots, T-1$}
      \State Solve the problem (e.g., using SGD) with sample-paths $(\beta_{s}, \Sigma_{s})$ drawn from the $\model$.
      \begin{equation}
        \label{eqn:rho}
        \small
        p_{t:T}\opt \in
        \argmin_{ p_{t}, \ldots, p_{T}} 
        \left\{ V^{p_{t:T}}_{t}(\beta_{t},\Sigma_{t}, B_{t})
          \defeq \E_{t}  \left[
            \sum\nolimits_{s=t}^{T} c_{s}(p_{s} \, | \, \beta_{s}, \Sigma_{s})
          \right]
          ~~\Big|~~   \sum\nolimits_{s=t}^{T}g_{s}(p_{s}, \beta_s, \Sigma_s)  \leq B_{t}
            \right\}
      \end{equation}
      \State Sample arms $A_{t,i} \sim p^{*}_{t}(\cdot \,|\,X_{t,i})$. Observe rewards $R_{t,i}$.
      \State Update state $(\beta_{t+1}, \Sigma_{t+1})$
      according to the transitions~\eqref{eqn:posterior_update}.
      \State Update constraint $B_{t+1} = B_{t} - g_{t}(p^{*}_{t}, \beta_t, \Sigma_t)$.
      \EndFor 
      \State Output: Final posterior state $(\beta_{T}, \Sigma_T)$ and final allocation $p^{*}_{T} \in \argmin_{p_{T}} c_{s}(p_{T} \,|\, \beta_{T}, \Sigma_{T})$.
    \end{algorithmic}
  \end{algorithm}


For small horizon adaptive experiments ($T \le 10$), the $\modelfull$ gives a tractable dynamic program that can guide
experimentation in batched settings. With this in hand, one can apply methods
from approximate dynamic programming and reinforcement learning to develop
adaptive designs to optimize performance in this model.  While the
$\model$ is far more tractable than the original DP
formulation described in Section~\ref{section:oae-dp}, it still has some of the inherent
difficulties of solving dynamic programs, especially since it is a continuous
state and action dynamic program.  But fully solving the dynamic program may
not be necessary.  \citet{che2023adaptive} notes even a simple algorithm based
on model-predictive control (MPC) can achieve large improvements compared to
standard adaptive policies---including sophisticated RL algorithms.
By crystallizing the algorithm design principle behind this method, 
we extend this simple yet robust heuristic to the substantially more general formulation $\model$.

The key capability we leverage from the $\model$ is the ability to plan over
 short time horizons $T$. Instead of solving for an optimal dynamic policy,
$\algo$ instead plans a sequence of static sampling allocations $\rho_{s}$ to
be used for future epochs. It does so by solving an open-loop \emph{planning
  problem}, following the MPC design principle: \vspace{1em}

\noindent\makebox{
    \parbox{\dimexpr\linewidth-2\fboxsep-2\fboxrule\relax}
    {\centering
    {\bf MPC design principle}: At every epoch $t$, given the current state $(\beta_{t}, \Sigma_{t}, B_t)$,\\
    solve for the optimal \emph{static} sampling allocations $(\rho_{t},\ldots,\rho_{T-1})$.
  }
}
\vspace{0.2em} 

\begin{wrapfigure}{Y}{0.4\textwidth}
  \vspace{-.3cm}  
    \includegraphics[height=1.8in, width=2.3in]{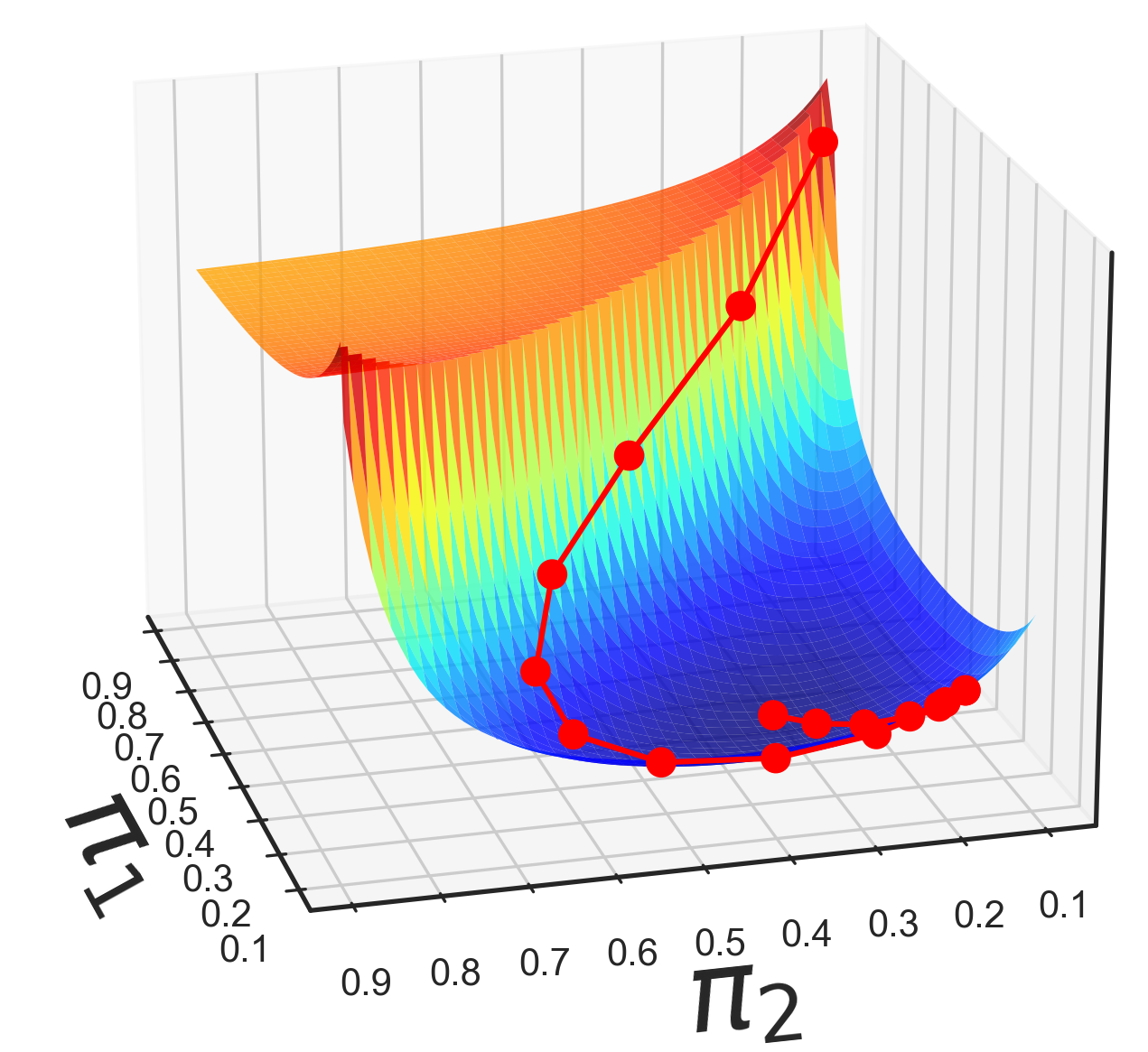}
    \caption{$\model$ provides a differentiable MDP that can be minimized via
      stochastic gradient descent. For a 3-armed experiment with horizon
      $T = 10$, we plot simple regret (red) over static sampling allocations
      $\pi(\mc{H}):= p = (p_{1}, p_{2}, 1-p_{1}-p_{2})$ optimized using the Adam
      optimizer.}
      \vspace{-.3cm}
  \label{fig:sgd}
\end{wrapfigure}
At any epoch, this leads to an optimization problem over static (constant)
sampling allocations that only depend on the previous data through the current
posterior state (Algorithm~\ref{alg:rho}).  The current sampling allocation
gets deployed and additional data is observed; by updating the posterior
beliefs using the Gaussian approximation of the estimators $\hat{\theta}_{t}$
computed in each batch~\eqref{eqn:dynamics}, we can flexibly re-solve the
optimization problem whenever new data is obtained.  Despite its simplicity,
we demonstrate in the following that the MPC design principle can reap the
benefits of adaptivity. In particular, this method can be seen as a natural
dynamic extension of a static A/B test.

The planning problem~\eqref{eqn:rho} over static sampling allocations is
computationally efficient to solve, compared to its counterparts involving
dynamic allocations that depend on future observations.  In fact, we can
leverage the differentiability of the state transitions of the asymptotic
model~\eqref{eqn:dynamics}, and solve this optimization
problem~\eqref{eqn:rho} directly with stochastic gradient descent methods,
rather than any dynamic programming. As we illustrate in Figure~\ref{fig:sgd}, we are nevertheless still solving a non-convex stochastic program and do not expect formal guarantees. 
Given the AI community's success in heuristically solving such problems
in the past decade, we believe standard software tools in ML (e.g., PyTorch and Weights \& Biases) can provide good adaptive designs in practice. We validate this claim empirically on a wealth of examples in the next section.

\paragraph{Robustness Guarantee} 
A key advantages of $\algo$ over other alternatives is that it flexibly
handles custom objectives, constraints, and non-stationarity, while
guaranteeing performance in all of these myriad settings.  This is concretely
realized by the following theorem, which shows that it always achieves a lower
Bayesian regret compared to a static A/B test across all time horizons
$T$---even across these custom objectives, constraints, and non-stationarity
arrivals.
\begin{theorem}[Policy Improvement]
\label{theorem:policy_improvement} Suppose that for any $t$, $(\beta_{t}, \Sigma_{t})$, and $B$, there exists a static sequence $p_{t},...,p_{T}$ such that $\sum\nolimits_{s=t}^{T} g_{s}(p_{s} ) \leq B$ is feasible.
Consider any static sequence $p=(p_{0},\ldots,p_{T})$
of feasible sampling allocation policies.
Let $V_{t}^{p_{0:T}}$
be the corresponding value function under the Bayesian model and $V_{t}^{\algo}(\beta_{t},\Sigma_{t})$
be the value function of $\algo$. For all $t,\beta_{t},\Sigma_{t}$ and budget $B_{t}$,
\[
    V_{t}^{\algo}(\beta_{t},\Sigma_{t}) \leq \min_{p_{0:T}}
    \left\{
    V_{t}^{p_{0:T}}(\beta_{t},\Sigma_{t})
    ~~\Big|~~
   \sum\nolimits_{s=t}^{T} g_{s}(p_{s}) \leq B_{t}
    \right\}.
\]
\end{theorem}
\noindent The proof is in Appendix~\ref{section:proof-policy-improvement}.

Given that a static A/B test is often the de-facto choice of sampling
allocation, this result guarantees that $\algo$ will achieve the baseline
level of performance, even in settings where standard heuristics such as
Thompson Sampling cannot be applied without modification. In contrast, to
apply standard heuristics to new settings requires ad-hoc changes to the
policy without any guarantees that the modified sampling principle will work
appropriately.

This guarantee is even more crucial under non-stationarity. In non-stationary
settings adaptive algorithms can often perform much worse than static designs as we concretely demonstrate in the next section,
as adaptive policies can overfit towards early but temporary performance. As
long as the non-stationarity is \emph{predictable}, that is the batch context
distributions $\mu_{s}$ are known ahead of time, $\algo$ is guaranteed to do
at least as well as the best static design. While this may seem like a strong
requirement, this is naturally satisfied when modeling temporal
non-stationarity (e.g. day-of-the-week effects or novelty effects).  Temporal
non-stationarity can be modeled by having the context be the epoch $x = t$; in
this case future context arrival distributions are $\mu_{s} = 1\{x = s\}$ and
are deterministic.  In more complicated settings, e.g. if the context are
highly user-specific features, we can instead use samples from the context
distributions $\mu_{s}$, e.g., collected through a historical dataset.

Beyond guranteeing performance against static designs, $\algo$ can match performance and even outperform
other adaptive policies on standard instances by planning with the Bayesian model. In doing so, it 
calibrates exploration to the time horizon and the signal-to-noise ratio.
It uses more information about the experimental setting to achieve this, such as the horizon length $T$, which is often known in advance. $\algo$ also requires the ability to estimate the hessian $H_{t}$ and gradient covariance $I_{t}$.
This is straightforward for standard statistical settings, such as least-squares or logistic regression.
Using this additional information, it does sensible things. Early in the experiment,
when the remaining sample size is large, it explores widely. 
Towards the end of the experiment, it focuses on a few, promising treatment arms.
Ultimately though, we validate the algorithmic design through evaluation in realistic experimental environments.


We emphasize that $\algo$ is a simple heuristic and 
that the goal of this paper is not to propose the most optimal or theoretically grounded algorithm. As 
an example of its limitations, $\algo$ converges to a greedy policy when optimizing only for 
cumulative regret. While greedy policies perform well in short horizon settings, they 
can perform poorly in longer experiments.   
Our approach serves as an initial proof-of-concept for our mathematical programming framework, and its
strong empirical performance provides promising validation for our current framework.  
We hope to spur further work on scalable solution methods and algorithm designs.


\section{Numerical Experiments}
\label{section:experiments}

Using the simple heuristic $\algo$, we illustrate the potential of our mathematical programming framework across 
multiple problem instances.
 We examine two important challenges that are highly relevant to real-world experimentation applications: 
\emph{non-stationarity} and \emph{personalization}. Both can pose difficulties for standard adaptive experimentation algorithms. For the former, we make use of the ASOS Digital Experiments Dataset, introduced in~\cite{liu2021datasets}, which contains 78 real experiments run at ASOS, a global fashion retailer with over 26 million active customers (as of 2024). For the latter, we create a synthetic environment for personalized content ranking for online platforms, inspired by the content recommendation systems of Instagram and Facebook \cite{mahapatra2020insta, lada2021fb}. 

For both environments, we compare the performance of $\algo$ against three baseline methods: Uniform allocation, Thompson sampling \cite{agrawal13thompson, RussoVaKaOsWe18}, and Top-two Thompson sampling \cite{Russo20}. We select these baseline methods for their widespread adoption, state-of-the-art performance \citep{chapelle2011empirical}, and importantly, the ability to naturally handle the batched setting.\footnote{Many bandit algorithms, such as UCB, do not easily extend to the batched setting. For example, in a non-personalized setting, UCB would assign the same action to all individuals. Traditional Bayesian optimization approaches \citep{frazier2018tutorial} face a similar difficulty.} All of the baseline methods maintain a Gaussian belief. We provide more details about the baselines below.
\begin{itemize}
    \item \textbf{Uniform.} A non-adaptive policy that uniformly allocates samples across all arms.  
    \item \textbf{Thompson sampling (TS).} The main idea behind Thompson sampling is to sample from the posterior distribution of the optimal action \citep{RussoVa16}. To implement this, at time $t$, TS first samples a parameter $\hat{\beta}_{t}$ from a prior belief $\hat{\beta}_{t} \sim 
    N(\beta_{t}, \Sigma_{t})$. Given a reward model\footnote{Recall our notation from \eqref{eqn:parametric reward model} that $r(x,a) = f(x,a;\theta^*)$.} $f(x,a;\theta)$ and context $X_{t,i}$, TS then makes an allocation decision by choosing $A_{t,i} = \argmax_{a} \, f(X_{t,i}, a; \hat{\beta}_{t})$. See \cite{agrawal13thompson, RussoVaKaOsWe18} for additional details. As mentioned above, although TS is typically presented in the fully-sequential setting, we can naturally adapt TS to our batch setting by simply repeating the sampling procedure for each unit in the batch.
    \item \textbf{Top-two Thompson sampling (TTTS).} Top-two Thompson sampling, introduced in \cite{Russo20}, first identifies $A_{t,i}$ via the same procedure as TS. With probability $\beta \in (0,1)$, TTTS discards $A_{t,i}$ and repeatedly samples parameters $\hat{\beta}_{t} \sim N(\beta_{t}, \Sigma_{t})$ until the optimal action $\argmax_{a} f(X_{t,i}, a; \hat{\beta}_{t}) \neq A_{t,i}$. This encourages the algorithm to explore outside of the best arm and is specifically designed to minimize simple regret.
\end{itemize}

\subsection{Non-stationarity via the ASOS Digital Experiments Dataset}
 Our first empirical study focuses on the issue of non-stationarity and makes use of the 78 real experiments included in the ASOS Digital Experiments Dataset, introduced in~\cite{liu2021datasets}. Each experiment contains one treatment and one control variant and includes up to four different metrics (i.e., outcomes of interest), which results in 241 unique benchmark settings.  Associated with each metric is sequence of sample means and variances, recorded at either 12-hour or daily intervals. The length of the experiments range from 2 to 132 recorded intervals. 


We construct a simulation environment using the ASOS dataset as follows. To model non-stationarity, we view the context $X_{t,i}$ to be time; i.e., we have that $X_{t, i} \equiv t$. We take the conditional mean rewards to be the sample means provided in the dataset, while individual rewards are assumed to be sampled from a Gaussian distribution parameterized by the sample means and sample variances from the dataset. Since each experiment in the dataset contains only two arms (treatment and control), we also generated \emph{synthetic arms} to mimic multi-armed experimentation problems that arise in practice. The means of the synthetic arms are set to be the sample mean of the control arm shifted by a $N(0,1)$ noise term multiplied by the gap between the sample means of treatment and control arms. The variances of the synthetic arms are set to be identical to the sample variance of the treatment arm. In doing so, we generate arms with similar average rewards to the treatment arm and display similar non-stationarity in the means across epochs.

 We construct our simulation with 10 experimentation epochs consisting of 10 arms and three batch sizes: 
 10,000, 100,000, and 250,000 observations per epoch. We assume that the experimenter starts with a prior $N(\beta_0, \Sigma_0)$ where the structure of $\beta_0$ and $\Sigma_0$ depend on the assumed model.  The objective is to minimize Bayesian simple regret (Example ~\ref{example:simple_regret}). In this nonstationary setting, we set the population distribution $\mu$ to be uniform over $[T]$, resulting in
\begin{equation*}
    \label{eq:simple_regret_asos}
    \mathsf{BayesSimpleRegret}_{T}(p_{0:T}) =  \E_{0}\left[\E_{T}\Biggl[\max_{a} \sum\nolimits_{t=1}^T r(t,a) \Biggl] - \sum\nolimits_{a} p_{T}(a) \, \E_{T}\Biggl[\sum\nolimits_{t=1}^T r(t,a)\Biggr]\right],
\end{equation*}
 
 \noindent \newline \textbf{Reward models.} Based on the temporal variation within the ASOS dataset and the simple regret objective, we first introduce two models that form the basis of each experimentation policy. 
\begin{itemize}
    \item  \textit{Non-contextual.} Here, the parameter vector is given by $\theta = (\theta_1, \ldots, \theta_{K}) \in \mathbb{R}^K$, where we learn one parameter per arm. We then set $\phi(t,a)$ so that
 \[f(t,a;\theta) = \phi(t,a)^\top \theta = \theta_{a},\] 
 with arm effects being constant across time. 
 

    \item \textit{Contextual (interactive effects).} The parameter vector is given by
    \[
    \theta = (\underbrace{\theta_1, \ldots, \theta_K}_{\text{arm constants}}, \underbrace{\theta_{1,1}, \ldots, \theta_{1,K}}_{\text{time 1}}, \ldots, \underbrace{\theta_{T,1}, \ldots, \theta_{T,K}}_{\text{time } T}),
    \]
    and the feature map $\phi(t,a)$ is set so that
    rewards are modeled as   
\[f(t,a;\theta) = \phi(t,a)^\top \theta = \theta_a + \theta_{t,a}.\] 
In contrast to the non-contextual model, arm effects can now vary across time, better resembling the true data generating model.
\end{itemize}

\noindent \textbf{Implementation details.} We implement Uniform, TS, TTTS, and $\algo$ under both reward model specifications above.
Note that TS and TTTS under the interactive effects model is exactly Deconfounded TS \cite{QinRusso2023}, an extension of TS designed specifically to handle non-stationary environments. 
To initialize the prior for all policies, each policy starts out with a uniform prior $N(0, \lambda I)$ where $\lambda$ scales inversely 
proportional to the batch size $b_tn$. 
The nature of these time dependent models implies that the minimum eigenvalue of the sample covariance 
matrices $\lambda_{\text{min}} (\Phi_t^\top \Phi_t) = 0$ at each batch. Since the Bayesian posterior updates 
in~\eqref{eqn:approx_posterior} require invertibility of $\Sigma_t$, we choose 
to scale $\lambda$ with the batch size to maintain invertibility 
and stable condition numbers as the experiment progresses. 
Empirically, we find that this scaling is necessary to maintain stable performance across all policies. 

The values in the ASOS dataset are reported in a cumulative fashion, but it is ambiguous why the data were reported in this way. We interpret the data by making the assumption that users are observed once at each time step (e.g. daily active user metric) rather than cumulatively (e.g. total clicks over a time period). We preprocess the data by subtracting off the values of the previous time step.

\subsubsection{Results and Discussion for the ASOS Dataset}
We run 750 experiment simulations for each of the 241 settings in the ASOS dataset, resulting in 180,750 evaluations of each policy and batch size. Surprisingly, most of adaptive experimentation algorithms benchmarked on the different settings within the ASOS dataset offer little to no gains over uniform allocation. 

\noindent \newline \textbf{Performance comparison between TS-based policies versus $\algo$.} While TS-based policies offer strong theoretical guarantees in ideal settings, Figure \ref{fig:main_plot} shows that under a batch size of $100,000$, all TS-based policies actually perform worse than uniform in both average reward obtained and identifying the best arm. In contrast, $\algo$ offers strong performance. In fact, for \emph{every hyperparameter setting} (i.e., the choice of learning rate, $\algo$'s only hyperparameter), $\algo$ outperforms TS-based policies even when choosing between the best of TTTS and the greedier TS in each individual setting. This type of robustness against hyperparameters is highly desirable in practice. We observe the same trends across batch sizes of $10,000$ and $250,000$, where $\algo$ is the only policy to outperform Uniform. All policies have lower raw regret values as batch sizes increase, showing that while all policies perform better with more information, $\algo$ is particularly robust to various amounts of statistical power.  

Sometimes, an experimenter may have validation data to tune RHO's learning rate in individual settings. We simulate this possibility and find that choosing from a sweep over five different learning rates in each setting yields substantial performance gains. Overall, we find that RHO is robust to different learning rate choices, 
but can also offer significant gains when tuned properly.

    

\noindent \newline \textbf{Failure of existing adaptive experimentation policies.} To understand why existing adaptive experimentation policies like TS or TTTS fail to offer significant gains over uniform allocation, we note that the settings within the ASOS dataset are often incredibly challenging with severe non-stationarity and a low signal-to-noise ratio. As previously mentioned, Figure~\ref{fig:diff-in-means} shows an example of a setting with non-stationary means. Under non-stationary behavior, adaptive algorithms may fail to explore, allocating samples to arms that show initial promise but eventually underperform the experiment progresses. In addition, the average gap between arms in the ASOS dataset is 0.0093 while the average measurement variance is 33.76, ranging from 0.0007 to over 2870. As these  data are from real experiments, we believe that our results serve not just to advocate for the validity of RHO as a method, but also sheds light on the robustness (or lack thereof) of standard adaptive experimentation algorithms in challenging, non-ideal, but realistic conditions. 


\begin{table}[t]
\centering
\small
\renewcommand{\arraystretch}{1.2}
\begin{tabular}{ccccc}
\toprule
& \multicolumn{2}{c}{Simple Regret $<$ Uniform} &  \multicolumn{1}{c}{Simple Regret $>$ Uniform} \\
Policy & \% Settings & \% Avg Regret of Uniform & \% Avg Regret of Uniform \\
\cmidrule(lr){2-3} \cmidrule(lr){4-4}
Non-contextual TTTS~\cite{Russo20}   & 39.1  & 92.1 & 124.8 \\
Contextual TTTS~\cite{QinRusso2023}   &  51.8 & 89.7 & 110.7 \\
Contextual RHO  & 60.5  & 90.4 & 106.9 \\
\bottomrule
\end{tabular}
\caption{While TS and TTTS policies can achieve lower average regret on settings where they do better than Uniform, $\algo$ is much more stable in situations where it underperforms Uniform. $\algo$ achieves lower regret than Uniform in 445/732 settings (60.5\%), more than Non-contextual TTTS (39.1\%) and Deconfounded (Contextual) TTTS (51.8\%).}
\label{100000_batch}
\end{table}

\noindent \newline \textbf{Understanding the performance of $\algo$.} We observe that the main performance gains of $\algo$ come from its consistency and robustness in challenging scenarios where adaptive algorithms underperform Uniform. In Table \ref{100000_batch}, we examine the two events of underperforming and outperforming Uniform separately. We find that while RHO obtains slightly higher regret compared to TS when both are \emph{conditioned on} outperforming Uniform, overall $\algo$ performs better than Uniform in many more settings (60.5\% vs 51.8\% and 39.1\%). 
TS policies also have heavier regret tails, doing much worse in challenging situations compared to RHO. Similar behavior is detailed in Appendix \ref{section:more-experiments} is found in settings with $10,000$ and $250,000$ batch sizes. 

\noindent  \newline  \textbf{Model misspecification.} Even when the model is misspecified, $\algo$ can do better than other similarly misspecified models. When comparing policies under the non-contextual model (which cannot explain the variations due to interdependent arm and day effects), Table \ref{Misspecified Table} shows that $\algo$ still outperforms Uniform while TS and TTTS can suffer large drops in performance.  

\begin{table}[t]
\small
\centering
\renewcommand{\arraystretch}{1.2}
\begin{tabular}{cccc}
\toprule
& \multicolumn{3}{c}{Batch Size} \\
\cmidrule(lr){2-4}
Policy & 10,000 & 100,000 & 250,000 \\
\midrule
Non-contextual TS   & 0.0006  & -1.01 & -0.004 \\
Non-contextual TTTS   &  0.003 & -1.02 & -0.003 \\
Non-contextual RHO  & 0.099  & 0.003 & 0.004 \\
\bottomrule
\end{tabular}
\caption{Difference in raw regret values of methods under misspecified (non-contextual) models compared to Uniform with a correctly specified model. Non-contextual $\algo$ is able to consistently outperform Uniform under model misspecification, while TS-based policies struggle. }
\label{Misspecified Table}
\end{table}

\begin{figure}[t]
  \centering
  \includegraphics[width=0.7\textwidth]{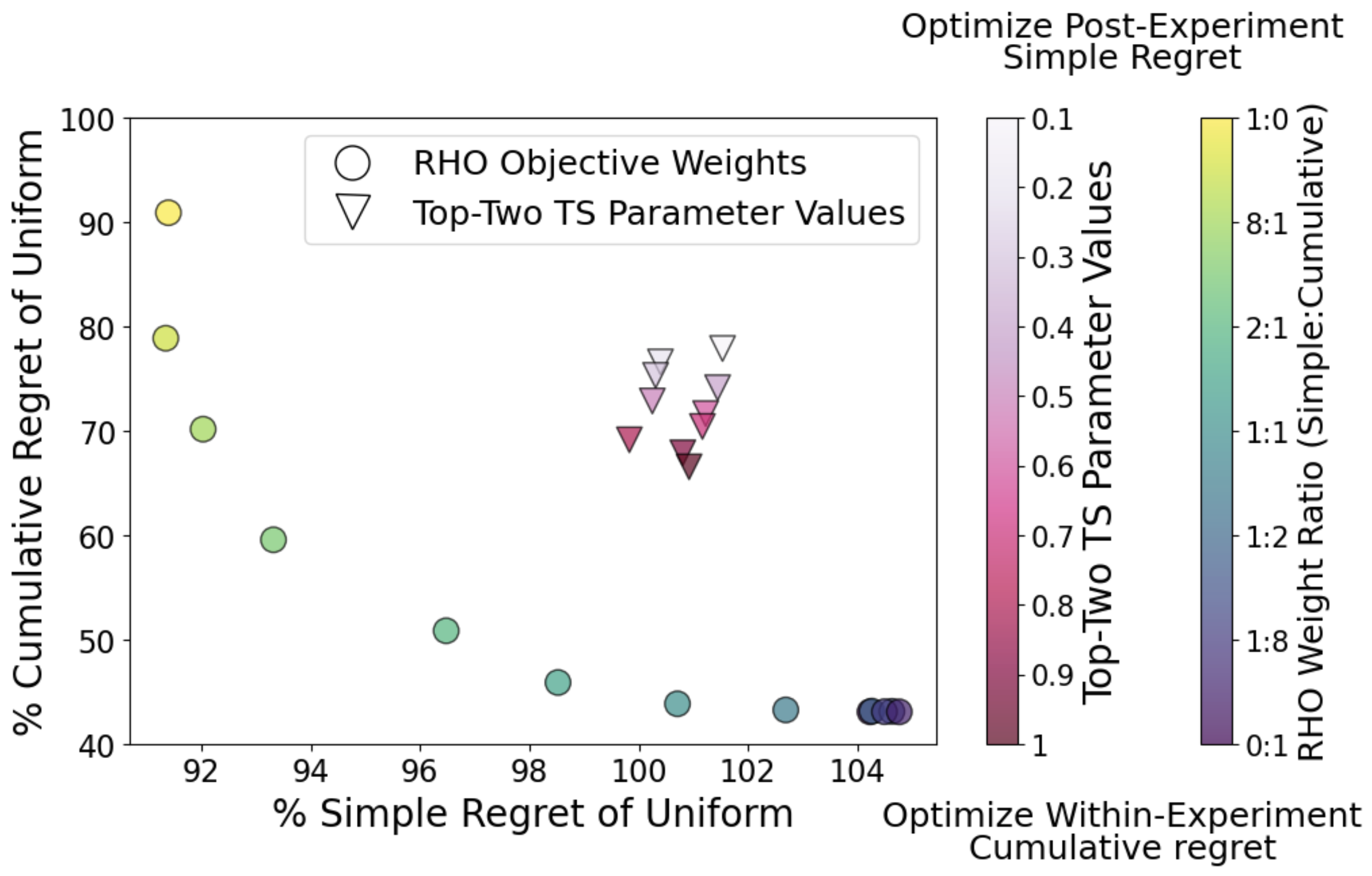}
  \caption{\textbf{(Personalized Value Models)} Pareto frontier of the tradeoff between simple and cumulative regret with $5$ epochs of experimentation in a synthetic personalized ranking simulation. Lighter colors corresponding to 
  TTTS indicate parameter values closer to $0$, while darker values correspond to values closer to $1$. Note that 
  a value of $1$ is exactly TS. Similarly, lighter values for $\algo$ correspond to higher weights on simple regret
  while darker values correspond to higher weights on cumulative regret. $\algo$ is able to efficiently trade off 
  between the two objectives, with many points strictly dominating all values associated with TTTS. As more weight is put
  on the simple regret term, the simple regret incurred decreases monotonically.
  $\algo$ trades off in a principled and interpretable way simply by setting the weights equal to the number of individuals under 
  each objective. On the other hand, there is no principled way of selecting the parameter values for TTTS, and most
  do not outperform Uniform in terms of simple regret. 
  }
  \label{fig:pareto-ranking}
\end{figure}

\subsection{Personalized Value Models for Ranking and Recommendations}
\label{subsection:personalization}

We further benchmark and explore the performance of $\algo$ in a synthetic personalization setting. 
Classical works in personalization with contextual bandits focus on settings that involve sequentially 
choosing individual pieces of content to recommend to users \cite{li2010contextual, ChapelleLi11, agrawal13thompson}, 
with the goal of minimizing cumulative regret as content is recommended. 
Let $x_{i} \in \mathbb{R}^d$ be the features corresponding to user $i \in \{1, \ldots, N\}$.  
The basic linear model used in \cite{li2010contextual} involves a separate parameter $\theta_{a} \in \mathbb{R}^d$ 
for each piece of content (arm) $a$ such that $r(x_{i}, a_{i}) = x_{i}^\top \theta_{a}^{*}$.
Content is sequentially recommended to various users based on beliefs about the parameter $\theta_{a}^{*}$. Note that there is a separate parameter $\theta_{a}^{*}$ to be learned for each action (piece of content). 

However, as platforms like Instagram and Facebook scale significantly in size and recommend 
content to billions of users, sequentially recommending and estimating the value 
of individual pieces of content may be infeasible. In addition, learning separate coefficients $\theta_{a}$ for each piece of content like the traditional contextual bandit setting is impractical. Therefore, these platforms use a \emph{user value model}, a linear function that scalarizes a set of machine learning signals into a single value for each piece of content \cite{mahapatra2020insta, lada2021fb}. 
Define user features $x$ and content features $z$. The experimenter has a 
known feature map $\phi(x,z) \in \mathbb{R}^d$ that generates $d$ ranking signals. 
Given weights $w_{x}$ for a user $x$, the ``value'' (or ``score'') assigned to content $z$ would be modeled as $w_{x}^\top \phi(x,z)$. For each piece of content $z$ in the universe of content $\mc{Z}$, this value can be calculated and content can be ranked based on this model. 
Instead of choosing each individual piece of content to recommend to users, 
each experimentation epoch involves choosing ranking weights $w_{x}$ to deploy 
for each user $x$ such that each user sees $b$ pieces of content ranked by $w_{x}$. 

\noindent \newline \textbf{Ranking simulation setup.} There are $T$ epochs of experimentation with $n_{t}$ users at each epoch $t$. 
Each user has features $x_{i} \in \R^{d}$ 
and there is a selection of $|\mathcal Z|$ pieces of content that can be recommended to each user. At each epoch $t$, the experimenter can recommend $b$ pieces of content to each user. 
User features $x$ and content features $z \in \R^{d}$ are simulated from Gaussian distributions: 
\[
x \sim N(\mu_{\text{user}}, \Sigma_{\text{user}}), \quad z \sim N(\mu_{\text{content}}, \Sigma_{\text{content}}).
\]
We let $\phi(x,z) := x \odot z$, where $\odot$ denotes the element-wise product. 
Given a true parameter $\theta^{*}$, we let the expected reward of recommending content $z$ to user $x$ be 
\begin{equation}
\label{eq:ranking-model}
    r_{\mathcal Z}(x, z) = f_{\mathcal Z}(x, z; \theta^{*}) = (x \odot z)^\top \theta^{*} = \langle x, z \rangle _{\theta^{*}},
\end{equation}
where $\langle x, z \rangle _{\theta^{*}}$ represents a $\theta^{*}$-weighted inner product between $x$ and $z$ and the subscript $\mathcal Z$ indicates that this is the \emph{user-content} reward function (later, we will define the primary reward function, which depends on $x$ and $w$). The individual rewards are distributed according to $N(\langle x, z \rangle _{\theta^{*}}, s^2)$ for some fixed measurement variance $s^2$.

The action space consists of $K$ predefined rankers $\{w^{(1)}, \ldots, w^{(K)}\}$, with $w^{(k)} \in \mathbb{R}^d$, that the experimenter must choose between. For each user $x$, the experimenter may choose a personalized ranker.
Choosing ranker $w$ for user $x$ chooses a set of items $\mathcal Z(x, w)$ to be recommended, defined to be the top $b$ items when sorted by $\langle x, z \rangle _{w}$. Let  $r(x, w)$ be the total reward over the user-content rewards for the top $b$ items
\begin{equation}
  \label{eq:rank-content-set}
  \textstyle    r(x, w) = f(x, w; \theta^*) = \sum_{z \in \mathcal Z(x, w)} f_{\mathcal Z}(x, z;\theta^*)
~~\mbox{where}~~
  \mathcal Z(x, w) := \text{top-$b$}_{z \in \mc{Z}} \; \langle x, z \rangle _{w}.
\end{equation}


\noindent \newline \textbf{Objectives.} 
For the ranking experiment, we consider a range of objectives, where the experimenter is interested in balancing cumulative regret and simple regret (or maximizing the cumulative and final rewards).
The tension between these two objectives is a recurring dilemma that experimentation practitioners often face. One way to view this trade-off is to consider the direct impact on users: during the experiment phase the experimenter may have to experiment on $\sum_{t=0}^{T-1} n_{t}$ users, while the final policy will be deployed on $n_{T}$ users. The experimenter will ideally minimize the cumulative regret incurred within the experiment (i.e., not recommending low quality content) while finding a good policy to deploy after the experiment. If $n_{T} \gg \sum_{t=0}^{T-1} n_{t}$, then the experimenter should focus primarily on minimizing simple regret, while if $\sum_{t=0}^{T-1} n_{t} \gg n_T$, then the experimenter should minimize cumulative regret.

\noindent \newline \textbf{Implementation details.} Given a context $X_{t,i}$, TS chooses an action by sampling $\hat{\beta}_{t} \sim N(\beta_{t}, \Sigma_{t})$ and then selecting $A_{t,i} = \argmax_{w} f(X_{t,i}, w; \hat{\beta}_{t})$. $\algo$ directly optimizes the simple and cumulative regret objective, where the tradeoff between these two terms is automatically optimized based on the number of individuals assigned to each objective $\sum_{t=0}^{T-1} n_{t}$ and $n_{T}$.

\subsubsection{Results and Discussion for Personalized Value Models}
In this section, we present several experiments based on the personalized recommendations setting. We examine each algorithm's ability to (1) balance simple and cumulative regret and (2) optimize simple regret only. Lastly, we also present a more traditional, single-item recommendation system formulation (i.e., without ranking or value models) that is closer to the idealized contextual bandit setting under which TS-based policies are designed.


\noindent \newline \textbf{Balancing simple and cumulative regret.} We first benchmark the algorithms' ability 
to efficiently trade off between simple and cumulative regret. We fix the within-experiment phase at $T=5$ waves of 
$n_t = 100$ users for $t < T$. There are $K = 10$ different rankers to choose from that recommend $b = 4$ items out of $10$ for each 
user. To trade off between the two objectives, we vary the parameter of TTTS from $0.1$ to $1$ 
in increments of $0.1$ as it controls the level of exploration. 
A parameter of $1.0$ exactly corresponds to standard TS, which behaves more greedily \cite{Russo20}. To vary the level of exploration of $\algo$, we vary $n_{T}$, which corresponds to the number of users the final policy will be deployed upon. 

\begin{figure}[t]

  \centering
  \hspace{-.9cm}
    \subfloat[\centering Simple Regret Performance]{\label{fig:simple-regret-ranking}{\includegraphics[height=6cm]{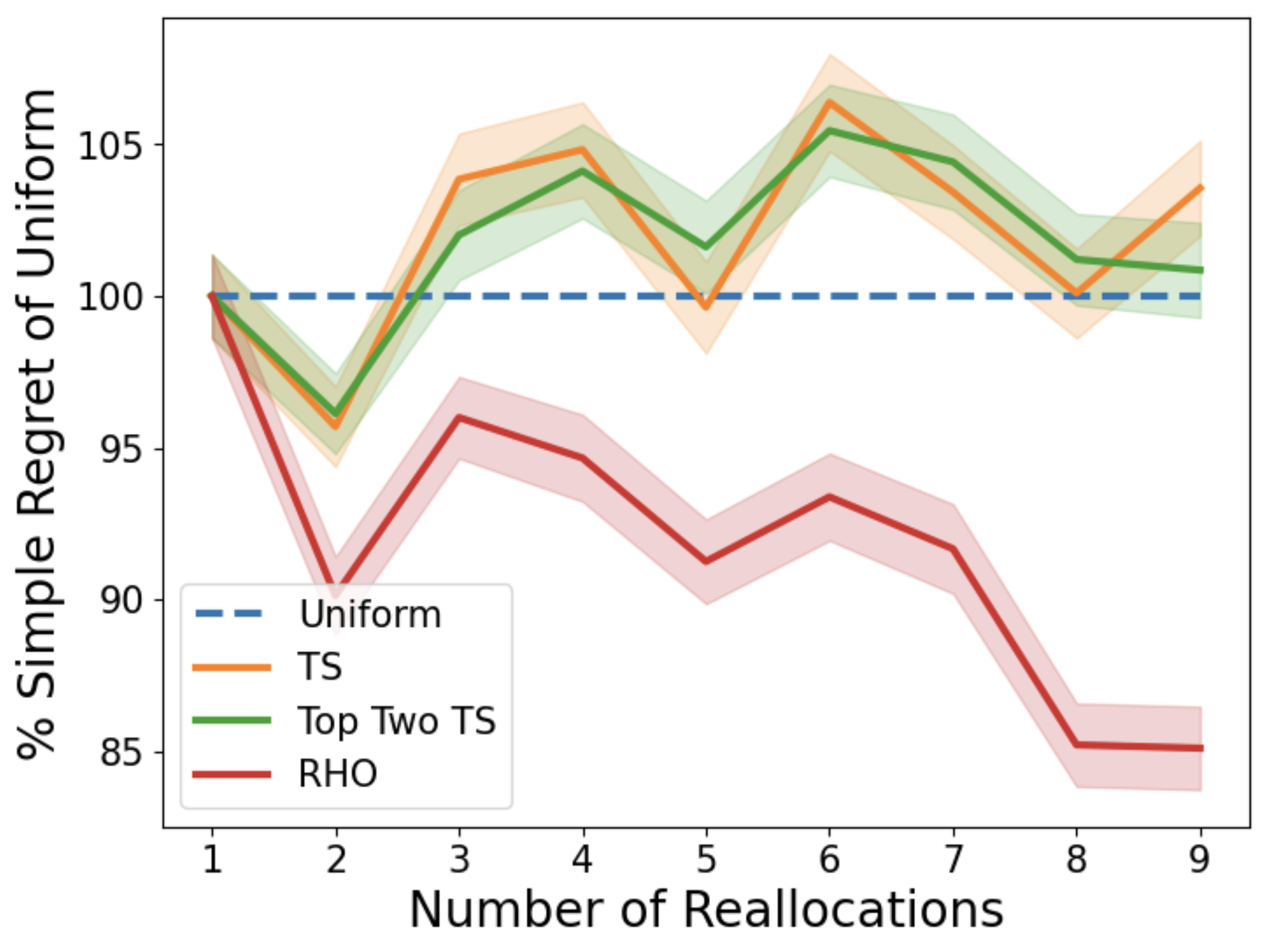}} }%
    \hspace{.3cm}
    \subfloat[\centering Cumulative Regret Performance When $\algo$ Optimizes for Simple Regret]{\label{fig:cumul-regret-ranking}{\includegraphics[height=6cm]{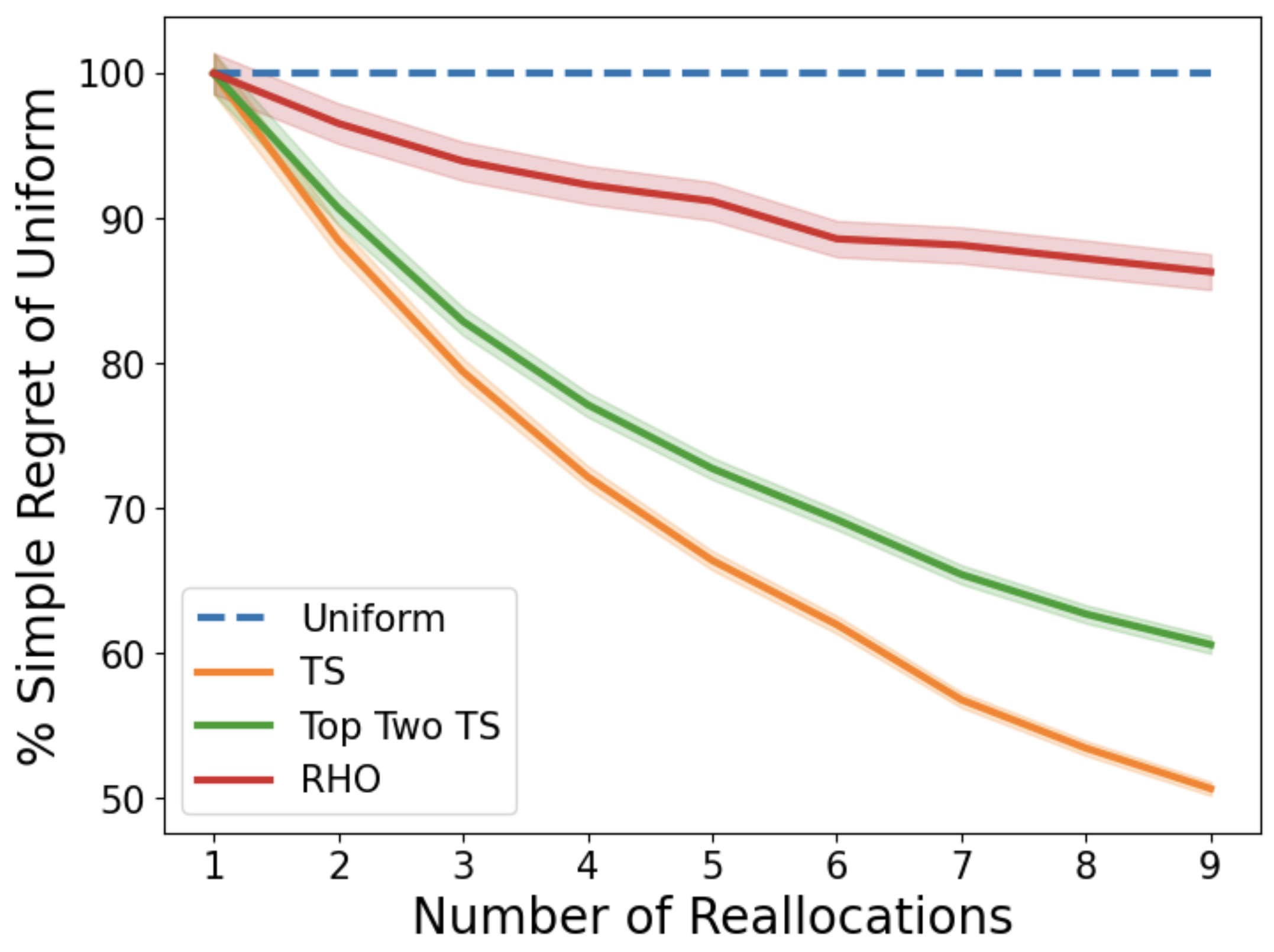} }}%
    \vspace{.4cm}
    \caption{\label{fig:simple-regret-epochs-ranking} Plot of simple and cumulative regret performance 
    of various policies and $\algo$ when it optimizes only for simple regret. \textbf{(Left)} $\algo$ is the 
    only policy that consistently outperforms Uniform. \textbf{(Right)} Plotting the cumulative regret performance 
    shows that both TS and TTTS incur lower cumulative regret than $\algo$, potentially showing that they are 
    too greedy, giving a possible explanation as to why they fail compared to $\algo$ and Uniform. }
\end{figure}
Figure \ref{fig:pareto-ranking} shows how $\algo$ is able to efficiently 
trade off between simple and cumulative regret. As more weight is put on the simple regret term 
by increasing $n_{T}$, the simple regret achieved monotonically decreases. As noted in the caption of Figure \ref{fig:pareto-ranking}, $\algo$ efficiently trades off between simple and cumulative regret and often \emph{strictly dominates} the performance of all variants of TTTS along both axes. It is noteworthy that $\algo$ handles this tradeoff in a principled and interpretable way simply by setting the weights equal to the number of individuals under each objective. On the other hand, there is no principled way of selecting the parameter values for TTTS, and most do not outperform Uniform in terms of simple regret.

We can also consider the \emph{weighted objective} directly, with the weight on cumulative and simple regret terms to be exactly $\sum_{t=0}^{T-1} n_{t}$ and $n_{T}$. Under this alternative setup, $\algo$ achieves the best objective value in $9$ out of the $13$ combinations of $\sum_{t=0}^{T-1} n_{t}$ and $n_{T}$ tested in Figure \ref{fig:pareto-ranking}. 




It is worth noting that when only optimizing for cumulative regret, $\algo$ exactly recovers a greedy policy. 
The strong performance of greedy policies over TS-based policies over short time horizons has been observed in \cite{RussoVaKaOsWe18}, further supporting $\algo$'s ability to plan when there are limited opportunities for adaptivity. 

\noindent \newline \textbf{Simple regret minimization with varying reallocation epochs.} 
Figure \ref{fig:simple-regret-ranking} measures simple regret performance of various 
policies while varying the number of reallocation epochs within the experiment. $\algo$ is able to consistently outperform Uniform, however, TS and TTTS do not. To understand why this is this case, \ref{fig:cumul-regret-ranking} plots the cumulative regret performance of the various policies. It is noteworthy that both TS and TTTS incur lower cumulative regret compared to both $\algo$ and Uniform, potentially giving an explanation as to why TS and TTTS fail to outperform Uniform in terms of simple regret. This also shows that $\algo$ automatically calibrates its level of exploration such that 
when only optimizing for simple regret, it explores potentially bad options instead of sampling just the best arm.  

\noindent \newline \textbf{A tradition single-item formulation.}
\begin{figure}[t]
  \centering
  \hspace{-.9cm}
    \subfloat[\centering Simple Regret Performance, $s^{2} = 0.2$, $\sigma^2 = 1$ ]{\label{fig:simple-regret-one-item}{\includegraphics[height=5cm]{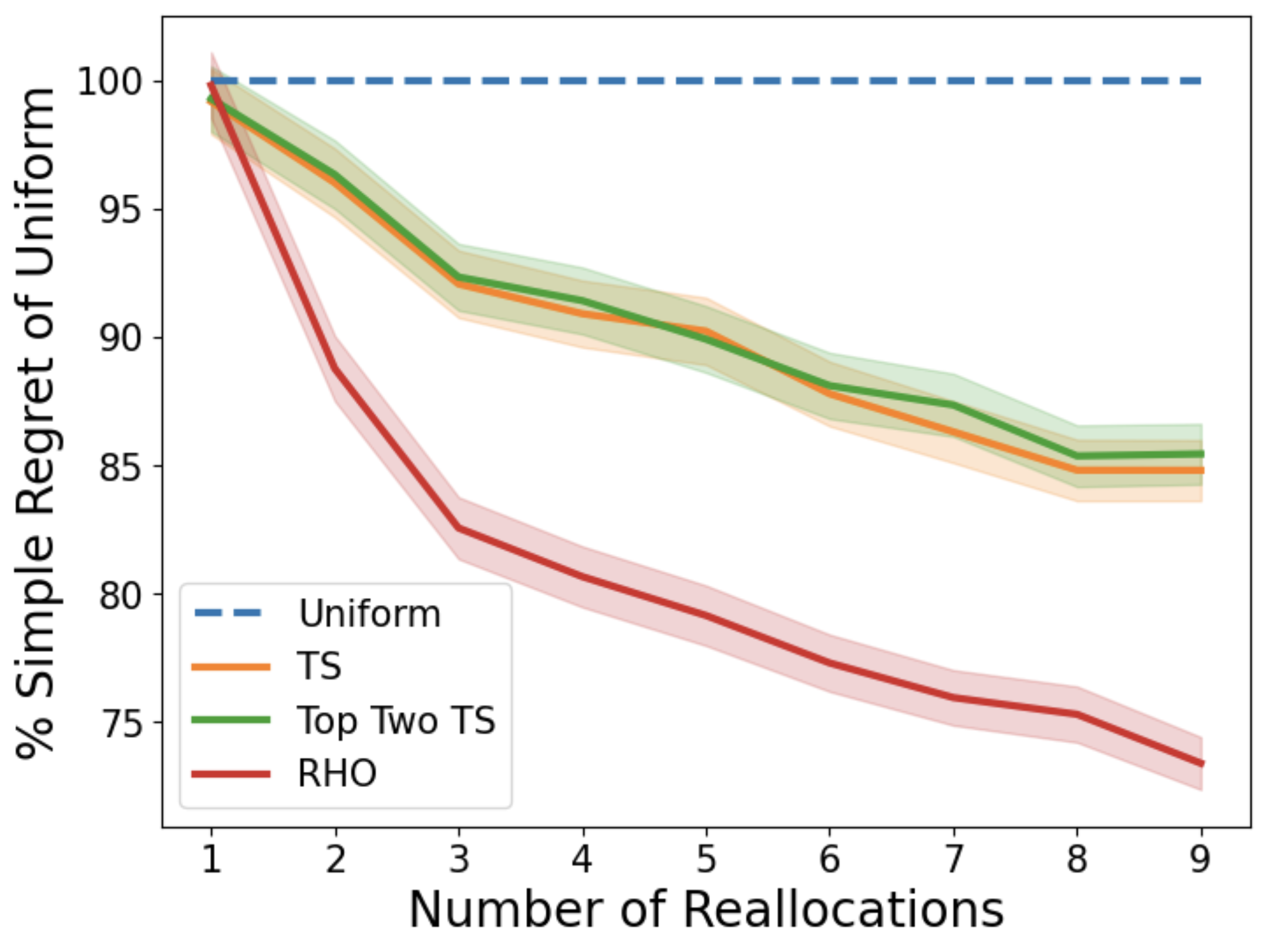}} }%
    \hspace{.3cm}
    \subfloat[\centering Pareto Frontier Between Simple and Cumulative Regret, epochs = $5$ ]{\label{fig:pareto-one-item}{\includegraphics[height=6cm]{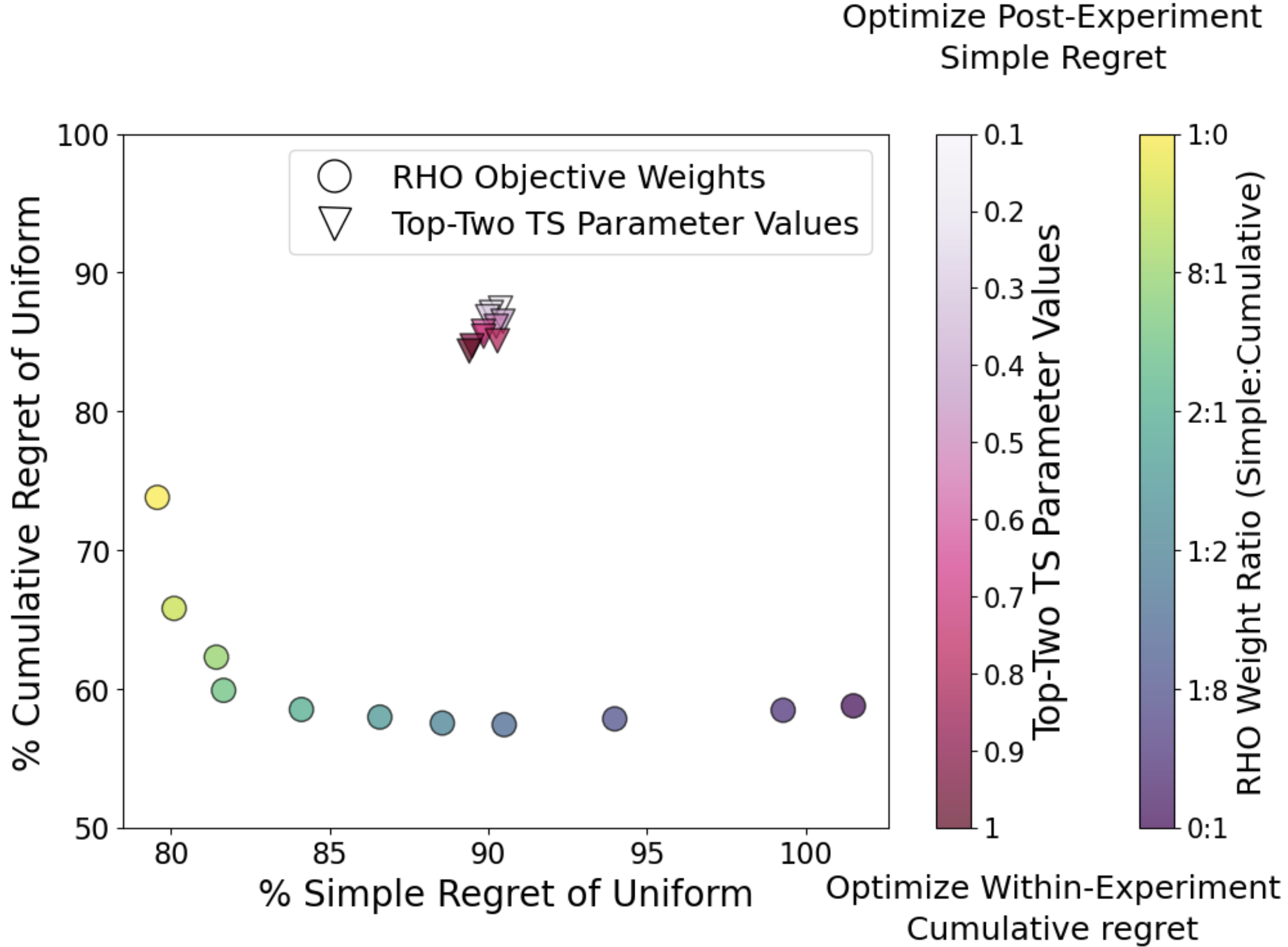} }}%
    \vspace{.4cm}
    \caption{\label{fig:performance-one-item} \textbf{(Left)} TS and TTTS outperform Uniform in simple regret 
    minimization across various experiment lengths, unlike in the ranking example. 
    This indicates that TS may be better suited to this setting. 
    $\algo$ RHO still outperforms TS policies by a large margin. \textbf{(Right)} Lighter values linked to TS
    indicate parameters closer to $0$, while darker values indicate parameters closer to $1$. Lighter values 
    for $\algo$ mean higher weight on simple regret, while darker values are higher weights on cumulative regret. 
    $\algo$ is again able to efficiently trade off between simple and cumulative regret while TS policies have 
    relatively homogeneous performance. 
        }
\end{figure}
Many works have shown that TS has strong empirical and theoretical performance in a more 
traditional personalization setting \cite{ChapelleLi11, agrawal13thompson}. Instead of choosing a ranking policy based on a value model, the traditional setting typically involves item-by-item, fully-sequential content recommendation (i.e., no batches of users or ranking and presenting multiple items per user). This involves learning separate coefficients $\theta_{z}$ for each action, or content to be recommended \cite{li2010contextual}. Similarly to the ranking setting, we have $x \sim N(\mu_{\text{user}}, \sigma_{\text{user}}^{2}I)$ and rewards for user $x$ and content $z$ are distributed as $N(x^\top \theta_{z}, s^2)$.

Each epoch involves a batch of $100$ samples. Figure \ref{fig:simple-regret-one-item}
shows that unlike in the ranking example (Figure \ref{fig:simple-regret-ranking}), TS policies outperform Uniform by a wide margin in simple regret minimization. In contrast, $\algo$ showcases strong performance in both settings, exhibiting its ability to adapt and optimize for specific conditions. Figure \ref{fig:pareto-one-item} showcases how $\algo$ is able to efficiently optimize both simple and cumulative regret when various weights are placed on each objective. Values for TS are rather homogeneous and many objective values of $\algo$ strictly dominate those of TS. 

\begin{table}[h!]
\centering
\small
\begin{tabular}{lcccc}
\toprule
\textbf{Policy} & \(s^2 = 0.2, \sigma^2_{\text{user}} = 0.1\) & \(s^2 = 0.2, \sigma^2_{\text{user}} = 5\) & \(s^2 = 1, \sigma^2_{\text{user}} = 0.1\) & \(s^2 = 1, \sigma^2_{\text{user}} = 5\) \\
\midrule
(Gumbel) TS        & 67.0 & 82.8 & 94.22 & 91.7 \\
Top-Two TS         & 69.1 & 93.8 & 93.8  & 94.0 \\
RHO                       & \textbf{61.8} & \textbf{80.3} & \textbf{81.1}  & \textbf{86.1} \\
\midrule
(Student’s t) TS   & 77.1 & 102  & 90.6  & 91.6 \\
Top-Two TS         & 76.1 & 97.8 & 90.8  & 88.9 \\
RHO                       & \textbf{60.5} & \textbf{83.0} & \textbf{66.5}  & \textbf{83.3} \\
\bottomrule
\end{tabular}
\caption{There are a total of $10$ epochs. Performance of policies under non-standard noise distributions, the Gumbel and Student's $t$. We also vary the measurement variance level and context distribution. In all settings, $\algo$ outperforms 
  TS policies, and performs particularly well under the student's t distribution. }
  \label{tab:one-item-diff-noise}
\end{table}

While the experiment in Figure \ref{fig:performance-one-item} involves noise 
sampled from a Gaussian distribution, we further benchmark the algorithms' performance under non-standard noise distributions. We choose the Gumbel distribution which involves high skewness, and the Student's $t$ distribution, which involves heavy tails. Table \ref{tab:one-item-diff-noise} keeps the number of epochs constant at $10$ and shows the performance of policies under these distributions as well as various measurement variance and context distribution levels. We find that under all circumstances, $\algo$ is able to outperform. These results suggest that $\algo$ can be robust to other noise distributions.

\subsection{Budget Constraints}
We further benchmark $\algo$'s ability to handle budget constraints, a large practical concern. Our basic simulation setup follows that of the single item formulation for ranking and recommendations in section~\ref{subsection:personalization}, but we repeat the setup here for clarity. There are $T$ epochs of experimentation with $n_{t}$ individuals at each epoch $t$. Each individual has features $x_{i}\in \R^{d}$ where $x \sim N(\mu, \Sigma)$ and there are $K$ arms. Playing arm $a_{i}$ on individual $x_{i}$ yields a linear reward of $R(x_{i},a_{i}) = x_{i}^\top \theta_{a_{i}} + \varepsilon_{i}$, where $\varepsilon \sim N(0, s^{2})$. Furthermore, each arm is associated with a fixed cost $c_{t}(a)$ which may vary across epochs. At each time $t$, the costs of each arm are sampled from a Half Normal distribution with mean $0$ and variance $1$.

\noindent \newline \textbf{Objectives.} We consider a range of objectives, where the experimenter is interested in balancing cumulative and/or simple regret while maintaining costs at each epoch $t$ under a fixed budget $B_{t}$. While section~\ref{subsection:personalization} only looks at the tradeoff between simple and cumulative regret, this setting with a fixed budget is relevant as real-world experiments are often constrained by their costs. 

\noindent \newline \textbf{Algorithmic Details.} To satisfy the budget constraint $B_{t}$, we easily incorporate projected gradient descent into $\algo$. Since $\algo$ involves stochastic gradient descent, we first take a gradient step on the regret objective. Then we use gradient steps to minimize 
\[
 g_{t}(p_{t}) = \sum_{a \in [K]} \left(\sum_{i=1}^{n_{t}} p_{t}(x_{i}, a)\right) c(a),
\]
until $g_{t}(p_{t}) \leq B_{t}$. We compare $\algo$ to Budgeted Thompson Sampling~\cite{xia2015thompson} which works as follows. Given a context $x_{i}$, Budgeted TS chooses an action by sampling $\what{\beta}_{t} \sim N(\beta_{t},\Sigma_{t})$ and then selecting $a_{i} = \argmax_{a \in [K]}\frac{ x_{i}^\top \what{\beta}_{t}}{c_{t}(a)}$. To weight the costs to be more or less important, we introduce a scaling factor $b \in \R$ such that Budgeted TS plays $a_{i} = \argmax_{a \in [K]}\frac{ x_{i}^\top \what{\beta}_{t}}{\exp(b*c_{t}(a))}$. Setting $b=0$ yields a denominator of $1$ and no influence by the costs. To tradeoff between simple and cumulative regret, we use a Top-Two TS~\cite{Russo20} variant of Budgeted TS. 
\begin{figure}[t]
  \centering
  \includegraphics[width=0.65\textwidth]{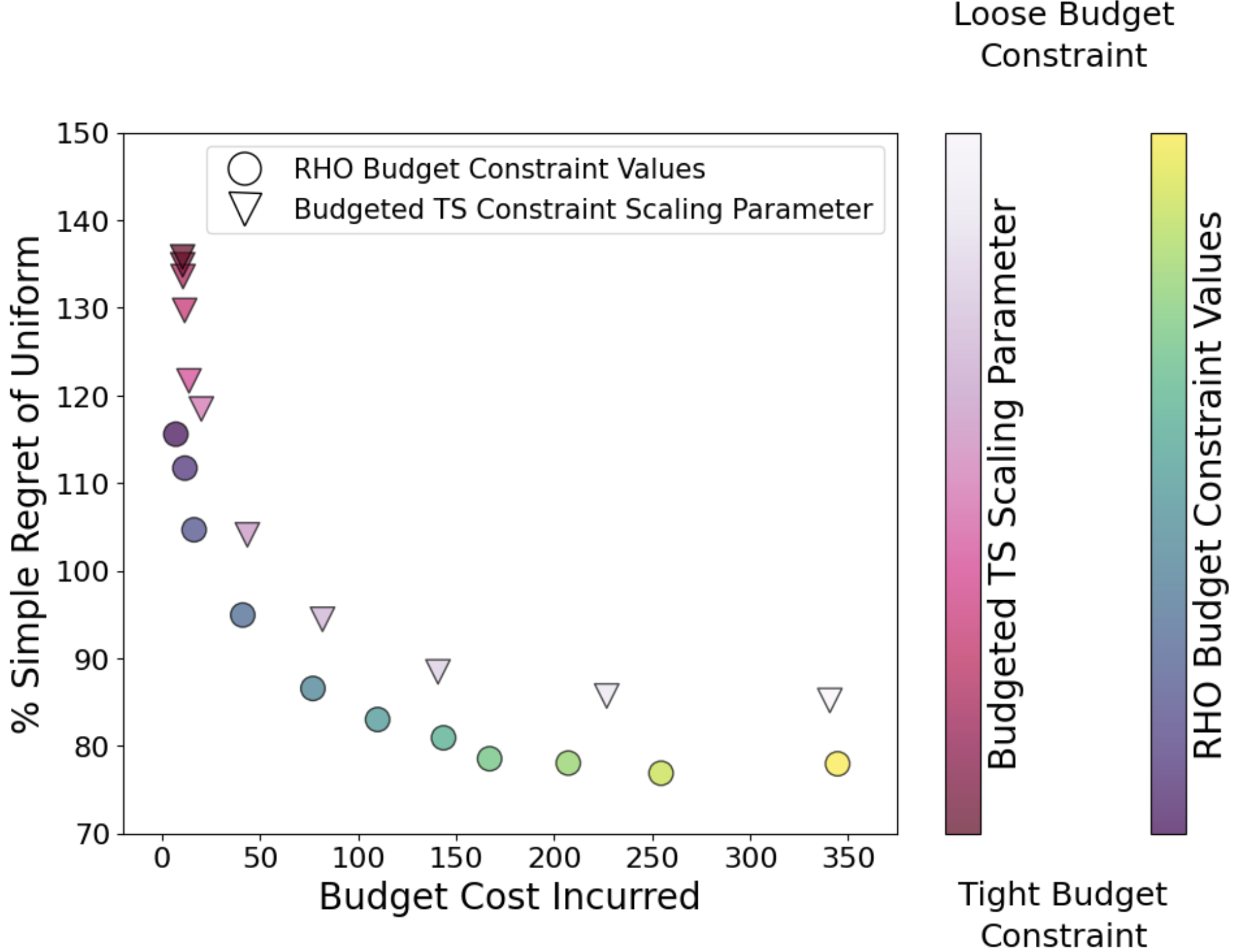}
  \caption{Pareto frontier between simple regret incurred and the average budget cost used by the algorithm throughout the experiment. $\algo$ strictly dominates that of Budgeted TS. Additionally, satisfying budget constraints with $\algo$ is easy since it uses projected gradient descent to exactly satisfy the constraint in expectation. In contrast, while Budgeted TS can trade off between these objectives, there is no principled method of setting the cost scaling parameter $b \in \R$ and is difficult to tune.  
  }
  \label{fig:pareto-simple-constraints}
\end{figure}

\subsubsection{Results and Discussion for Budget Constraints}

We first run an experiment involving trading off between simple and cumulative regret in section~\ref{subsection:personalization} while imposing an exact budget constraint. Figure~\ref{fig:pareto-constraints} showcases $\algo$'s ability to efficiently tradeoff between these two objectives while exactly satisfying the budget constraint via projected gradient descent. In contrast, Budgeted Top-Two TS involves two conflicting parameters that are difficult to tune. In Figure~\ref{fig:pareto-simple-constraints}, we look at the pareto frontier between the budget costs and simple regret incurred. We find that $\algo$ strictly dominates Budgeted TS.


\section{Asymptotics for Batch Adaptive Experiments}
\label{section:asymptotics}




The theoretical foundation for the $\model$ presented in
Section~\ref{section:algorithms} is the asymptotic
normality~\eqref{eqn:informal-clt} of the estimators $\what{\theta}_{n,t}$
obtained by empirical risk minimization~\eqref{eqn:batch-estimator} over a
batch.  We now prove this result rigorously. The key technical challenge is to
show not only that this applies for the estimator computed in a single batch,
but also for the entire sequence of estimators, which justifies the dynamic
programming formulation.  
While we focus on the case where the Hessian
$H_{t}$~\eqref{eqn:hessian} is invertible in  Section~\ref{section:algorithms},
the main results we prove in this section allow the empirical and population Hessians to be singular.
This enables us to apply our asymptotic results to a much wider range of adaptive policies, 
feature mappings, and non-stationary context arrivals.
Our result applies to general loss minimization problems and significantly generalizes the validity of  sequential normal approximations compared to~\citet{CheNa23} who focus on non-contextual and stationary settings.

The main difficulty is in showing convergence of sampling statistics under
sampling probabilities that are themselves stochastic, as they are influenced by previous measurements.
Standard asymptotic normality results for adaptively sampled data
show results for the special case where $\what{\theta}_{n,t}$ is the 
OLS estimator~\cite{HadadHiZhAt21, LuedtkeVa16, LuedtkeVa18, ZhangJaMu20}.
This typically requires a uniform lower bound on the minimum eigenvalue of the design matrix.
Even when the sampling policy $\pi_{t} > \delta$ is bounded from below for some $\delta > 0$ 
---a standard assumption for inference under adaptive sampling which we do not make---the design matrix could still be non-invertible
due to non-stationary context arrivals. 
    
\begin{example}~\label{example:monday_tuesday}(Monday/Tuesday).
   Consider a batched experiment with $T = 2$ epochs (corresponding to days of the week) and $K = 1$ arm.
  The context is simply the day-of-the-week, i.e. $x = t$. We consider a simple reward model with coefficient vector
  $\theta = (\theta^{\text{day}}_{0}, \theta^{\text{day}}_{1}, \theta^{\text{arm}})\in \R^{3}$
  where rewards are composed of a time fixed effect and an arm fixed effect
  and the loss function is least squares:
  \[
    r(t,a) = \phi(x,a)^\top \theta = \theta^{\text{day}}_{t} + \theta^{\text{arm}}
  \]
  At $t = 0$, suppose 100 sampling units arrive. They will all share the same context $x = t = 0$ and are all assigned 
  to treatment $a=0$. The sample covariance matrix $\sum_{i=1}^{100} \phi(0,0)\phi(0,0)^\top$ will then yield
  \[
    \lambda_{\min}\left(\sum_{i=1}^{100} \phi(0,0)\phi(0,0)^\top \right) = 
    \lambda_{\min}\left( 
      \begin{bmatrix} 
        100 & 0 & 100 \\
        0  & 0 & 0 \\
        100 & 0 & 100\\
      \end{bmatrix}
    \right) = 0
  \]
  This is because the context $t = 1$ is never observed in epoch $t = 0$.
\end{example}
To handle these cases, our proof method instead shows convergence of batched rewards scaled by the Hessian.
This approach obviates the need for minimum eigenvalues of population/empirical Hessians
and the sampling allocations to be bounded away from zero. 

\subsection{Diffusion scaling}
\label{section:diffusion}

Recall that $\mc{D}_{t} := \{\left(X_{t,i}, A_{t,i}, R_{n,t,i} \right)\}_{i=1}^{n_{t}}$ denotes
the data collected in batch $t$, and let $\mc{D}_{t,i} = \left( X_{t,i}, A_{t,i}, R_{n,t,i} \right)$ 
denote the tuple of observations for sampling unit $i$.
We let $\ell(\theta|\mc{D}_{t,i}) = \ell(\theta| X_{t,i}, A_{t,i}, R_{n,t,i})$ 
be the statistical loss for unit $i$ and $L_{n_{t}}(\theta | \mc{D}_{t})$ denote 
the empirical risk minimization objective under the batch dataset $\mc{D}_{t}$.
We let $\what{\theta}_{n,t}$ be a minimizer of the empirical loss: 
\begin{equation}
\label{eqn:erm_prelimit}
\what{\theta}_{n,t} \in \argmin_{\theta \in \Theta}
\left\{L_{n_{t}}(\theta | \mc{D}_{t}) 
  = \frac{1}{n_{t}} \sum_{t=1}^{n_{t}} \ell(\theta| X_{t,i}, A_{t,i}, R_{n,t,i})
  \right\}
\end{equation}
We emphasize that this loss function depends on the sampling allocation 
selected in batch $t$, and review the data-generating process for batch $t$:
\[
  X_{t,i} \simiid \mu_{t}, \quad A_{t,i} \simiid p_{t}(\cdot | X_{t,i}), \quad R_{n,t,i} \simiid \nu(\cdot | X_{t,i}, A_{t,i})  
\]
where $\mu_{t}$ can be different for different $t$'s and $\nu$ is unknown. Since we allow $p_{t}(\cdot | X_{t,i})$ to sum to less than one and even be the zero action $p_{t}(\cdot | X_{t,i}) = \textbf{0}$, if no actions are taken and no observations are received, then we set $\what{\theta}_{n,t} = 0$. 
In this section, we let $\E_{t}$ denote expectation with respect to the above data-generating distribution.

We can define several key quantities related to the curvature of the loss
function.  Recalling the population Hessian and gradient covariance at time
$t$~\eqref{eqn:fisher}, their
empirical counterparts are
\begin{subequations}
  \label{eqn:emp_fisher}
  \begin{align}
  \what{H}_{n,t} &\defeq \frac{1}{n_{t}} \sum\nolimits_{t=1}^{n_{t}} \nabla_{\theta}^{2} \ell(\what{\theta}_{n,t} |\mc{D}_{t,i}) 
  \label{eqn:emp_hessian}\\
  \what{I}_{n,t} &\defeq \frac{1}{n_{t}} \sum\nolimits_{t=1}^{n_{t}}  \nabla_{\theta} \ell(\what{\theta}_{n,t} |\mc{D}_{t,i}) \nabla_{\theta} \ell(\what{\theta}_{n,t} |\mc{D}_{t,i})^{\top}.
  \label{eqn:emp_grad_cov}
  \end{align}
\end{subequations}
To illustrate, under a linear reward model with the least-squares loss
function, $r(x,a) = \phi(x,a)^{\top}\theta\opt$, and residual variance
$\var(R|X,A) = 1$, these quantities correspond with the population and
empirical design matrices, where we define $\Phi_{n,t}\in \R^{n_{t} \times d}$
to be the matrix of feature vectors 
\[
  H_{t} = I_{t} = \E_{t}[\phi(X_{t,i}, A_{t,i})\phi(X_{t,i}, A_{t,i})^{\top}],\qquad 
  \what{H}_{n,t} = \what{I}_{n,t} = \frac{1}{n} \sum_{i=1}^{n_{t}} \phi(X_{t,i}, A_{t,i})\phi(X_{t,i}, A_{t,i})^\top .
\]
To formalize our use of normal approximations, we consider a scaling parameter $n$ 
so as $n\to \infty$, batch sizes grow large while the true parameter becomes harder to
identify
\begin{equation}
  \label{eq:scaling}
  r_{n}(x,a) =  f(x,a;\theta\opt_{n}) = f(x,a;\theta\opt/\sqrt{n}), \quad \quad n_{t} = n,
\end{equation}
so that the signal-to-noise ratio remains fixed as batch size grows.
For simplicity, we assume that the batch size remains constant across epochs
such that $n_{t} = n$, but our theory trivially extends 
to varying batch sizes.
Intuitively, scaling the parameter $\theta\opt$ also scales the gap between
average rewards of each arm, making it harder to identify the best arm.  For
example, if $\theta\opt$ represents the mean of a distribution, then scaling
$\theta\opt$ by $1/\sqrt{n}$ scales the gap between average rewards by the
same factor. In the case of linear and logistic regression, scaling
$\theta\opt$ reduces the effect size of the covariate on the linear response
and log-odds respectively.
This is a classical scaling used to analyze the statistical power of binary hypothesis tests
and heavy traffic behavior of queueing systems~\cite{Whitt02}; more  recently, it been used to analyze adaptive experiments in the sequential limit $T \to \infty$~\cite{KalvitZe21, WagerXu21, FanGl21, Adusumilli21,
  AramanCa22}
or the large batch limit $n \to \infty$~\cite{HiranoPo23}. Our work extends the classical \emph{Gaussian experiment}~\cite{VanDerVaart98, LeCamYa00}
to sequential M-estimators.

In the general case where $H_{t}$ is allowed to be non-invertible, the statistical estimator we analyze is
not $\what{\theta}_{n,t}$ but rather $\what{H}_{n,t} \what{\theta}_{n,t}$. This is because when $H_{t}^{-1}$ does not exist,
asymptotic normality does not necessarily hold for the minimizer $\what{\theta}_{n,t}$ and even the minimizer may not be unique.
However, $\what{H}_{n,t} \what{\theta}_{n,t}$ satisfies a well-defined central limit theorem and the limiting random variable gives rise 
to a well-defined generalization of the Bayesian posterior update.
Precisely, the sufficient statistic we consider is
\begin{equation}
  \label{eqn:sufficient_stats}
  \what{\Psi}_{t} := \sqrt{n} \what{H}_{n,t} \what{\theta}_{n,t}.
\end{equation}
The $\sqrt{n}$ scaling is needed for $\what{\Psi}_{t}$ to be $O(1)$ because the
target parameter $\theta\opt_{n}$ is $O(1/\sqrt{n})$.


\subsection{Sequential Gaussian Experiment for M-estimators}

We show that conditional on previous observations $\what{\theta}_{0}, ..., \what{\theta}_{t-1}$,
which affect the arm assignments $A_{t,i}$ through the policy $\pi_{t}(\cdot | x, \what{\Psi}_{0:t-1})$,
$\what{\Psi}_{t}$ converges in distribution to a Gaussian random variable.
We refer to this as the \emph{Batched Limit Sequential Experiment}.
\begin{definition}
  \label{def:blse}
  The Batched Limit Sequential Experiment is characterized by observations
  $G_0, \ldots, G_{T-1}$ with conditional distributions
  \begin{equation*}
    G_t | G_0, \ldots, G_{t-1} \sim
    N \left(H_{t}\theta\opt,I_{t} \right)
  \end{equation*}
\end{definition}
\noindent If $H_{t}$ is non-invertible, this only provides a partial
observation of $\theta\opt$ as it is well-known that unbiased estimators do
not exist in this case.  Nevertheless, one can still perform Bayesian updating
with this partial signal.

In order to rigorously show convergence, we require mild regularity assumption on the policy.
\begin{assumption}
  \label{assumption:policy} We consider policies $\pi = \{ \pi_{t} \}_{t=1}^{T}$ where
  \begin{enumerate}
  \item \label{item:sufficiency} \emph{Sufficient Statistic}: $\pi$ depends on the
    data only through the sufficient statistics,
\begin{equation*}
  \pi_{t}(\cdot | x, \mc{H}_{t})=\pi_{t}\left(\cdot |x, \what{\Psi}_{0:t-1} \right) ~~~\mbox{for all}~t= 0, \cdots, T-1
  \end{equation*}
  where $\what{\Psi}_{0:t} = (\what{\Psi}_{1}, \ldots, \what{\Psi}_{t})$
  \item \label{item:continuity} \emph{Continuity}: The policy 
    $\pi_{t}(a | x, \what{\Psi}_{0:t-1})$ is continuous in inputs $\what{\Psi}_{0:t}$ for all
    $a,t$.
  \end{enumerate}
\end{assumption}
 \noindent We also allow the $\pi_{t}$ to depend on the Hessian and the gradient covariance matrices as well. 
These are mild assumptions, and for standard settings such as linear-contextual bandits,
this class of policies includes Linear Thompson Sampling~\cite{AgrawalGo13b}. 

We do not assume the Hessian of the population
loss $\E_{t} [\ell(\theta| \mc{D}_{i}^{t})]$ is strictly positive definite, which could be
the case if $\pi_{t}$ does not sample certain treatment arms or if there is
severe non-stationarity in the context distribution.  As a result, we relax
standard assumptions regarding strong convexity of the population loss.
Instead, we require strong convexity only for the parameters in the range of
the population Hessian, which is a much weaker assumption that is consistent
with the fact that the minimum eigenvalue $\lambda_{\min}(H_{t}) = 0$ can
equal zero.

Let $P_{t}$ be the orthogonal projection matrix for the range space of the population Hessian $H_{t}$. 
We require the following mild assumptions, which 
are generally satisfied by all standard convex loss functions (e.g. mean squared error, logistic loss).
\begin{assumption} 
  \label{assumption:m-estimation} 
\begin{enumerate}
  \item  \label{item:strong-convexity}
    \emph{Strong Convexity}: The population loss $\ell (\theta| \cdot)$ is twice differentiable with respect 
    to $\theta$ and there exists $\mu > 0$ such that for any $\theta\in\Theta$, 
    \begin{equation}
      \label{eqn:strong-convexity}
      \E_{t} [\ell(\theta| \mc{D}_{t,i})] -  \E_{t} [\ell(\theta\opt_{n}| \mc{D}_{t,i})] \geq \mu \norm{P_{t}(\theta - \theta\opt_{n})}_{2}^{2}
    \end{equation} 

  \item \label{item:lipschitz} \emph{Lipschitz Condition}: there exists $c_{1}>0, c_{2} > 0$ such that for any 
  $\mc{D}_{t,i} = (X_{t,i}, A_{t,i}, R_{n,t,i})$ and $\theta\in\Theta$, 
    \begin{equation}
      \label{eqn:lipschitz}
      \norm{\ell (\theta| \mc{D}_{t,i}) - \ell (\theta\opt_{n}| \mc{D}_{t,i}) }_{2} \leq c_{1} \norm{P_{t}(\theta - \theta\opt_{n})}_{2}
    \end{equation}
    \begin{equation}
      \label{eqn:lipschitz-hessian}
      \norm{\nabla^{2}_{\theta} \ell (\theta| \mc{D}_{t,i}) - \nabla^{2}_{\theta} \ell (\theta\opt_{n}| \mc{D}_{t,i}) }_{2} \leq c_{2} \norm{P_{t}(\theta - \theta\opt_{n})}_{2}
    \end{equation}
  \item \label{item:boundedness} \emph{Boundedness Condition}: there exists $c_{3}, c_{4} > 0$ such that 
    \begin{align}
      \norm{\nabla \ell (\theta| \mc{D}_{t,i})}_{2} \leq c_{3}, \; \nabla^{2} \ell (\theta| \mc{D}_{t,i}) \preceq c_{4} I \text{ for all } X_{t,i}, A_{t,i}, R_{t,i}, \theta \in \Theta
    \end{align}
    Furthermore, $\| \theta\opt ||_2 < \infty$. 
  \item \label{item:optimality} \emph{Optimality Condition}: At each epoch $t$, the empirical risk minimizer 
  $\what{\theta}_{n,t} \in \argmin_{\theta \in \Theta} L_{n_{t}}(\theta | \mc{D}_{t})$ \eqref{eqn:erm_prelimit} is 
  attained in the interior of the constraint set $\Theta$. 
\end{enumerate}
\end{assumption}

It is worth mentioning what is not assumed. First, we do not require sampling probabilities
to be bounded away from zero, i.e. $\pi_{t}(a|x, \what{\Psi}_{0:t-1})>0$. Second, unlike standard results
for inference of bandit algorithms and M-estimation in general~\cite{ZhangJaMu20, Shao2021Berry, VanDerVaart98, VanDerVaartWe96, LeCamYa00}, we do not require that the Hessian 
is invertible, i.e., we allow both the empirical and population versions to have 
a zero eigenvalue. For linear least-squares with $r(x,a) = \phi(x,a)^{\top}\theta\opt$,
this corresponds to a singular design matrix,
\[
  \lambda_{\min}\left( \frac{1}{n_{t}}\sum_{i=1}^{n_{t}} \phi(X_{t,i}, A_{t,i})\phi(X_{t,i}, A_{t,i})^\top \right) = 0, 
  \quad \quad \lambda_{\min}\left( \E_{t}[\phi \phi^{\top}] \right) = 0.
\]
Rather than being a purely technical matter, this is critical for being able to
show the validity of the contextual Gaussian sequential experiment under non-stationary context arrivals,
as we will discuss later.

Finally, we require the following assumption on the Hessian of the loss function. 

\begin{assumption}
  \label{assumption:hessian}
  For each $t$,
    $\E_{t} \left[ \left\| \frac{1}{\sqrt{n}} \sum\limits_{i=1}^{n} \left(\nabla^{2}_{\theta} \ell(\what{\theta}_{n,t}| \mc{D}_{i}^{t}) -  \nabla^{2}_{\theta} \ell(\theta\opt_{n}| \mc{D}_{i}^{t})\right) \right\|_{2} \right] \to 0$ as $n\to \infty$. 
\end{assumption}
\noindent This assumption is automatically satisfied by a range of common models such as 
linear models, as their Hessians are solely dependent on the data and 
independent of the parameter. When performing traditional M-estimation \cite{VanDerVaart98, LeCamYa00}, consistency of the empirical risk minimizer yields consistent estimates of the Hessian and Fisher information, which is only needed for valid inference purposes. This assumption is needed in our formulation as we prove convergence of the statistic $\what{\Psi}_{n,t}$ which involves the empirical Hessian, instead of just $\what{\theta}_{n,t}$. 

We now present our main asymptotic result, which shows that as $n\to \infty$, the contextual
Gaussian sequential experiment is a valid approximation for the batch statistics
even under adaptive sampling policies. We prove this in
Section~\ref{section:proof-limit}. 
\begin{theorem}
  \label{theorem:limit}
  Let Assumptions~\ref{assumption:policy},~\ref{assumption:m-estimation},~\ref{assumption:hessian} hold. 
  Under a fixed policy $\{ \pi_{t} \}_{t=0}^{T-1}$,
  the Gaussian sequential experiment in Definition~\ref{def:blse} provides a
  valid asymptotic approximation for 
    $\what{\Psi}_{t} := \sqrt{n} \what{H}_{n,t} \what{\theta}_{n,t}$
  \begin{equation}
    \label{eqn:weak-convergence}
    (\what{\Psi}_{0},\ldots,\what{\Psi}_{T-1}) \cd
    (G_{0},...,G_{T-1})
    ~~~\mbox{as batch size}~n \to \infty.
  \end{equation}
\end{theorem}
\noindent Note that if $p_{t}(\cdot | X_{t,i})$ is close to the zero action, then the Hessian and Fisher information~\eqref{eqn:emp_fisher} will approach $0$. If $p_{t}(\cdot | X_{t,i}) = \textbf{0}$, then $G_{t}$ and $\what{\Psi}_{t}$ will be equal to $0$. 



\subsection{Asymptotic Validity of $\model$}

Upon observing a Gaussian random variable with $N(H_{t}\theta\opt, n_{t}^{-1}I_{t})$,
the posterior update for $\theta\opt$ 
is given by the following.
\begin{definition}
  \label{definition:generalized_mdp}
  Let $\theta\opt \sim N(\beta_0, \Sigma_0)$. Given observations
  $\{G_{t} \}_{t=1}^{T}$ where
  $G_{t} | G_{1:t-1} \sim N(H_{t}\theta\opt, n_{t}^{-1}I_{t} )$, the
  joint distribution of posterior states
  $\{(\beta_{t}, \Sigma_{t}) \}_{t=0}^{T}$ are characterized by the
  recursive relation
  \begin{subequations}
    \label{eqn:gen-dynamics}
    \begin{align}
      \mbox{Posterior variance:}~~~~ \Sigma_{t+1}^{-1}
      & \defeq \Sigma_{t}^{-1} + H_{t} I_{t}^{\dagger} H_{t}
        \label{eqn:gen-dynamics-var} \\
      \mbox{Posterior mean:}~~~~\beta_{t+1}
      &:= \beta_{t} + (\Sigma_{t} - \Sigma_{t+1})^{1/2} Z_{t}
        \label{eqn:gen-dynamics-mean}
    \end{align}
  \end{subequations}
  where $H_{t}$ and $I_{t}$ are defined in~\eqref{eqn:hessian}
  and~\eqref{eqn:grad_cov} respectively, and
  $Z_0, \ldots, Z_{T-1} \simiid N(0, I_{d})$ are standard normal variables.
\end{definition}

Unlike the MDP~\eqref{eqn:dynamics} in Section~\ref{section:algorithms}, the above MDP does not have $n$ in the update for $\Sigma$ as this is under the rescaled rewards $r_{n}$.
By rescaling removing the $\sqrt{n}$ factor in $\Phi_{n,t}$, we can recover the same scaling as in~\eqref{eqn:dynamics}.
Under this scaling, we let $R_{n,t} = \{ R_{n,t,i} \}_{i=1}^{n_{t}}$ be the vector of rewards collected from epoch $t$.
Note that $R_{n,t}$ depends on $n$ through the number of observations $n_{t}$ and also the
scaling of the average rewards $r_{n}$ as $\E[R_{n,t,i}|X_{t,i} = x, A_{t,i} = a] = r_{n}(x,a)$. 
As the batch size grows large, the population-level quantities become relevant.

Naturally, the dynamics~\eqref{eqn:gen-dynamics} cannot be used exactly since the population covariance is unknown and
the measurement $\what{\Psi}_{t}$ is used instead of $G_{t}$. So instead, we can
plug-in the empirical counterparts to the asymptotic quantities, leading to an
approximate Bayesian model $(\beta_{n,t}, \Sigma_{n,t})$, which is compatible
with the data obtained from the batched experiment.
\begin{definition}
  The {\bf approximate Bayesian model} is defined by the following update rule for posterior parameters
  $(\beta_{n,t}, \Sigma_{n,t})$, upon observing aggregated rewards $\what{\Psi}_{t}$ and contexts $\Phi_{n,t}$:
  \begin{subequations}
    \label{eqn:approx_posterior}
    \begin{align}
      \mbox{Posterior variance:}~~~~ \Sigma_{n,t+1}^{-1}
      &\defeq \Sigma_{n,t}^{-1} + \left( \what{H}_{n,t} \what{I}_{n,t}^{\dagger} \what{H}_{n,t} \right)
        \label{eqn:approx_posterior-var} \\
      \mbox{Posterior mean:}~~~~\beta_{n,t+1}
      &\defeq \Sigma_{n,t+1} \left( \Sigma_{n,t}^{-1}\beta_{t} 
      + \what{H}_{n,t} \what{I}_{n,t}^{\dagger} \what{\Psi}_{t} \right)
        \label{eqn:approx_posterior-mean}
    \end{align}
  \end{subequations}
\end{definition}

Using the main asymptotic result in Theorem~\ref{theorem:limit}, we can show
that trajectories of the approximate Bayesian model converge in distribution
to trajectories of the asymptotic model as $n\to \infty$,
\begin{corollary}
  \label{cor:trajectory_limit}
  Let $\pi_{t}(\beta_{t}, \Sigma_{t})$ be a policy that is continuous
  a.s. under $(\beta_{t}, \Sigma_{t})$ drawn from~\eqref{eqn:gen-dynamics}.  Let
  Assumptions~\ref{assumption:policy}, ~\ref{assumption:m-estimation}, ~\ref{assumption:hessian} hold.
  Consider any fixed $\theta$ and prior
  $(\beta_{n,0}, \Sigma_{n,0}) = (\beta_{0}, \Sigma_{0})$.  The posterior
  states of~\eqref{eqn:approx_posterior} converge in distribution to the
  states~\eqref{eqn:gen-dynamics} as $n\to \infty$,
  \[
    (\beta_{n,1}, \Sigma_{n,1},\ldots, \beta_{n,t}, \Sigma_{n,t}) \cd (\beta_{1}, \Sigma_{1},\ldots, \beta_{t}, \Sigma_{t}).
  \]
\end{corollary}
\noindent This in turn implies convergence of the performance metrics and the
value functions. Ultimately, this justifies the use of the asymptotic model as
a guide for evaluating and optimizing performance metrics.
\begin{corollary}
  \label{cor:value_limit}
  If the objective functions $c_{s}(\pi_{s}, \beta_{s}, \Sigma_{s})$ are all
  continuous in all arguments a.s. under $(\beta_{t}, \Sigma_{t})$ drawn
  from~\eqref{eqn:gen-dynamics}, then
  $c_{s}(\pi_{s}, \beta_{n,s}, \Sigma_{n,s})\cd c_{s}(\pi_{s}, \beta_{s},
  \Sigma_{s})$ for all $s\leq T$.  If in addition,
  $c_{s} = O(\norm{\beta}_{2} + \norm{\Sigma}_{2})$ then we also have
  convergence of the value functions for all $t=0,...,T-1$
  \[
    \E_{t} \left[ \sum_{s=t}^{T} c_{s}(\pi_{s} | \beta_{n,s}, \Sigma_{n,s})\right] \to \E_{t} \left[ \sum_{s=t}^{T} c_{s}(\pi_{s} | \beta_{s}, \Sigma_{s})  \right]
  \]
\end{corollary}
\noindent The exact form of the approximate posterior updates
in~\eqref{eqn:approx_posterior} is useful for showing theoretical convergence,
but also has implications for posterior updates in batched settings.

\paragraph{Implications on practical performance: Scaling the design matrix or
  prior.} While the covariance update~\eqref{eqn:approx_posterior} scales the
design matrix by the batch size in order to obtain a sensible limit, we find
that this is also practically relevant.  By scaling down the covariance
matrix, or alternatively scaling up the prior by $n_{t}$, we manage to avoid
numerical difficulties that emerge when updating with large batches,
especially if the design matrix is singular or low-rank.  This numerical
difficulties emerge when trying to invert $\Sigma_{t+1}^{-1}$, which is going
have an extremely small minimum eigenvalue if the low-rank design matrix
dominates the prior.


\section{Discussion}
\label{section:discussion}

Normal approximations have been informally proposed in the simulation
optimization literature~\cite[Section 3.1]{KimNe07}, and normality is a common
\emph{assumption} in simulation optimization~\cite{HongNeXu15, ChenChLePu15},
Bayesian optimization~\cite{FrazierPoDa08, GonzalezOsLa16, LamWiWo16, WuFr19,
JiangJiBaKaGaGa19}, and bandits~\cite{LattimoreSz19}.  Our main theoretical
contribution establishes the validity of sequential Gaussian approximations
for general M-estimators in the large batch limit. Our formalization is
connected to Gaussian approximations for bandit
algorithms~\cite{LuedtkeVa16,ZhangJaMu20, HadadHiZhAt21}, and provides a
general guarantee that can model a wide range of M-estimators under
nonstationary settings and small policy weights. In
particular,~\citet{HiranoPo23} derives sequential Gaussian approximations for
non-contextual settings, and study inferential questions such as power
calculations.

Managing the exploration-exploitation trade-off computationally is a classical
idea in reinforcement learning~\cite{GhavamzadehMaPiTa15}, with applications
in ranking and selection~\cite{PengChChFu18} and Bayesian experimental
design~\cite{ChalonerVe95, RyanDrMcPe16, FosterIvMaRa21} where the goal is to
optimize information gain. For pure-exploration problems,~\citet{BhatFaMoSi20}
use covariate balancing to sequentially assign treatments as subjects arrive
stochastically, and derive a state state collapse behavior in specific
distributional settings.  Our work focuses on small-horizon adaptive
experiments similar to \citet{che2023adaptive}, but we significantly expand
the scope of computational methods (e.g., $\algo$) to handle practical problem
instances involving batched \& delayed feedback, non-stationarity, multiple
objectives \& constraints, and personalization.

As we emphasize throughout, our main goal is to improve upon the static
randomized control trial in a consistent and robust manner across problem
instances. There is much work to be done to materialize our mathematical
programming perspective. The simple SGD-based algorithm $\algo$ we study
cannot easily handle stochastic constraints and has to resort to Langragian
dual relaxations. We also do not expect it to perform well for large horizons
$T$, or problems that require substantial adaptivity. We hope this work spurs
future methodological innovations in solution approaches and corresponding
empirical foundations over which solvers can be benchmarked against a wide
range of instances. The mathematical programming community has long adopted
this two-pronged approach, and we expect a similar approach will be fruitful
in adaptive experimentation.




\bibliographystyle{abbrvnat}

\ifdefined\useorstyle
\setlength{\bibsep}{.0em}
\else
\setlength{\bibsep}{.7em}
\fi

\bibliography{adaptive_bib,extra_refs}

\ifdefined\useorstyle

\ECSwitch


\ECHead{Appendix}

\else
\newpage
\appendix

\fi

\section{Further Experimental Results}
\label{section:more-experiments}

\begin{tabular}{ccccc}
\toprule
& \multicolumn{2}{c}{Simple Regret $<$ Uniform} &  \multicolumn{1}{c}{Simple Regret $>$ Uniform} \\
\textbf{Policy} $b_{t} = 250,000$ & \% Settings & \% Avg Regret of Uniform & \% Avg Regret of Uniform \\
\cmidrule(lr){2-3} \cmidrule(lr){4-4}
Non-contextual TTTS~\cite{Russo20}   & 41.1  & 82.8 & 119.8 \\
Contextual TTTS~\cite{QinRusso2023}   &  55.1 & 83.9 & 107.2 \\
Contextual RHO  & 57.1  & 86.39 & 105.8 \\
\bottomrule
\end{tabular}

\vspace{1em}

\noindent \begin{tabular}{ccccc}
\toprule
& \multicolumn{2}{c}{Simple Regret $<$ Uniform} &  \multicolumn{1}{c}{Simple Regret $>$ Uniform} \\
\textbf{Policy} $b_{t} = 10,000$ & \% Settings & \% Avg Regret of Uniform & \% Avg Regret of Uniform \\
\cmidrule(lr){2-3} \cmidrule(lr){4-4}
Non-contextual TTTS~\cite{Russo20}   & 48.5  & 93.5 & 119.8 \\
Contextual TTTS~\cite{QinRusso2023}   &  50.6 & 94.2 & 103.3 \\
Contextual RHO  & 65.1  & 93.7 & 103.2 \\
\bottomrule
\end{tabular}

\section{Derivations of Gaussian sequential experiment}

\subsection{Proof of Theorem \ref{theorem:limit}}
\label{section:proof-limit}  
Recall the Gaussian sequential experiment given in Definition~\ref{def:blse}:
for all $0< t \le T-1$
\begin{equation*}
  G_{0}\sim N(H_{0}\theta\opt, I_{0}),
  ~~G_{t}|G_{t-1},\ldots,G_{0} \sim N(H_{t}\theta\opt, I_{t})
\end{equation*}
where the population Hessian and Fisher Information at time $t$ respectively are: 
\[
  H_{t} := \E_{t}[\nabla^{2} \ell (0| \mathcal{D}_{i}^{t})], \qquad I_{t} := \E_{t}[\nabla \ell (0| \mathcal{D}_{i}^{t}) \nabla \ell(0| \mathcal{D}_{i}^{t})^\top].
\]
Note that the Hessian and Fisher Information are evaluated at zero since $\theta\opt_{n} \to 0$ as $n\to \infty$.

To recapitulate, $\mc{D}_{i}^{t} := (X_{t,i}, A_{t,i}, R_{n,t,i})$, $\theta\opt$ is the true parameter, $\theta\opt_{n} = \theta\opt / \sqrt{n}$ is the scaled parameter where 
the $\sqrt{n}$ scaling maintains the difficulty of the experiment as the batch size grows. We also 
assume that $n_{t} = n$ for simplicity, but it is trivial to extend our result to differing batch sizes. 
Furthermore, 
\[
  X_{t,i} \simiid \mu_{t}, \quad A_{t,i} \simiid p_{t}(\cdot | X_{t,i}), \quad R_{n,t,i} \simiid \nu(\cdot | X_{t,i}, A_{t,i}).  
\]
Note that due to the dependence of $H_{t}$, and $I_{t}$ on $p_{t}$ and $p_{t}(\cdot | x) = \pi_{t}(\cdot| x, \mc{H}_{t})$ on previous observations,
$G_{t}$ depends on the outcomes of $G_{t-1},...,G_{0}$. Define 
\[
  \what{H}_{n,t} := \frac{1}{n} \sum_{i=1}^{n} \nabla^{2} \ell (\what{\theta}_{n,t}| \mathcal{D}_{i}^{t}), \qquad H_{n,t} := \frac{1}{n} \sum_{i=1}^{n} \nabla_{\theta}^{2} \ell(\theta\opt_{n} | \mathcal{D}_{i}^{t}). 
\]
We use a shorthand $\what{\Psi}_{n,t}$ to 
denote the estimator
\[
  \what{\Psi}_{n,t} := \sqrt{n} \what{H}_{n,t}\what{\theta}_{n,t}
\]
Since $p_{t}(\cdot | X_{t,i}) \in \mc{S}_{K} := \{(x_{1}, ..., x_{K}) \mid x_{i} \geq 0 \text{ and } \sum_{i=1}^{K} x_{i} \; \forall i \in [K]\}$, meaning $p_{t}$ can sum to less than one, we assume that if no action is sampled, then $\mc{D}_{t,i}$ is excluded from the loss minimization objective. Furthermore, if no actions are sampled for all $X_{t,i}$, then we set $\what{\theta}_{n,t} = 0$. 
As $\what{\theta}_{n,t} \in \argmin_{\theta \in \Theta} L_{n}(\theta| \mc{D}_{t})~\eqref{eqn:erm_prelimit}$ and $\what{\theta}_{n,t}$ is in the 
interior of the constraint set $\Theta$ by assumption \ref{item:optimality}, it follows that $\nabla_{\theta} L_{n}(\what{\theta}_{n,t}| \mc{D}_{t}) = 0$.
Therefore, we can take a Taylor Expansion of the gradient of the empirical loss around $\what{\theta}_{n,t}$: 
\begin{equation}
  \label{eqn:taylor-expansion}
  0 = \nabla_{\theta} L_{n}(\what{\theta}_{n,t}| \mc{D}_{t})
  =  \nabla_{\theta} L_{n}(\theta\opt_{n}| \mc{D}_{t}) + \frac{1}{n} \sum_{i=1}^{n} \int_{0}^{1}\nabla^{2} \ell(\theta_{s}| \mc{D}_{t,i}) (\what{\theta}_{n,t} - \theta^{*}_{n}) ds
\end{equation}
where $\theta_{s} = \theta\opt_{n} + s(\what{\theta}_{n,t} - \theta\opt_{n})$. We first focus our attention 
on the quantity $\sqrt{n} H_{n,t} \what{\theta}_{n,t}$ instead of $\what{\Psi}_{n,t} := \sqrt{n} \what{H}_{n,t} \what{\theta}_{n,t}$
, as the former is more amenable to analysis since $H_{n,t} \to H_{t}$ directly by the Law of Large Numbers. 
It follows that 
$\sqrt{n} H_{n,t} \what{\theta}_{n,t}$ can be written as
\begin{align}
  \sqrt{n} H_{n,t} \what{\theta}_{n,t} &= H_{t}\theta^{*}  + (H_{n,t} - H_{t})\theta^{*} -\frac{1}{\sqrt{n}} \sum_{i=1}^{n} \nabla_{\theta} \ell(\theta\opt_{n}| \mc{D}_{i}^{t}) \\
    &-  \sqrt{n}\int_{0}^{1}(\frac{1}{n} \sum_{i=1}^{ n } \nabla^{2} \ell(\theta_{s}| \mc{D}_{t,i}) - H_{n,t}) (\what{\theta}_{n,t}-\theta^{\ast}_{n}) ds.
\end{align}
 
\paragraph{Induction} We use an inductive argument to prove the weak
convergence~\eqref{eqn:weak-convergence} for a policy $\pi$ and reward process
$R_{n,t}$ satisfying Assumptions~\ref{assumption:policy}, ~\ref{assumption:m-estimation}, and ~\ref{assumption:hessian}. 

For $t = 0$, $p_{0}, \mu_{0}$ are fixed, implying that $\mc{D}_{i}^{0}$ are identically and independently 
distributed (i.i.d). Thus, $\nabla^{2}_{\theta} \ell(\theta\opt_{n}|\mc{D}_{i}^{0})$ and $\nabla_{\theta} \ell(\theta\opt_{n}|\mc{D}_{i}^{0})$ are i.i.d. 
Since $\nabla^{2}_{\theta} \ell(\theta\opt_{n}|\mc{D}_{i}^{0})$ is uniformly bounded by Assumption~\ref{item:boundedness}, 
$H_{n,0} \overset{a.s.}{\to} H_{0}$ by the Strong Law of Large Numbers. 
To bound the last term in the expansion \eqref{eqn:taylor-expansion}, 
we rely on the following proposition, which we prove in Section \ref{section:proof-bound-theta-n}. 
\begin{proposition} 
  \label{prop:bound-theta-n}
  Under condition (\ref{eqn:strong-convexity}) there exists a constant $m_{1} > 0$ that depends on $\mu, c_{1}, \text{ and } d$, such that
  \begin{equation}
    \label{eqn:bound-theta-n}
    \E[\|P_{t} (\what{\theta}_{n,t} - \theta^{*}_{n})\|^{p}_{2} ] \leq m_{1} n^{-p/2}
  \end{equation} 
\end{proposition}

\noindent By Assumption \ref{item:lipschitz} and Proposition \ref{prop:bound-theta-n}, we have 
\begin{align*}
  \E[\sqrt{n}\int_{0}^{1}(\frac{1}{n} \sum_{i=1}^{ n } \nabla^{2} \ell(\theta_{s}| \mc{D}_{i}^{0}) - H_{n,0}) (\what{\theta}_{n,0}-\theta^{\ast}_{n}) ds] 
  \leq c_{2} \sqrt{n} \E[||P_{0}(\what{\theta}_{n,0} - \theta^{*}_{n})||^{2}_{2}] \leq c_{2} m_{1}^{2} n^{-1/2}
\end{align*}
Therefore, we can apply the Lindeberg CLT to obtain
\[
  \sqrt{n} H_{n,0} \what{\theta}_{n,0} \cd N \left(H_{0} \theta^{*}, I_{0} \right).
\]
By Assumption \ref{assumption:hessian} and the Continuous Mapping Theorem,
\[
  \what{\Psi}_{n,0} \cd N \left(H_{0} \theta^{*}, I_{0} \right).
\]
Next, suppose
\begin{align}
  \label{eqn:inductive-hypothesis}
  \left(\what{\Psi}_{n,0},\ldots, \what{\Psi}_{n,t} \right) &\cd (G_{0},...,G_{t}).
\end{align}
To show weak convergence to $G_{t+1}$, it is sufficient to show 
\begin{equation}
  \label{eqn:inductive-weak-convergence}
  \E[f(\what{\Psi}_{n,0},\ldots,\what{\Psi}_{n,t+1})] \to
  \E[f(G_{0},...,G_{t+1})]~~\mbox{for any
bounded Lipschitz function}~f.
\end{equation}
Fixing a bounded Lipschitz $f$, assume without loss of generality that
\begin{equation*}
  \sup_{x,y\in\R^{(t+1)\times d}}|f(x)-f(y)|\leq 1
  ~~~\mbox{and}~~~f\in\text{Lip}(\R^{(t+1)\times d}).
\end{equation*}
We first set up some basic notation. For any
$\what{\Psi}_{0:t} \in \R^{(t+1) \times d}$, define the conditional expectation
operator on a random variable $W$
\begin{equation*}
  \E_{\what{\Psi}_{0:t}}[W] \defeq \E\left[ W \Bigg|\left\{ \what{\Psi}_{n,s} \right\}_{s=0}^t = \what{\Psi}_{0:t}\right].
\end{equation*}
Then, conditional on realizations of previous estimators up to time $t$, we
define a shorthand for the conditional expectation of $f$ and its limiting
counterpart.  Suppressing the dependence on $f$, let
$\what{g}_{n}, g_{n},g:\R^{t\times d} \to \R$ be
\begin{equation}
  \begin{aligned}
    \label{eqn:gs}
  \what{g}_{n}(\what{\Psi}_{0:t}) &\defeq
    \E_{\what{\Psi}_{0:t}}\left[f\left(\what{\Psi}_{n,0}, \ldots, \what{\Psi}_{n,t}, \what{\Psi}_{n,t+1} \right)\right]
    = \E\left[f\left(\what{\Psi}_{0:t}, \what{\Psi}_{n,t+1}\right)\right] \\
  g_{n}(\what{\Psi}_{0:t}) &\defeq
             \E_{\what{\Psi}_{0:t}}\left[f\left(\what{\Psi}_{n,0}, \ldots, \what{\Psi}_{n,t}, \sqrt{n}H_{n,t+1}\hat{\theta}_{n,t+1} \right)\right]
             = \E\left[f\left(\what{\Psi}_{0:t}, \sqrt{n}H_{n,t+1}\hat{\theta}_{n,t+1}\right)\right] \\
  g(\what{\Psi}_{0:t}) & \defeq \E_{\what{\Psi}_{0:t}}
            \left[f\left(\what{\Psi}_{n,0}, \ldots, \what{\Psi}_{n,t}, 
            I_{t+1}^{1/2}(\what{\Psi}_{0:t})Z + H_{t+1}(\what{\Psi}_{0:t})\theta^{*} \right)\right] \\
         & = \E\left[f\left(\what{\Psi}_{0:t}, I_{t+1}^{1/2}(\what{\Psi}_{0:t})Z + H_{t+1}(\what{\Psi}_{0:t})\theta^{*} \right)\right] 
\end{aligned}
\end{equation}
where $Z\sim N(0,I)$ and the conditional Hessian and Fisher information is determined by the
allocation $\pi_{t+1}(\what{\Psi}_{0:t})$, which depends on previous observations.  Conditional on the history $\what{\Psi}_{0:t}$, $\what{\Psi}_{n,t}$ depends on
$\what{\Psi}_{0:t}$ only through the sampling probabilities $\pi_{t+1}(\what{\Psi}_{0:t})$.

To show the weak convergence~\eqref{eqn:inductive-weak-convergence}, decompose the
difference between
$\E[f(\what{\Psi}_{n,0}, \ldots, \what{\Psi}_{n,t+1})]$ and
$\E[f(G_{0},\ldots,G_{t+1})]$ in terms of $\what{g}_{n}$, $g_{n}$ and $g$
\begin{align}
  &|\mathbb{E}[f(\what{\Psi}_{n,0}, \ldots, \what{\Psi}_{n, t+1})]
     - \mathbb{E}[f(G_{0},...,G_{t+1})]| \label{eqn:start} \\ 
  & =|\mathbb{E}[\what{g}_{n}(\what{\Psi}_{n,0}, \ldots, \what{\Psi}_{n,t})]
    - \mathbb{E}[g_{n}(\what{\Psi}_{n,0}, \ldots, \what{\Psi}_{n,t})] 
    + \mathbb{E}[g_{n}(\what{\Psi}_{n,0}, \ldots, \what{\Psi}_{n,t})]
    -\mathbb{E}[g(G_{0},...,G_{t})]|  \nonumber  \\
  & \leq|\mathbb{E}[\what{g}_{n}(\what{\Psi}_{n,0}, \ldots, \what{\Psi}_{n,t})]
    -\mathbb{E}[g_{n}(\what{\Psi}_{n,0}, \ldots, \what{\Psi}_{n,t})]|
    +|\E[g_{n}(\what{\Psi}_{n,0}, \ldots, \what{\Psi}_{n,t})]-\mathbb{E}[g(G_{0},...,G_{t})]|. \nonumber
\end{align}
First, for any bounded lipschitz function $f$, 
\begin{align*}
  \left|\E\left[f\left(\what{\Psi}_{0:t}, \what{\Psi}_{n,t+1}\right)\right] - \E\left[f\left(\what{\Psi}_{0:t}, \sqrt{n}H_{n,t+1}\hat{\theta}_{n,t+1}\right)\right]\right| 
  &\leq \E\left[ \left\| \what{\Psi}_{n,t+1} -  \sqrt{n}H_{n,t+1}\what{\theta}_{n,t+1}\right\|_{2}\right] \\
  &\leq \E\left[ \left\| \sqrt{n} (\what{H}_{n,t+1} - H_{n,t+1})\right\| \left\| \what{\theta}_{n,t+1}\right\|_{2}\right]
\end{align*}
By Assumption~\ref{assumption:hessian}, we have 
\begin{align}
    \label{eqn:empirical-g-bound}
    |\mathbb{E}[\what{g}_{n}(\what{\Psi}_{n,0}, \ldots, \what{\Psi}_{n,t})]
    -\mathbb{E}[g_{n}(\what{\Psi}_{n,0}, \ldots, \what{\Psi}_{n,t})]| \to 0.
\end{align}
We can further decompose as follows
\begin{align}
  |\E[g_{n}(\what{\Psi}_{n,0}, \ldots, \what{\Psi}_{n,t})]-\mathbb{E}[g(G_{0},...,G_{t})]| 
  &\leq |\E[g_{n}(\what{\Psi}_{n,0}, \ldots, \what{\Psi}_{n,t})]-\mathbb{E}[g(\what{\Psi}_{n,0}, \ldots, \what{\Psi}_{n,t})]| \\ 
  &+ |\E[g(\what{\Psi}_{n,0}, \ldots, \what{\Psi}_{n,t})]-\mathbb{E}[g(G_{0},...,G_{t})]|. \nonumber
\end{align}
By Assumption~\ref{assumption:policy}\ref{item:continuity} and dominated
convergence, $g$ is continuous almost surely under $(G_0, \ldots, G_{t})$.
From the inductive hypothesis~\eqref{eqn:inductive-hypothesis}, the continuous mapping theorem implies
\begin{equation}
\label{eqn:g-g-bound}
  \E[g(\what{\Psi}_{n,0}, \ldots, \what{\Psi}_{n,t})] \to \E[g(G_{1},...,G_{t})].
\end{equation}

\paragraph{Uniform convergence of $g_n \to g$}
It remains to show the convergence
\[
  \E[g_{n}(\what{\Psi}_{n,0},\ldots,\what{\Psi}_{n,t})] \to \E[g(\what{\Psi}_{n,0},\ldots,\what{\Psi}_{n,t})].
\]  While one would
expect pointwise convergence of $g_n(\what{\Psi}_{0:t})$ to $g(\what{\Psi}_{0:t})$ by the CLT,
proving the above requires controlling the convergence across random
realizations of the sampling probabilities
$\pi_{t+1}(\what{\Psi}_{n,0},\ldots,\what{\Psi}_{n,t})$.  This is
complicated by the fact that we allow the sampling probabilities to be zero.  To
this end, we use the Stein's method for multivariate distributions to provide
rates of convergence for the CLT that are uniform across realizations
$\what{\Psi}_{0:t}$. We use the bounded Lipschitz distance $d_{\text{BL}}$ to metrize
weak convergence.
\begin{equation*}
  d_{\text{BL}}(\mu, \nu) \defeq \sup\left\{
    |\mathbb{E}_{R \sim \mu}\bar{f}(R)-\mathbb{E}_{R \sim \nu}\bar{f}(R)|:
    \bar{f}\in\text{Lip}(\mathbb{R}^{\numarm}),
    \sup_{x,y\in\mathbb{R}^{\numarm}}|\bar{f}(x)-\bar{f}(y)|\leq1\right\}.
\end{equation*}
Define 
\[
  b_{n, t+1}(\what{\Psi}_{0:t}) = H_{t+1}(\what{\Psi}_{0:t})\theta^{*} 
                          - \sqrt{n}H_{n,t+1}\hat{\theta}_{n,t+1} 
                          - \frac{1}{\sqrt{n}} \sum_{i=1}^{n} \nabla_{\theta} \ell(\theta\opt_{n}|\mc{D}_{i}^{t+1}).
\] 
To bound $|\E[g_{n}(\what{\Psi}_{n,0},\ldots,\what{\Psi}_{n,t})] - g(\what{\Psi}_{n,0},\ldots,\what{\Psi}_{n,t})|$,
decompose
\begin{align}
  & |\E[g_{n}(\what{\Psi}_{n,0},\ldots,\what{\Psi}_{n,t}) - g(\what{\Psi}_{n,0},\ldots,\what{\Psi}_{n,t})]| \nonumber  \\
  & = |\E[\E_{\what{\Psi}_{0:t}}[f(\what{\Psi}_{0:t}, \sqrt{n}H_{n,t+1}\hat{\theta}_{n,t+1})
    - f(\what{\Psi}_{0:t},
    I_{t+1}^{1/2}(\what{\Psi}_{0:t})Z + H_{t+1}(\what{\Psi}_{0:t})\theta^{*})]]|
    \nonumber \\
    & \leq \E[\E_{\what{\Psi}_{0:t}}[d_{\text{BL}}(\frac{1}{\sqrt{n}} \sum_{i=1}^{ n} \nabla_{\theta} \ell(\theta\opt_{n}|\mc{D}_{i}^{t+1}),
    I_{t+1}^{1/2}(\what{\Psi}_{0:t})Z + b_{n,t+1}(\what{\Psi}_{0:t}))]] \nonumber \\
  & \leq \E[\E_{\what{\Psi}_{0:t}}[d_{\text{BL}}(\frac{1}{\sqrt{n}} \sum_{i=1}^{ n} \nabla_{\theta} \ell(\theta\opt_{n}|\mc{D}_{i}^{t+1}),
  I_{t+1}^{1/2}(\what{\Psi}_{0:t})Z)]
    + \E_{\what{\Psi}_{0:t}}[d_{\text{BL}}(
      I_{t+1}^{1/2}(\what{\Psi}_{0:t})Z,
      I_{t+1}^{1/2}(\what{\Psi}_{0:t})Z + b_{n,t+1}(\what{\Psi}_{0:t}))]]
\label{eqn:bl-triangle}
\end{align}
We now proceed to bound both terms in the inequality~\eqref{eqn:bl-triangle}. The first term measures the 
difference between a centered $\sqrt{n}H_{n,t+1}\what{\theta}_{n,t+1}$ and its Gaussian limit, and can 
be bounded by noting that 
\begin{align}
  \label{eqn:cov-n}
  \text{Cov}_{\what{\Psi}_{0:t}}
  \left(\frac{1}{\sqrt{n}}\nabla_{\theta} \ell(\theta\opt_{n}|\mc{D}_{i}^{t+1})\right)
  &= I_{t+1}(\what{\Psi}_{0:t}). \nonumber
\end{align}
The second term accounts purely for the bias incurred compared to its true limit. 
To ease notation, we often omit the dependence on $\what{\Psi}_{0:t}$. 

\paragraph{Bounding the first term in inequality~\eqref{eqn:bl-triangle}}
Before we proceed, it is helpful to define the following quantities, which describe
smoothness of the derivatives of any function $f\in\mathcal{C}^{3}$
\[
  M_{1}(f) = \sup_{x} \norm{\nabla f}_{2} 
  ~~~~M_{2}(f) = \sup_{x} \norm{\nabla^{2}f}_{op} 
  ~~~~M_{3}(f) = \sup_{x} \norm{\nabla^{3}f}_{op}
\]
The following bound, which we prove in
Section~\ref{section:proof-clt-sample-mean}, quantifies the rate of
convergence for the CLT using the multivariate Stein's
method~\citet{Meckes2009stein}. 
\begin{proposition}
  \label{prop:clt-sample-mean}
  For any $f \in \mathcal{C}^{3}$,
  \[
    |\E f(\frac{1}{\sqrt{n}} \sum_{i=1}^{ n} \nabla_{\theta} \ell(\theta\opt_{n}|\mc{D}_{i}^{t+1})) - \E f(I_{t+1}^{1/2} Z) |
    \leq 
    C_{1} n^{-1/2} M_{2}(f) + C_{2}n^{-1/2} M_{3}(f)
  \]
  where $C_{1}$ and $C_{2}$ depend polynomially on $d, c_{3}$.
\end{proposition}

It remains to show convergence of $\frac{1}{\sqrt{n}} \sum_{i=1}^{ n} \ell(\theta\opt_{n}|\mc{D}_{i}^{t+1})$ to $I_{t+1}^{1/2}Z$ for
Lipschitz test functions, which is required to control the bounded Lipschitz
distance.  We use standard Gaussian smoothing arguments found in
\citet{Meckes2009gauss}: by convolving the test function with a Gaussian
density, one obtains a smoother function for which the result of
Proposition~\ref{prop:clt-sample-mean} is applicable. At the same time, the
amount of smoothing is controlled to ensure the bias with the original test
function is small.

\begin{lemma}[{\citet[Corollary 3.5]{Meckes2009gauss}}]
\label{lemma:convolution}
For any 1-Lipschitz function $f$, consider the Gaussian convolution 
$(f * \phi_{\delta})(x) := \E[f(x + \delta Z)]$, where $Z\sim N(0,I_{d})$.
\[
  \begin{aligned}
  M_{2}(f * \phi_{\delta}) 
  &\leq M_{1}(f) \sup_{\theta:\norm{\theta}_{2} = 1} \norm{\nabla \phi_{\delta}^{\top} \theta}
  \leq \sqrt{\frac{2}{\pi}} \frac{1}{\delta} \\
  M_{3}(f * \phi_{\delta}) 
  &\leq M_{1}(f) \sup_{\theta:\norm{\theta}_{2} = 1} \norm{\theta^{\top} \nabla^{2} \phi_{\delta} \theta}
  \leq \frac{\sqrt{2}}{\delta^{2}}
  \end{aligned}
\]
Moreover, for any random vector $X\in \R^{d}$,
$$ \E[(f * \phi_{\delta})(X) - f(X)] \leq \E[\delta\norm{Z}_{2}] \leq \delta \sqrt{d} $$
\end{lemma}

Thus, for any 1-Lipschitz function $f$, we have the bound
\[
  \begin{aligned}
    |\E f(\frac{1}{\sqrt{n}} \sum_{i=1}^{ n} \ell(\theta\opt_{n}|\mc{D}_{i}^{t+1})) - \E f(I_{t+1}^{1/2} Z) |
    &\leq |\E f(\frac{1}{\sqrt{n}} \sum_{i=1}^{ n} \ell(\theta\opt_{n}|\mc{D}_{i}^{t+1})) - \E (f * \phi_{\delta})(\frac{1}{\sqrt{n}} \sum_{i=1}^{ n} \ell(\theta\opt_{n}|\mc{D}_{i}^{t+1})) | \\
    &\qquad+ |\E (f * \phi_{\delta})(\frac{1}{\sqrt{n}} \sum_{i=1}^{ n} \ell(\theta\opt_{n}|\mc{D}_{i}^{t+1})) - \E (f * \phi_{\delta})(I_{t+1}^{1/2} Z) | \\
    &\qquad + |\E (f * \phi_{\delta})(I_{t+1}^{1/2} Z)
    - \E f(I_{t+1}^{1/2} Z)| \\
    &\leq 2\delta \sqrt{d}
       +  \sqrt{\frac{2}{\pi}} \frac{1}{\delta} C_{1} n^{-1/2}
        + \frac{\sqrt{2}}{\delta^{2}} C_{2} n^{-1/2}
  \end{aligned}
\]
Optimizing over $\delta$, we obtain
\[
  d_{\text{BL}}(\frac{1}{\sqrt{n}} \sum_{i=1}^{ n} \ell(\theta\opt_{n}|\mc{D}_{i}^{t+1}), I_{t+1}^{1/2} Z)
  \leq C_{t+1} n^{-1/6}
\]
for some constant $C_{t+1}$ that depends only on 
$d$ and $c_{3}$ but not on $n$ or $\what{\Psi}_{0:t}$.  

\paragraph{Bounding the second term in inequality~\eqref{eqn:bl-triangle}}
Finally, it remains to show uniform convergence of the second term in the
bound~\eqref{eqn:bl-triangle}.
\begin{equation*}
  \E\left[\E_{\what{\Psi}_{0:t}}\left[d_{\text{BL}}\left(I_{t+1}^{1/2}(\what{\Psi}_{0:t})Z,
  I_{t+1}^{1/2}(\what{\Psi}_{0:t})Z + b_{n,t+1}(\what{\Psi}_{0:t})\right) \right]\right].
\end{equation*}
Since the stochastic terms have identical distributions, by a simple coupling argument, for any bounded Lipschitz function $f$
\begin{align}
  \label{eqn:bound-bias}
& \E_{\what{\Psi}_{0:t}}\left[ \E[f\left(I_{t+1}^{1/2}(\what{\Psi}_{0:t})Z \right)] - 
        \E[ f\left(I_{t+1}^{1/2}(\what{\Psi}_{0:t})Z + b_{n,t+1}(\what{\Psi}_{0:t}) \right)] \ \right] \nonumber  \\ 
&\leq \E_{\what{\Psi}_{0:t}}  \left[ \E \left[\norm{I_{t+1}^{1/2}(\what{\Psi}_{0:t})Z - I_{t+1}^{1/2}(\what{\Psi}_{0:t})Z - b_{n,t+1}(\what{\Psi}_{0:t})}_{2} \right] \right] \nonumber\\
&= \E_{\what{\Psi}_{0:t}} \left[ \E \left[\norm{b_{n,t+1}(\what{\Psi}_{0:t})}_{2} \right] \right] \nonumber \\
&= \E_{\what{\Psi}_{0:t}}  \left[ \E \left[ \norm{H_{t+1}(\what{\Psi}_{0:t})\theta\opt 
- \sqrt{n}H_{n,t+1}\hat{\theta}_{n,t+1} 
- \frac{1}{\sqrt{n}} \sum_{i=1}^{n} \nabla_{\theta} \ell(\theta\opt_{n}|\mc{D}_{i}^{t+1})}_{2} \right] \right] \nonumber \\
&= \E_{\what{\Psi}_{0:t}}  \left[ \E \left[ \norm{(H_{t+1} - H_{n,t+1})\theta\opt 
+ \sqrt{n}\int_{0}^{1}(\frac{1}{n} \sum_{i=1}^{ n } \nabla^{2} \ell(\theta_{s}| \mc{D}_{t,i}) 
- H_{n,t}) (\what{\theta}_{n,t}-\theta^{\ast}_{n}) ds}_{2} \right] \right] \nonumber\\
& \leq \E_{\what{\Psi}_{0:t}}  \left[ \E \left[ \norm{(H_{t+1} - H_{n,t+1})}_{2}\norm{\theta\opt}_{2} 
+ \norm{\sqrt{n}\int_{0}^{1}(\frac{1}{n} \sum_{i=1}^{ n } \nabla^{2} \ell(\theta_{s}| \mc{D}_{t,i}) 
- H_{n,t}) (\what{\theta}_{n,t}-\theta^{\ast}_{n}) ds}_{2} \right] \right] 
\end{align}
By Assumption \ref{item:lipschitz} and Proposition \ref{prop:bound-theta-n}, we can bound the 
second term in~\eqref{eqn:bound-bias} as
\begin{align}
\label{eqn:taylor-bias}
  \E[\sqrt{n}\int_{0}^{1}(\frac{1}{n} \sum_{i=1}^{ n } \nabla^{2} \ell(\theta_{s}| \mc{D}_{i}^{0}) - H_{n,t+1}) (\what{\theta}_{n,t+1}-\theta^{\ast}_{n}) ds] 
  \leq c_{2} \sqrt{n} \E[||P_{t+1}(\what{\theta}_{n,t+1} - \theta^{*}_{n})||^{2}_{2}] \leq c_{2} m_{1}^{2} n^{-1/2}.
\end{align}
Next, we bound the expected spectral norm $\E \left[\norm{ H_{n, t+1} - H_{t+1} }_{2}\right]$.
Conditional on $\what{\Psi}_{0:t}$,  
$H_{n,t+1}(\what{\Psi}_{0:t}) - H_{t+1}(\what{\Psi}_{0:t}) = \sum_{i=1}^{n} \frac{\varphi_{n,t+1,i}}{n}$ is 
a a sum of iid, mean-zero, random matrices where
\[
  \varphi_{n,t+1,i} = \left( \nabla_{\theta}^{2} \ell(\theta\opt_{n}|\mc{D}_{i}^{t+1}) - \E_{\what{\Psi}_{0:t}}[\nabla_{\theta}^{2} \ell(\theta\opt_{n}|\mc{D}_{i}^{t+1})] \right)
\]
By Assumption~\ref{item:boundedness}, $\norm{\frac{\varphi_{n,t+1,i}}{n}} \leq 2c_{4}/n$.
Moreover,
\[
\lambda_{\max}\left(\sum_{i=1}^{n} \E \left(\frac{\varphi_{n,t+1,i}}{n}\right)^{2} \right)  
= \frac{1}{n}\lambda_{\max}\left( \frac{1}{n}\sum_{i=1}^{n} \E\varphi_{n,t+1,i}^{2} \right)
\leq \frac{1}{n} \lambda_{\max}\left( \E\varphi_{n,t+1,i}^{2} \right) \leq c_{4}^{2} / n
\]
By the matrix Bernstein inequality we have
\begin{align}
\label{eqn: bernstein-bias}
  \E_{\what{\Psi}_{0:t}} \left[ \E \left[\norm{ H_{t+1}(\what{\Psi}_{0:t}) - H_{n,t+1}(\what{\Psi}_{0:t})}_{2} \right] \right] \leq \sqrt{\frac{c_{4}^{2} \log d}{n}} + \frac{2c_{4} \log d}{n}
\end{align}
Using results from~\eqref{eqn:taylor-bias},~\eqref{eqn: bernstein-bias} and by collecting constants we have 
\begin{equation*}
  \E\left[\E_{\what{\Psi}_{0:t}}\left[d_{\text{BL}}\left(I_{t+1}^{1/2}(\what{\Psi}_{0:t})Z,
  I_{t+1}^{1/2}(\what{\Psi}_{0:t})Z + b_{n,t+1}(\what{\Psi}_{0:t})\right) \right]\right] \leq D_{t+1}n^{-1/2},
\end{equation*}
where $D_{t+1}>0$ is a constant that depends on $||\theta^{*}||_{2}, c_{2}, c_{4}$, and $m_{1}$.

\paragraph{Conclusion} Altogether, we have shown
\[
  |\E[g_{n}(\what{\Psi}_{n,0},\ldots,\what{\Psi}_{n,t}) -
  g(\what{\Psi}_{n,0},\ldots,\what{\Psi}_{n,t})]| \leq C_{t+1}n^{-1/6} + D_{t+1}n^{-1/2} 
\]
for constants $C_{t+1}, D_{t+1}$ that depend polynomially on $c_{1}, c_{2}, c_{3}, c_{4}$ as well as on $d$.
Alongside~\eqref{eqn:empirical-g-bound} and~\eqref{eqn:g-g-bound}, this shows that
$\E[f(\what{\Psi}_{n,0},\ldots,\what{\Psi}_{n,t+1})] \to \E[f(G_{0},\ldots,G_{t+1})]$ as
$n\to \infty$ for any bounded Lipschitz function $f$, which implies the
desired weak convergence.

\subsubsection{Proof of Proposition~\ref{prop:clt-sample-mean}}
\label{section:proof-clt-sample-mean}

In order to quantify the rate of the CLT, we use the following result by
\citet{Meckes2009stein} which provides a characterization of Stein's method
for random vectors with arbitrary covariance matrices.
\begin{lemma}[{\citet[Theorem 3]{Meckes2009stein}}]
  \label{theorem:meckes}
  Let $(W, W')$ be an exchangeable pair of random vectors in $\R^{d}$. 
  Suppose that there exists $\lambda > 0$, a positive semi-definite matrix $\Sigma$,
  a random matrix $E$ such that
  \[
    \begin{aligned}
      \E[W' - W | W] &= -\lambda W \\
      \E[(W' - W)(W' - W)^{\top}|W] &= 2\lambda \Sigma + \E[E|W]
    \end{aligned}
  \]
  Then for any $f \in \mathcal{C}^{3}$,
  \[
    |\E f(W) - \E f(\Sigma^{1/2} Z) |
    \leq \frac{1}{\lambda}
    \left[ 
        \frac{\sqrt{d}}{4} M_{2}(f) \E \norm{E}_{H.S.}
        + \frac{1}{9} M_{3}(f) \E \norm{W' - W}^{3}
    \right]
  \]
where $\norm{\cdot}_{H.S.}$ is the Hilbert-Schmidt norm.
\end{lemma}

The rest of the proof is similar to that of~\citet[Theorem 7]{ChatterjeeMe08},
with slight modifications due to the fact that we have a non-identity
covariance. For simplicity, define 
$W = \frac{1}{\sqrt{n}} \sum_{i=1}^{n} \nabla \ell_{\theta_{n}^{*}} (\mc{D}_{i}^{t+1})$ and 
$X_{j} :=  \frac{\nabla \ell_{\theta_{n}^{*}} (\mc{D}_{j}^{t+1})}{\sqrt{n}}$.
For any index $j$, we construct an independent copy $Y_j$ of $X_j$.  We
construct an exchangeable pair $(W,W')$ by selecting an random
index $I\in \{1,...,n\}$ chosen uniformly and independently from
$W$ and letting
$$ W' = W - \frac{X_{I}}{\sqrt{n}} + \frac{Y_{I}}{\sqrt{n}}. $$

We can observe then that
\[
  \begin{aligned}
    \E [W' - W|W]
    &= \frac{1}{\sqrt{n}} \E[Y_{I} - X_{I}|W] \\
    &= \frac{1}{n^{3/2}} \sum_{j=1}^{n} \E[Y_{j} - X_{j}|W] \\
    &= -\frac{1}{n} W
  \end{aligned}
\]
by independence of $Y_j$ and $W$. This pair satisfies the first condition of 
Theorem~\ref{theorem:meckes} with $\lambda = 1/n$.

It also satisfies the second condition of Theorem~\ref{theorem:meckes} with:
\begin{align*}
  E_{a,a'} 
& = \frac{1}{n^{2}} \sum_{j=1}^{n} \left(X_{j,a}X_{j,a'} - (I_{t+1} )_{a,a'} \right)
\end{align*}
by independence of $X_{i}$ and $Y_{i}$.

Thus, we have that
\[
  \begin{aligned}
    \E E_{a,a'}^{2} &= \frac{1}{n^{4}} \sum_{j=1}^{n} \E \left( \E[X_{j,a}X_{j,a'} - (I_{t+1} )_{a,a'}|W] \right)^{2} \\
    &\leq  \frac{1}{n^{4}} \sum_{j=1}^{n} \E\left[ \left( X_{j,a}X_{j,a'} - (I_{t+1} )_{a,a'} \right)^{2} \right]  \\
    &= \frac{1}{n^{3}} \E\left[ \left( X_{1,a}X_{1,a'}  \right)^{2} - (I_{t+1} )_{a,a'}^{2}\right]  \\
  \end{aligned}
\]
This gives us a bound on the Hilbert-Schmidt norm:
\[
  \begin{aligned}
    \E \norm{E}_{H.S.} &\leq \sqrt{\sum_{a,a'} \E E_{a,a'}^{2}} 
    \leq \frac{1}{n^{3/2}} \sqrt{\sum_{a,a'}\E \left( X_{1,a}X_{1,a'}  \right)^{2} - (I_{t+1} )_{a,a'}^{2}}
    \leq \frac{1}{n^{3/2}} \sqrt{\E\norm{X_{1}}_{2}^{4}}
  \end{aligned}
\]
We can further bound $\E\norm{X_{1}}_{2}^{4}$ by
a constant $C_{1}$ that is polynomial in $c_{3}$.

Finally, we can
bound $\E \norm{W' - W}_{2}^{3}$ as follows:
\[
  \begin{aligned}
    \E \norm{W' - W}_{2}^{3}
    =  \frac{1}{n^{3/2}} \E \norm{Y_{I} - X_{I}}_{2}^{3}
    \leq \frac{1}{n^{3/2}} 8\E \norm{X_{1}}_{2}^{3}
  \end{aligned}
\]
where the final inequality uses independence of $Y_{j}$ and $X_{j}$ as well as Holder's inequality.
We can bound $\E \norm{X_{1}}_{2}^{3}$ by another constant $C_{2}$ that is polynomial in $c_{3}$. 
Plugging these bounds into the statement of Theorem~\ref{theorem:meckes}, we obtain the stated result.

\subsection{Proof of Proposition~\ref{prop:bound-theta-n}}
\label{section:proof-bound-theta-n}

Our proof follows that of~\cite[Lemma 5.2]{Shao2021Berry} with modifications because we only assume strong convexity with 
  respect to the set of parameters projected into the range space of $P_{t}$. 
  Given a sampling decision $p_{t}$, for each $n$, let
  \[
  A_{j,n} = \{\theta : 2^{j-1} < \sqrt{n}\|P_{t}(\theta - \theta\opt_{n})\|_{2} \leq 2^j\}, \quad j \geq 1.
  \]
  First, by strong convexity in the projected parameter space~\eqref{item:strong-convexity}, we have 
  \[
  \inf_{\theta \in A_{j,n}} \E_{t}[\ell (\theta | \mathcal{D}_{i}^{t})] - \E_{t}[\ell (\theta_{n}\opt|\mathcal{D}_{i}^{t})] \geq \mu \inf_{\theta \in A_{j,n}} \|P_{t}(\theta - \theta\opt_{n})\|^{2}_{2} \geq \mu n^{-1} 4^{j}.
  \]
  Define 
  \[
    M(p_{t}, \theta, \delta) = \sup_{\theta: \|P_{t} (\theta - \theta\opt_{n})\| \leq \delta} 
  \sqrt{n} \left| \left(\frac{1}{n}\sum_{i=1}^{ n}  \ell (\theta | \mathcal{D}_{i}^{t}) - \E_{t}[\ell (\theta | \mathcal{D}_{i}^{t})]\right) - \left(\frac{1}{n}\sum_{i=1}^{ n}  \ell (\theta_{n}\opt | \mathcal{D}_{i}^{t}) - \E_{t}[ \ell (\theta_{n}\opt | \mathcal{D}_{i}^{t})]\right)  \right|,
  \] 
  and let $\delta_{j} = 2^{j} n^{-1/2}$.
  Since $\what{\theta}_{n,t}$ minimizes the empirical loss $L_{n}(\theta| \mc{D}_{t}) := \frac{1}{n}\sum_{i=1}^{ n}  \ell (\theta | \mathcal{D}_{i}^{t})~\eqref{eqn:erm_prelimit}$ by Assumption~\eqref{item:optimality}, then
  \begin{align}
  \label{eqn: consistency-series}
      \nonumber \mathbb{E}_{t} [\sqrt{n} \|P_{t} (\what{\theta}_{n,t} - \theta\opt_{n})\|_{2} ]^{p} 
  \nonumber &\leq \sum_{j \geq 1} 2^{jp} \mathbb{P} (\what{\theta}_{n,t} \in A_{j,n}) \\ 
  \nonumber &\leq \sum_{j \geq 1} 2^{jp} \mathbb{P} \left( \inf_{\theta \in A_{j,n}} L_{n}(\theta | \mc{D}_{t}) - L_{n}(\theta\opt_{n} | \mc{D}_{t}) \leq 0 \right) \\
  \nonumber &\leq \sum_{j \geq 1} 2^{jp} \mathbb{P} \left(\inf_{\theta \in A_{j,n}} \E_{t}[\ell (\theta | \mathcal{D}_{i}^{t})]- \E_{t}[\ell (\theta_{n}\opt|\mathcal{D}_{i}^{t})] - \inf_{\theta \in A_{j,n}} L_{n}(\theta| \mc{D}_{t}) + L_{n}(\theta\opt_{n}| \mc{D}_{t}) \geq  \mu n^{-1}4^{j} \right) \\
   &\leq \sum_{j \geq 1} 2^{jp} \mathbb{P} \left( M(p_{t}, \theta, \delta_{j}) \geq \mu n^{-1/2} 4^{j}  \right). \\ \nonumber 
  \end{align}
  By~\cite[Lemma A.2]{Shao2021Berry}, we can bound
  \begin{equation}
    \label{eqn: sup-process-lemma}
    M(p_{t}, \theta, \delta) \leq C c_{1} \sqrt{d} \delta, 
  \end{equation}
  where $C > 0$ is an absolute constant.  Then, by substituting $\delta_{j}$ and an application of both Markov's Inequality and~\eqref{eqn: sup-process-lemma}, we can bound the series~\eqref{eqn: consistency-series} 
  \begin{align*}
    \sum_{j \geq 1} 2^{jp} \mathbb{P} \left( M(p_{t}, \theta, \delta_{j}) \geq \mu n^{-1/2} 4^{j}  \right) 
  &\leq \mu^{-p} n^{p/2}\sum_{j \geq 1} 2^{-jp} \E [M(p_{t}, \theta, \delta_{j})^{p}]. \\
  &\leq m_{1},
  \end{align*}
where $m_{1}>0$ is a constant that depends polynomially on $C, c_{1}, d,$ and $\mu$. Therefore, we obtain
\[
    \mathbb{E}_{t} [ \|P_{t} (\what{\theta}_{n,t} - \theta\opt_{n})\|_{2} ]^{p} \leq m_{1}n^{-p/2}.
\]



\section{Derivations for Bayesian adaptive experiment}
\label{section:proof-bae}
\subsection{Proof of Lemma~\ref{lemma:mdp}}
\label{section:proof-mdp}

Let $\Gamma_{t} := n_{t}H_{t}I_{t}^{-1}H_{t}$. Then,
We can simplify the Bayesian posterior update for the mean as follows:

\begin{align*}
 & \Sigma_{t+1}(\Sigma_{t}^{-1}\beta_{t}+\Gamma_{t}\beta_{t}-\Gamma_{t}\beta_{t}+\Gamma_{t}\hat{\beta}_{t})\\
 & =\beta_{t}+\Sigma_{t+1}\Gamma_{t}(\hat{\beta}_{t}-\beta_{t})
\end{align*}

Note that the posterior predictive distribution of $\hat{\beta}_{t}\sim N(\beta^{*},\Gamma_{t}^{-1})$
is given by 

\begin{align*}
\hat{\beta}_{t} & =\beta^{*}+s\Gamma_{t}^{-1/2}Z_{1}\\
 & =(\beta_{t}+\Sigma_{t}^{1/2}Z_{2})+s\Gamma_{t}^{-1/2}Z_{1}
\end{align*}
where $Z_{1}$ and $Z_{2}$ are independent $N(0,I_{d})$ random vectors.
This implies
\begin{align*}
\text{Var}\left(\Gamma_{t}(\hat{\beta}_{t}-\beta_{t})\right) & =\text{Var}\left(\Gamma_{t}(\Sigma_{t}^{1/2}Z_{2}+s\Gamma_{t}^{-1/2}Z_{1})\right)\\
 & =\left(\Gamma_{t}\Sigma_{t}\Gamma_{t}\right)+\Gamma_{t}\left(n_{t}^{-1}\Gamma_{t}^{-1}\right)\Gamma_{t}\\
 & =\left(\Gamma_{t}\Sigma_{t}\Gamma_{t}\right)+\Gamma_{t}
\end{align*}

To compute $\text{Var}\left(\Sigma_{t+1}\Gamma_{t}(\hat{\beta}_{t}-\beta_{t})\right)$,
we observe that

\begin{align*}
\text{Var}\left(\Sigma_{t+1}\Gamma_{t}(\hat{\beta}_{t}-\beta_{t})\right) & =\Sigma_{t+1}\left(\left(\Gamma_{t}\Sigma_{t}\Gamma_{t}\right)+\Gamma_{t}\right)\Sigma_{t+1}\\
 & =\Sigma_{t+1}\left(\Gamma_{t}\Sigma_{t}+I\right)\Gamma_{t}\Sigma_{t+1}
\end{align*}

We use the identity $(A+B)^{-1}=A^{-1}-(A+AB^{-1}A)^{-1}$, taking
$A=\Gamma_{t}$ and $B=\Sigma_{t}^{-1}$ and observe that

\begin{align*}
\Sigma_{t+1} & =\Gamma_{t}^{-1}-\left[\left(\Gamma_{t}\Sigma_{t}\Gamma_{t}\right)+\Gamma_{t}\right]^{-1}\\
 & =A^{-1}-\left[(AB^{-1}+I)A\right]^{-1}
\end{align*}
Further simplifying, we have
\begin{align*}
 & \Sigma_{t+1}\left(\left(\Gamma_{t}\Sigma_{t}\Gamma_{t}\right)+\Gamma_{t}\right)\Sigma_{t+1}\\
 & =\left(A^{-1}-\left[(AB^{-1}+I)A\right]^{-1}\right)\left((AB^{-1}+I)A\right)\left(A^{-1}-\left[(AB^{-1}+I)A\right]^{-1}\right)\\
 & =\left(A^{-1}-\left[(AB^{-1}+I)A\right]^{-1}\right)\left((AB^{-1}+I)-I\right)\\
 & =\left(A^{-1}-\left[(AB^{-1}+I)A\right]^{-1}\right)AB^{-1}\\
 & =\Sigma_{t+1}\Gamma_{t}\Sigma_{t}
\end{align*}
Finally, we can observe that $\Sigma_{t+1}\Gamma_{t}\Sigma_{t}$ is equal to $\Sigma_{t} - \Sigma_{t+1}$.
Since, $\Sigma_{t+1}\Gamma_{t}(\hat{\beta}_{t}-\beta_{t})$
is a mean-zero Gaussian random vector, it can be expressed as $\text{Var}\left(\Sigma_{t+1}\Gamma_{t}(\hat{\beta}_{t}-\beta_{t})\right)^{1/2}Z$.
So altogether the posterior update can be expressed as 
\[
\left(\Sigma_{t} - \Sigma_{t+1}\right)^{1/2}Z_{t}
\]

\subsection{Proof of Corollary~\ref{cor:trajectory_limit}}
\label{section:proof-bayes-limit}

Note that the posterior state~\eqref{eqn:approx_posterior} transitions are continuous 
functions of previous states $(\beta_{n,t}, \Sigma_{n,t})$, $\what{\Psi}_{n,t}$, $\what{H}_{n,t}$ and $\what{I}_{n,t}$.
Therefore, they are continuous functions of $\what{\Psi}_{n,0},\ldots, \what{\Psi}_{n,T-1}$ 
and 
$\what{H}_{n,0}, \what{I}_{n,0},\ldots, \what{H}_{n,t}, \what{I}_{n,t}$. 
As $n\to\infty$, the above quantities converge in distribution to $G_{0},\ldots,G_{T-1}$
and $H_{0}I_{0}^{\dagger}H_{0},\ldots H_{T-1}I_{T-1}^{\dagger}H_{T-1}$.
The continuous mapping theorem implies that 
\[
  (\beta_{n,0}, \Sigma_{n,0},\ldots,\beta_{n,T-1},\Sigma_{n,T-1}) \cd (\beta_{0}, \Sigma_{0},\ldots,\beta_{T-1},\Sigma_{T-1})
\]

\subsection{Proof of Corollary~\ref{cor:value_limit}}
\label{section:proof-bayes-limit}

Since $c_{s}$ are continuous functions of $(\beta_{n,t}, \Sigma_{n,t})$, by the continous mapping theorem 
$c_{s}(\pi_{s}, \beta_{n,s},\Sigma_{n,s}) \cd c_{s}(\pi_{s}, \beta_{s},\Sigma_{s})$.
Note that almost surely, we have that $\norm{\Sigma_{n,s}}_{2} \leq \norm{\Sigma_{0}}_{2}$
and $\norm{\beta_{n,s}} \leq \norm{\beta_{0}}_{2} + \sum_{t=1}^{T} \norm{\what{\Psi}_{n,t}}_{2}$. 
Note that by definition of $\Psi_{n,t}$,  $\E[\norm{\what{\Psi}_{n,t}}_{2}^{2}] \leq c_{4}\theta^{*} < \infty $ has a uniformly bounded second moment across $n$. Thus, $\beta_{n,t}$ and $\Sigma_{n,t}$ and $c$ are uniformly integrable and by dominated convergence theorem,
the expectation of $c_{s}$ converge.



\subsection{Proof of Theorem~\ref{theorem:policy_improvement}}
\label{section:proof-policy-improvement}

First observe that when at $T$ and $T-1$, the static policy and
RHO coincide so $V_{t}^{\algo}(\beta_{t},\Sigma_{t})=\max_{p}V_{t}^{p}(\beta_{t},\Sigma_{t})$ for any budget $B_{t}$.
Next, as an induction hypothesis, suppose for all $(\beta_{t+1},\Sigma_{t+1}, B_{t+1})$,

\[
V_{t+1}^{\algo}(\beta_{t+1},\Sigma_{t+1}, B_{t+1})
\leq 
\min_{p_{t+1:T}} \left\{  V_{t+1}^{p_{t+1:T}}(\beta_{t+1},\Sigma_{t+1}) ~~\Big|~~    \sum\nolimits_{s=t+1}^{T}g_{s}(p_{s})  \leq B_{t+1}
            \right\}.
\]

For the purpose of clarity, we will explicitly denote $B_{t}$ as part of the state.
Then for any $(\beta_{t},\Sigma_{t}, B_{t})$,

\begin{align*}
V_{t}^{\algo}(\beta_{t},\Sigma_{t}, B_{t}) 
& = c_{t}(\rho_{t}^{*}|\beta_{t},\Sigma_{t})
+\mathbb{E}_{t}[V_{t+1}^{\algo}(\beta_{t+1},\Sigma_{t+1}, 
B_{t} - g_{t}(\rho^{*}_{t})]\\
 & \leq c_{t}(\rho_{t}^{*}|\beta_{t},\Sigma_{t}) +\mathbb{E}_{t}\left[ 
 \min_{p_{t+1:T}} \left\{V_{t+1}^{p_{t+1:T}}(\beta_{t+1},\Sigma_{t+1}, ,B_{t}
 - g_{t}(\rho^{*}_{t}))  
 ~~\Big|~~
  \sum\nolimits_{s=t+1}^{T}g_{s}(p_{s})  \leq B_{t}
 - g_{t}(\rho^{*}_{t}))
 \right\}.
 \right] \\
 & \leq c_{t}(\rho_{t}^{*} \, | \, \beta_{t},\Sigma_{t}) \\
 &+\min_{p_{t+1:T}}
 \mathbb{E}_{t}
 \left\{
 \left[
 V_{t+1}^{p_{t+1:T}}(\beta_{t+1},\Sigma_{t+1},B_{t}
 - g_{t}(\rho^{*}_{t})))
 \right]
  ~~\Big|~~
  \sum\nolimits_{s=t+1}^{T}g_{s}(p_{s})  \leq B_{t}
 - g_{t}(\rho^{*}_{t})) \right\}\\
 & =\min_{p_{t:T}}\left\{ c_{t}(p_{t} \, | \,
 \beta_{t},\Sigma_{t})+\mathbb{E}_{t}[V_{t+1}^{p_{t+1:T}}(\beta_{t+1},\Sigma_{t+1},B_{t}
 - g_{t}(\rho^{*}_{t})))]
 ~~\Big|~~
  \sum\nolimits_{s=t}^{T}g_{s}(p_{s}) \leq B_{t}
 \right\} \\
 &= \max_{p_{t:T}} 
 \left\{
 V_{t}^{p_{t:T}}(\beta_{t}, \Sigma_{t}, B_{t} ) 
 ~~\Big|~~
  \sum\nolimits_{s=t}^{T}g_{s}(p_{s}) \leq B_{t}
 \right\}
\end{align*}
using the definition for the policy $\rho_{t}^{*}$.




\end{document}